%% file: main.tex
\documentclass{article}

    \PassOptionsToPackage{numbers, compress}{natbib}

\usepackage[preprint]{neurips_2025}
\usepackage{microtype}
\usepackage{graphicx}
\usepackage{amsmath}
\usepackage{amssymb}
\usepackage{booktabs}
\usepackage{makecell}
\usepackage{enumitem}
\usepackage{centernot}
\usepackage{amsfonts}
\usepackage[normalem]{ulem}
\usepackage{dsfont}
\usepackage{multirow}
\usepackage{cuted}
\usepackage{capt-of}
\usepackage{bbm}
\usepackage{balance}

\usepackage{natbib}
\usepackage{url}

\usepackage{subcaption}

\usepackage{xspace}
\usepackage{float}

\graphicspath{{figures/}}

\usepackage[capitalize,noabbrev]{cleveref}
\usepackage[textsize=tiny]{todonotes}

\newcommand{\modelname}{GRID}

\makeatletter
\DeclareRobustCommand\onedot{\futurelet\@let@token\@onedot}
\def\@onedot{\ifx\@let@token.\else.\null\fi\xspace}

\makeatother

\crefname{figure}{Fig.}{Figs.}
\Crefname{figure}{Fig.}{Figs.}

\title{
Grid: Omni Visual Generation}

\usepackage{authblk}

\author[1,$\dagger$]{Cong Wan}
\author[2,$\dagger$]{Xiangyang Luo}
\author[3]{Hao Luo}
\author[4]{Zijian Cai}
\author[5]{Yiren Song}
\author[1]{Yunlong Zhao}
\author[3,$\ddagger$]{Yifan Bai}
\author[3]{Fan Wang}
\author[1]{Yuhang He}
\author[1]{Yihong Gong}

\affil[1]{Xi'an Jiaotong University}
\affil[2]{Tsinghua University}
\affil[3]{DAMO Academy, Alibaba Group}
\affil[4]{Chinese Academy of Sciences}
\affil[5]{National University of Singapore}

\begin{document}

\maketitle
\vspace{-4mm}

\input{sections/0_abstract}

\input{sections/1_introduction}

\input{sections/3_method}

\input{sections/4_result}
\input{sections/5_capabilities}

\input{sections/2_related_work}

\input{sections/6_conclusion}

\clearpage

{
    \small
    \bibliographystyle{unsrtnat}
    \bibliography{main}
}

\newpage

\appendix
\onecolumn
\input{sections/7_appendix}

\end{document}

%% file: sections/0_abstract.tex
\begin{abstract}
Visual generation has witnessed remarkable progress in single-image tasks, yet extending these capabilities to temporal sequences remains challenging. Current approaches either build specialized video models from scratch with enormous computational costs or add separate motion modules to image generators, both requiring learning temporal dynamics anew. We observe that modern image generation models possess underutilized potential in handling structured layouts with implicit temporal understanding. Building on this insight, we introduce \modelname{}, which reformulates temporal sequences as grid layouts, enabling holistic processing of visual sequences while leveraging existing model capabilities. Through a parallel flow-matching training strategy with coarse-to-fine scheduling, our approach achieves up to \textbf{67$\times$} faster inference speeds while using \textbf{$<\frac{1}{1000}$} of the computational resources compared to specialized models. Extensive experiments demonstrate that \modelname{} not only excels in temporal tasks from Text-to-Video to 3D Editing but also preserves strong performance in image generation, establishing itself as an efficient and versatile \textbf{omni-solution} for visual generation.
\end{abstract}

%% file: sections/1_introduction.tex

\section{Introduction} 
\label{sec:intro}
Film strips demonstrate an elegant approach in visual arts: by arranging temporal sequences into structured grids, allowing time-based narratives to be displayed in layouts while maintaining their narrative coherence and visual connections. This organization does more than preserve chronological order - it enables efficient content manipulation, comparison, and editing. Drawing inspiration from this intuitive yet powerful organizational principle, we propose a fundamental question: \textbf{Can we directly reframe various temporal visual generation tasks as how to layout}, where key visual elements (such as multiple viewpoints or video frames) are treated as grid \textquotedblleft layout\textquotedblright?

\begin{figure*}[tp]
\centering
\includegraphics[width=\textwidth]{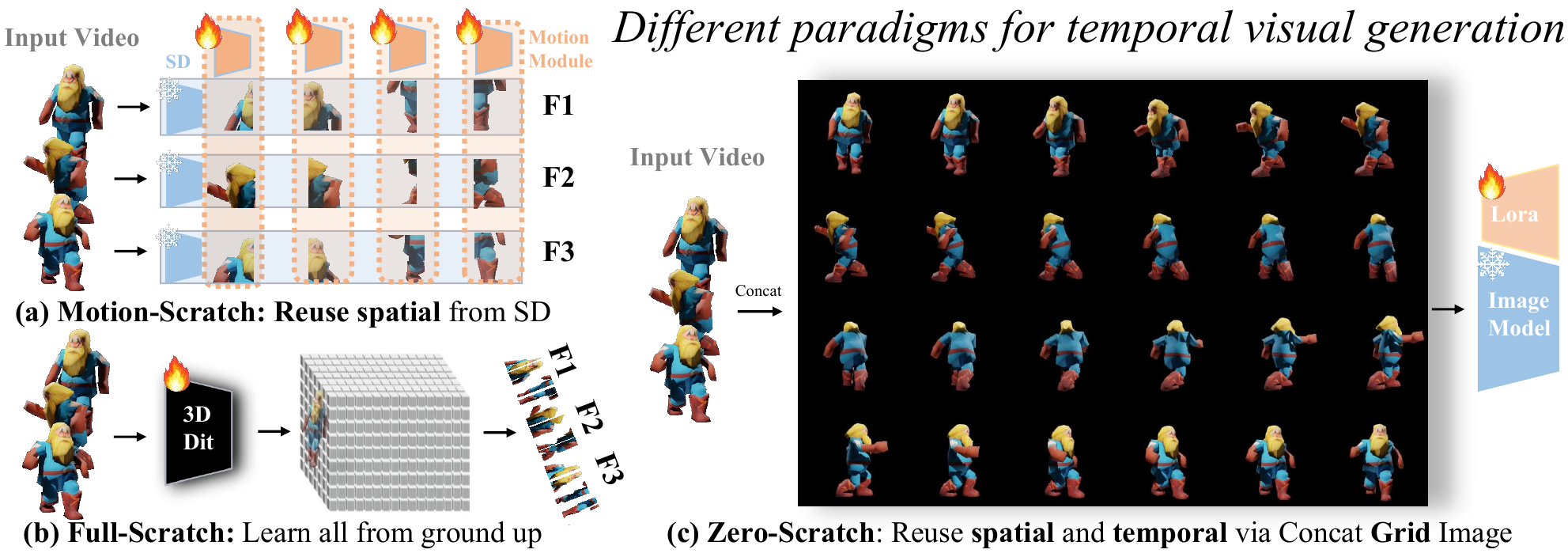}
\caption{
\textbf{Different paradigms for temporal visual generation}. 
(a) Motion-Scratch (e.g., SVD, AnimateDiff): learn temporal dynamics from scratch while reusing pretrained image models.
(b) Full-Scratch (e.g., Sora): learn everything from scratch, requiring massive data and computational resources. 
(c) Zero-Scratch (\textbf{GRID}): reuse both spatial and temporal capabilities through grid-based reformulation, leveraging pretrained models' inherent understanding.
}
\label{fig:compare}
\end{figure*}

To answer this, a natural starting point emerges from the recent breakthroughs in text-to-image generation. For single image generation, models like ~\cite{esser2024scaling, baldridge2024imagen, betker2023improving} have demonstrated remarkable capabilities in understanding and generating complex spatial relationships. For temporal visual generation, current approaches typically follow two distinct paths: (a) building specialized video models from scratch (e.g., Sora), which requires learning both spatial and temporal relationships with prohibitive computational costs, (b) treating image generators as single-frame producers and mainly train additional motion modules - while this avoids learning spatial generation from scratch, it still requires learning temporal dynamics entirely anew.

Guided by our layout-centric perspective, we argue that the inherent capabilities of image generation models are significantly underestimated. Modern image models already possess implicit understanding of both spatial relationships and basic temporal coherence, suggesting we might not need to learn either aspect entirely from scratch. To validate this hypothesis, we first test the ability of current image generation models to handle grid-arranged layouts through simple prompting (Figure \ref{fig:zero_eval}). Our experiments reveal that while these models show promising initial capabilities in understanding structured layouts, they still fall short in two fundamental aspects (detailed in Section~\ref{sec:appendix-zero-shot}):

\begin{itemize}
\item \textbf{Layout Control:} They fail to maintain both consistent grid structures and visual appearances across layouts.

\item \textbf{Motion Coherence:} When given specific motion instructions (\textit{e.g.}, \textquotedblleft rotate clockwise\textquotedblright), they cannot reliably create sequential movements across layouts.
\end{itemize}

To address these, we introduce \modelname{}, which \textbf{reformulates temporal sequences as grid layouts}, allowing image generation models to process the entire sequence holistically and learn both spatial relationships and motion patterns.

Building on this grid-based framework, we develop a \textbf{parallel flow-matching} training strategy that leverages large-scale web datasets, where video frames are arranged in grid layouts. The model learns to simultaneously generate all frames in these structured layouts through a base parallel matching loss, achieving consistent visual appearances and proper grid arrangements. This approach naturally utilizes the models' self-attention mechanisms to capture and maintain spatial relationships across the entire layout.

For precise motion control, we further incorporate dedicated temporal loss and motion-annotated datasets during fine-tuning. The temporal loss ensures smooth transitions between adjacent frames, while the motion annotations help learn specific patterns like ``rotate clockwise". These components are balanced through a coarse-to-fine training schedule to achieve both fluid motion and consistent spatial structure.

Through our carefully designed training paradigm, \modelname{} achieves remarkable efficiency gains, demonstrating a substantial \textbf{6-35$\times$} acceleration in inference speed compared to specialized expert models, while requiring merely \textbf{$\frac{1}{1000}$} of the training computational resources. Our framework exhibits exceptional versatility, achieving competitive or superior performance across a diverse spectrum of generation tasks, including Text-to-Video, Image-to-Video, and Multi-view generations, with performance improvements of up to \textbf{23$\%$}. Furthermore, we extend the capabilities of \modelname{} to encompass Video Style Transfer, Video Restoration, and 3D Editing tasks,
while preserving its original strong image generation capabilities for image tasks such as image editing and style transfer. This unique combination of expanded capabilities and preserved foundational strengths establishes \modelname{} as a \textbf{omni-solution} for visual generation.

\textbf{Our main contributions} are summarized as follows:

\begin{itemize}

\item \textbf{Novel Grid-based Framework:} We introduce a new paradigm that reformulates temporal sequences as grid layouts, enabling holistic processing of visual sequences through image generation models.

\item \textbf{Coarse-to-fine Training Strategy:} We develop a parallel flow-matching strategy combining layout matching and temporal coherence losses, with a coarse-to-fine training schedule that evolves from basic layouts to more precise motion control.

\item \textbf{Omni Generation:} We demonstrate strong performance across multiple visual generation tasks while maintaining low computational costs. Our method achieves results comparable to task-specific approaches, despite using a single, efficient framework.

\end{itemize}




\begin{figure*}[tp]
\centering
\includegraphics[width=\textwidth]{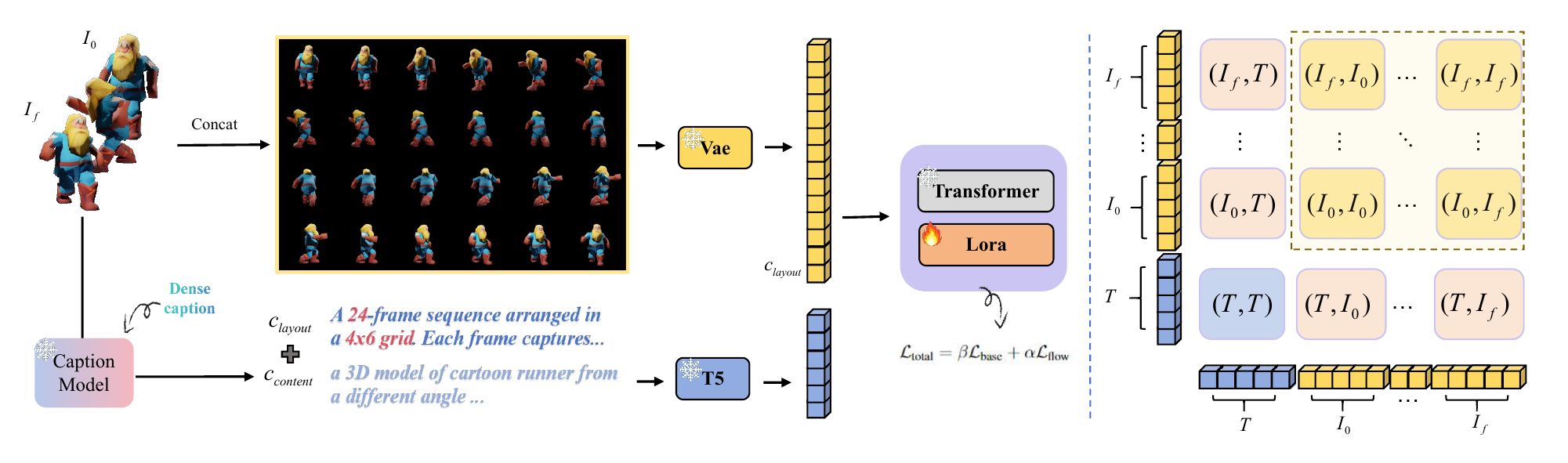}
\caption{
\textbf{Pipeline Overview}. Left: \modelname{} arranges videos into grid layouts, with text annotations combining layout format prefix and LLM-generated captions. The model is trained using LoRA fine-tuning on DIT blocks, incorporating both base loss and temporal loss to capture inter-frame relationships.
Right: Grid-based reformulation naturally extends model's built-in self-attention to include frame-wise self-attention, cross-frame attention, and text-to-frames cross-attention.
}
\label{fig:vffm_schematic}
\end{figure*}

%% file: sections/3_method.tex
\section{Layout Generation}
\label{sec:method}

Inspired by film strips that organize temporal sequences into structured grids, we present \modelname{}, a grid layout-driven framework that reformulates multiple visual generation tasks through grid-based representation. Our \modelname{} consists of three key components: 1) \textbf{Grid Representation}, which enables layout-based video organization for comprehensive visual generation; 2) \textbf{Parallel Flow Matching}, which ensures temporal coherence in successive grids; and 3) \textbf{Coarse-to-fine Training}, which enhances motion control capabilities. The framework architecture is illustrated in Figure~\ref{fig:vffm_schematic} (left).

\subsection{Grid Representation}
\label{subsubsec:grid_reform}

Existing text-to-image models, with inherent attention mechanisms, enable image manipulation and editing by generating new content from partial image information and semantic instructions, which inspires us to extend this capability to temporal generation by introducing a novel input paradigm, termed Grid Representation, that generates temporal content from keyframe visuals and semantic instructions.

Consider a general visual generation task that transforms an input condition $c_{content}$ (such as a text description $T$) into a sequence of images $(I_0, ..., I_f)$. We propose a grid layout specification $c_{\text{layout}}$ that arranges temporal frames into a structured grid within a single image, where each cell $(i,j)$ contains a specific image $I_{ij}$. As shown in Figure~\ref{fig:vffm_schematic} (right), when this grid structure is input into a conventional text-to-image model, the model's inherent attention mechanisms naturally extend their functionality to process this spatial arrangement as:

\begin{itemize}
    \item \textbf{Self-attention Expansion:} The standard self-attention mechanism $(I, I)$ (\textcolor{yellow}{yellow} block) expands into two distinct components:
        \begin{itemize}
        \item Intra-frame attention $(I_i, I_i)$: Maintains feature learning within individual grid cells
        \item Cross-frame attention $(I_i, I_j)$: Enables temporal relationships between different grid cells
        \end{itemize}
    
    \item \textbf{Cross-attention Extension:} The text-image cross-attention $(I, T)$ (\textcolor{pink}{pink} block) extends naturally to provide uniform text conditioning across all frame positions
\end{itemize}

Our approach demonstrates that thoughtful problem restructuring can be more effective than architectural modifications. By reorganizing the input space into a grid representation, standard text-to-image models can naturally handle temporal generation without architectural changes (see Appendix~\ref{subsec:attention} for detailed attention mechanism analysis). This grid-based design offers two key advantages:
First, it enables parallel generation of all frames and eliminates the error accumulation problems common in autoregressive approaches~\cite{tian2024visual}. Second, by leveraging the inherent consistency priors within pretrained image generation models, our approach effectively transfers their learned spatial consistency to temporal and multi-view coherence. This crucial advantage avoids the need for extensive pretraining on massive video datasets, as the grid representation naturally extends existing image-level understanding to sequence generation.
Additionally, through flexible layout conditioning ($c_{\text{layout}}$), our model shows strong generalization capabilities beyond training constraints as shown in Appendix and Supplementary Material, suggesting a promising solution to the fixed-length limitations of existing methods. Additionally, our grid representation supports diverse input types, including multi-view images and multi-frame sequences, laying the foundation for a comprehensive omni-generation model that bridges image and video domains.

\subsection{Parallel Flow Matching}
\label{subsubsec:training}

To fully leverage the potential of our grid representation, we employ parallel flow matching~\cite{esser2024scaling} to ensure temporal coherence across consecutive grids. For each training sample $\mathbf{I} = (I_{ij})$, we generate a corresponding text representation by integrating layout specifications with content descriptions: $c' = [c_{\text{layout}}, c_{\text{content}}]$. Here, $c_{\text{layout}}$ encodes the spatial structure (e.g., a sequence arranged in $m \times n$ grids), while $c_{\text{content}}$ captures the visual content as well as the temporal relationships between frames.

\textbf{Parallel Flow Evolution with Global Awareness.} 
Our grid representation integrates seamlessly with flow matching by organizing temporal frames into a unified grid image $\mathbf{I}$. This enables parallel evolution of frames through the following process:
\begin{equation}
\mathbf{I}_t = (1-t)\mathbf{I} + t\epsilon, \quad t \sim \mathcal{U}(0,1), \quad \epsilon \sim \mathcal{N}(0, I)
\end{equation}
Unlike autoregressive approaches that generate frames sequentially, our formulation allows all frames to evolve simultaneously from noise to target distribution through the model's native prediction process:
\begin{equation}
f: (\mathbf{I}_t, t, c') \rightarrow \epsilon - \mathbf{I}
\end{equation}
Each frame $(I_{ij})_t$ interacts with others within the grid, enabling mutual influence. This interaction naturally enforces temporal consistency across all sequences.

\subsection{Coarse-to-Fine Training}
\label{subsubsec:progressive}

Training models for temporal understanding in grid representation demands extensive video data to achieve key capabilities like identity preservation and motion consistency - essential features for video and multi-view generation that text-to-image models typically lack. This training process faces two main challenges from mixed quality of available data: the abundance of low-quality internet videos, and high computational costs of processing high-resolution footage. We tackle these limitations through a coarse-to-fine training strategy that combines two key components: data curriculum and loss dynamic. This dual approach optimizes both training efficiency and model performance, enabling effective use of diverse data sources while minimizing computational overhead. Our strategy enhances the capabilities of our flow-based framework without sacrificing training efficiency.

\textbf{Data Curriculum.} Our training strategy follows a Coarse-to-Fine approach, starting with foundational learning and advancing to refinement:
\begin{itemize}
    \item \textit{Coarse Phase:} In the initial phase, we utilize large-scale Internet datasets, including WebVid, TikTok, and Objaverse, which are designed with uniform $c_{\text{layout}}$ specifications. Although the content descriptions ($c_{\text{content}}$) are automatically generated by GLM-4V-9B~\cite{DBLP:conf/acl/DuQLDQY022} and may lack precise control details, the vast scale and diversity of this data—albeit at lower resolutions—provide a strong basis for developing robust spatial understanding and basic layout structures.
    \item \textit{Fine Phase:} Building on the foundational knowledge from the coarse phase, we transition to training with carefully curated, high-resolution samples. These samples are paired with detailed descriptions generated by GPT-4~\cite{openai2023gpt4}, offering explicit spatial and temporal instructions. As shown in Figure~\ref{fig:vffm_schematic}, these high-quality captions facilitate fine-grained control over complex layout variations, enabling the model to handle intricate spatial and temporal dynamics effectively.
\end{itemize}

\textbf{Loss Formulation.} Our training objective combines appearance accuracy with temporal consistency through a weighted sum:
\begin{equation}
\mathcal{L}_{\text{total}} = \mathcal{L}_{\text{base}} + \alpha\mathcal{L}_{\text{flow}}
\end{equation}
The base loss ensures accurate noise prediction at each position using mean squared error:
\begin{equation}
\mathcal{L}_{\text{base}} = \mathbb{E}_{t,\epsilon}[|\epsilon - \epsilon_\theta(\mathbf{I}, t, c')|^2]
\end{equation}
The flow loss enforces smooth temporal transitions by penalizing inconsistent changes between adjacent positions. For any position (i,j) in the grid, directional changes are:
\begin{equation}
\begin{aligned}
\Delta\epsilon^{ij} &= \begin{cases}
\epsilon^{ij} - \epsilon^{i,j-1} & \text{within row} \\
\epsilon^{i,0} - \epsilon^{i-1,n} & \text{across rows}
\end{cases}
\end{aligned}
\end{equation}
Similarly for predicted values:
\begin{equation}
\begin{aligned}
\Delta\epsilon_\theta^{ij} &= \begin{cases}
\epsilon_\theta^{ij} - \epsilon_\theta^{i,j-1} & \text{within row} \\
\epsilon_\theta^{i,0} - \epsilon_\theta^{i-1,n} & \text{across rows}
\end{cases}
\end{aligned}
\end{equation}
The flow loss then minimizes inconsistencies in these directional changes:
\begin{equation}
\mathcal{L}_{\text{flow}} = \mathbb{E}_{t,\epsilon}[|\Delta\epsilon - \Delta\epsilon_\theta(\mathbf{I}, t, c')|^2]
\end{equation}
The weight $\alpha$ gradually increases from 0 to a preset upper bound, allowing the model to first establish precise content generation capabilities before focusing on temporal dynamics. This staged evolution of the loss function complements our data curriculum, enabling the model to effectively learn both the spatial and temporal aspects of generation in a coordinated manner.

\begin{figure*}
\centering
\includegraphics[width=.99\textwidth,trim={0 0 0 0},clip]{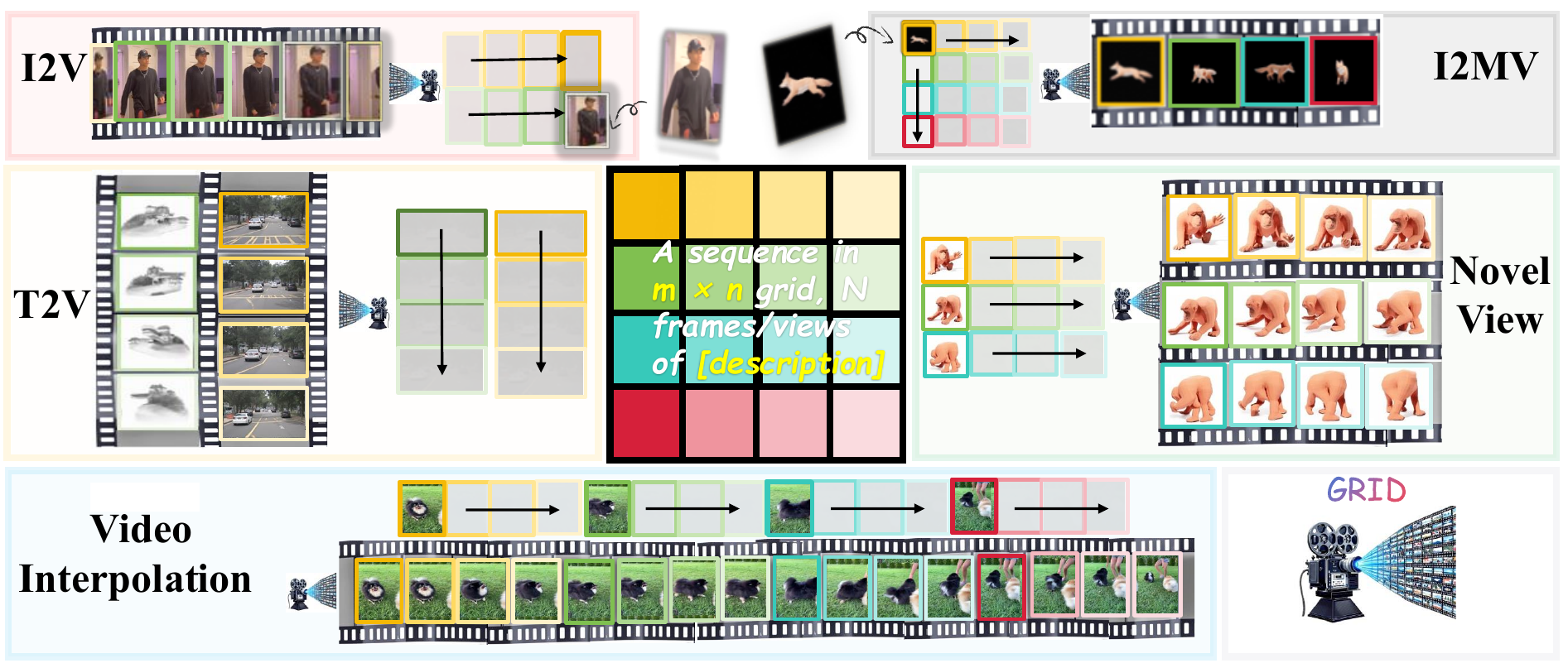}
\caption{\textbf{Omni Inference Framework}: By transforming temporal and view sequences into structured layout spaces, we enable a pure image-based model FLUX to tackle diverse video and multi-view tasks (text/image-to-video generation, video interpolation, and multi-view synthesis) through a unified pipeline without additional video-specific architectures.}
\label{fig:info_graphic}
\end{figure*}

\subsection{Omni Inference}
\label{subsec:unified_framework}

We propose an omni-inference framework designed to handle a wide range of generation tasks using a reference-guided grid layout initialization. The core idea of our approach is to unify different generation tasks by employing a well-structured initialization process combined with controlled grid noise injection. At the same time, we ensure consistency with the reference through the use of a binary mask.

Given a reference image $I_{\text{ref}}$ or key frames $(I_0,...,I_{m-1})$, we construct a grid structure $\mathbf{I}=(I_{ij})_{m\times n}$. For single-image expansion and frame interpolation tasks, we initialize the grid as:
\begin{equation}
I_{ij} = \begin{cases}
I_{\text{ref}} & \text{expansion} \\
(1-\frac{j}{n})I_{i,0} + \frac{j}{n}I_{i+1,0} & \text{interpolation}
\end{cases}
\end{equation}
The generation process requires both flexibility and reference consistency. To achieve this, we introduce controlled grid noise injection instead of starting from pure noise:
\begin{equation}
\mathbf{I}_T = (1-T)\mathbf{I} + T\epsilon, \quad \epsilon \sim \mathcal{N}(0, I)
\end{equation}
where $T$ denotes the time. This noise injection enables diverse generation while retaining the initialization structure.
To maintain reference consistency during generation, we employ a binary mask $M \in \{0,1\}^{m\times n}$:
\begin{equation}
M_{ij} = \begin{cases}
0 & \text{if } (i,j) \text{ contains reference frame} \\
1 & \text{otherwise}
\end{cases}
\end{equation}
This mask modulates the update process:
\begin{equation}
\mathbf{I}_t = (1-M) \odot \mathbf{I}_{\text{ref}} + M \odot \mathbf{I}_t
\end{equation}
ensuring reference frames remain unchanged while allowing other regions to evolve.
The noise level $T$ plays a key role in balancing generation quality. A large $T$ leads to pure noise with poor reference consistency, while a small $T$ yields near-duplicates. Our experiments show $T \in [0.8,1.0]$ a good balance between diversity and fidelity.

%% file: sections/4_result.tex

\section{Experiments} \label{sec:experiments}

\subsection{Experimental Setup}

\paragraph{Datasets} 
We train our model separately for video generation and multi-view generation tasks, both following a two-stage strategy:
(1) For coarse-level training, we combine video clips from WebVid~\cite{bain2021frozen}, and TikTok~\cite{jafarian2022self} arranged in 8×8 and 4×4 grid layouts for video generation, and 30K sequences from Objaverse~\cite{deitke2023objaverse} in 4×6 grids for multi-view generation. Each sequence is paired with automated captions and GLM-generated annotations emphasizing spatial and temporal relationships, using the sequence's inherent attributes (e.g., category labels) and visual content as queries. 
(2) For fine-grained control, we construct high-quality datasets of 1K sequences with structured annotations for each task. We first manually create exemplar annotations to establish a consistent format, then use these as few-shot examples for GPT-4o to generate precise control instructions while maintaining annotation consistency across the dataset.

\paragraph{Implementation Details} 
We implement \modelname{} based on the FLUX-dev, initializing from its pretrained weights. For video generation training, we adopt LoRA with ranks of 16-256, training for 10K steps with batch size 4 across 8 A800 GPUs using AdamW optimizer (learning rate 1e-4). The temporal loss weight $\alpha$ starts from 0 and gradually increases to a maximum of 0.5. For multi-view generation, we train on 30K sequences for 1.5K steps using LoRA rank 256 and Ours-EF using LoRA rank 16. During inference, we use a guidance scale of 3.5 and sampling step of 20.

\paragraph{Evaluation Protocol} 
We evaluate our model on three distinct generation tasks: 
(1) Text-to-video generation on UCF-101 dataset~\cite{soomro2012ucf101}, evaluated using FVD~\cite{unterthiner2019fvd} (I3D backbone) and IS~\cite{xu2018empirical}. We evaluate both 16-frame and 64-frame generation settings;
(2) Image-to-video generation on a randomly sampled subset of 100 TikTok videos, measured by FVD and CLIP$_{img}$ score;
(3) Multi-view generation on Objaverse, where we evaluate on 30 randomly selected objects with 24 frames per sequence at different viewpoints to assess 4D generation capabilities. We compute FVD, CLIP metrics, following~\cite{diffusion4d}.

\input{tables/t2v}
\subsection{Main Results}
We compare our approach with state-of-the-art methods across key domains: multiview generation (Animate124~\cite{zhao2023animate124}, 4DFY~\cite{bahmani20244dfy}, STAG4D~\cite{zeng2024stag4d}, 4DGen~\cite{yin20234dgen}), text-to-video and image-to-video generation (AnimateDiffv3~\cite{guo2023animatediff}, OpenSora1.2~\cite{opensora}, Cosmos~\cite{agarwal2025cosmos}, CogVideo5b~\cite{cogvideox}), and video frame interpolation (EMA-VFI~\cite{zhang2023extracting}, UPR-NetV~\cite{jin2023unified}, FIMamba~\cite{zhang2024vfimambavideoframeinterpolation}).

\paragraph{Multi-view Generation}
We evaluate on the Objaverse test set with 30 3D objects. As shown in Table~\ref{tab:multiview}, our method achieves \textbf{state-of-the-art performance} on both text-to-multiview and image-to-multiview tasks. For T2MV, we improve CLIP-F to \textbf{0.9427} and reduce FVD to \textbf{324.3}, while achieving \textbf{67$\times$} faster inference (6m vs. 405m) compared to 4DFY. For I2MV, we achieve \textbf{0.9486} CLIP-F score with \textbf{35$\times$} speedup over STAG4D. Ours-EF (lora rank 16) also demonstrates strong performance-speed trade-off.

\paragraph{Text-to-Video Generation}
As shown in Table~\ref{tab:comprehensive_results}, we achieve competitive FVD of 721.6 for 64-frame generation. For 16-frame generation, our method achieves \textbf{6.7$\times$ faster inference} (7.2s vs 48s) compared to CogVideo, with the efficiency gap widening to \textbf{5.5$\times$} for 64-frame tasks. Our staged training shows clear progression: Stage1 achieves FVD 455.3, improving to \textbf{401.1} with fine-grained annotations, and further to \textbf{382.5} with $\mathcal{L}_{flow}$.

\paragraph{Image-to-Video Generation}
We evaluate on the TikTok dataset containing 100 diverse short videos. Our method achieves breakthrough performance with FVD of \textbf{93.7} (\textbf{23\%} improvement) and CLIP$_{img}$ score of \textbf{0.9709}. Notably, our approach requires only \textbf{160M} parameters, compared to $>$400M for motion modeling or $>$1B for full generation in existing methods.

\paragraph{Video Frame Interpolation}
We evaluate on the UCF101 dataset for video frame interpolation~\cite{zhang2023extracting,jin2023unified,zhang2024vfimambavideoframeinterpolation}. As shown in Table~\ref{tab:vfi_results}, our approach achieves \textbf{state-of-the-art PSNR of 35.48}, matching EMA-VFI. For SSIM, all methods perform comparably around 0.970.

%% file: tables/t2v.tex
\begin{table*}[tp]
\centering
\caption{Quantitative comparison of Multi-view Generation Results on Text-to-Multiview and Image-to-Multiview Tasks. 
Time indicates the \textbf{whole} time cost during inference in A800.
}
\small
\resizebox{0.99\linewidth}{!}{%
\begin{tabular}{l|cccc||l|cccc}
\toprule
\multicolumn{5}{c||}{\textbf{Text-to-Multiview (T2MV)}} & 
\multicolumn{5}{c}{\textbf{Image-to-Multiview (I2MV)}} \\
\midrule
Method & CLIP-F↑ & CLIP-O↑ & FVD↓ & Time↓ &
Method & CLIP-F↑ & CLIP-O↑ & FVD↓ & Time↓ \\
\midrule
Animate124 & 0.7889 & 0.6005 & 411.6 & 180m &
STAG4D & 0.8803 & 0.6420 & 475.4 & 210m \\
4DFY & 0.8092 & 0.6163 & 390.4 & 405m &
4DGen & 0.8724 & 0.6397 & 525.2 & 130m \\
Ours-EF & 0.9060 & 0.6189 & 355.6 & 6m &
Ours-EF & 0.9392 & \textbf{0.6580} & \textbf{333.7} & 6m \\
\textbf{Ours} & \textbf{0.9427} & \textbf{0.6247} & \textbf{324.3} & \textbf{6m} &
\textbf{Ours} & \textbf{0.9486} & 0.6554 & 350.6 & \textbf{6m} \\
\bottomrule
\end{tabular}
}
\label{tab:multiview}
\end{table*}

\begin{table*}[tp]
\centering
\caption{\textbf{Comprehensive Generation Results.} Our model achieves competitive quality with \textbf{superior efficiency} across tasks. While existing methods are limited to 16-frame generation, our approach efficiently scales to 64-frame sequences with linear time cost. \underline{Underlined} and \textbf{bold} values indicate best results among our variants and all methods, respectively. Time shows average sampling time per sequence in A800 GPU. Para means \textbf{training} parameters.}
\small
\resizebox{0.99\linewidth}{!}{%
\begin{tabular}{l|ccc|ccc|ccc|c}
\toprule
\multirow{2}{*}{Method} & \multicolumn{3}{c|}{\textbf{Text-to-Video (16-frame)}} & \multicolumn{3}{c|}{\textbf{Text-to-Video (64-frame)}} & \multicolumn{3}{c|}{\textbf{Image-to-Video}} & \multirow{2}{*}{Para$\downarrow$} \\
\cmidrule(lr){2-4} \cmidrule(lr){5-7} \cmidrule(lr){8-10}
& FVD$\downarrow$ & IS$\uparrow$ & Time$\downarrow$ & FVD$\downarrow$ & IS$\uparrow$ & Time$\downarrow$ & FVD$\downarrow$ & CLIP$_{img}\uparrow$ & Time$\downarrow$ & \\
\midrule
AnimateDiffv3 & 464.1 & 35.24 & 12s & - & - & - & 250.9 & 0.9229 & 12s & 419M \\
VideoCrafter2 & 424.2 & 32.00 & 15s & - & - & - & - & - & - & 919M \\
OpenSora1.2 & 472.0 & \bf 39.07 & 12s & 1000.5 & \bf 37.11 & 66s & - & - & - & 1.5B \\
Cosmos & 399.7 & 35.54 & 275s & - & - & - & - & - & - & 7B \\
CogVideo5b & 410.5 & 38.8 & 13s & 740.1 & 34.82 & 132s & 122.5 & 0.9185 & 48s & 5B \\
\midrule
Ours(Stage1) & 455.3 & 32.46 & 7.2s & 1003.2 & 32.48 & 24s & 115.5 & 0.9598 & 7.2s & \multirow{3}{*}{\bf 160M} \\
Ours(Stage1+2) & 401.1 & 36.56 & 7.2s & 994.6 & 36.47 & 24s & 104.6 & 0.9695 & 7.2s \\
Ours(Full) & \bf \underline{382.5} & \underline{38.12} & \bf \underline{7.2}s & \bf \underline{721.6} & \underline{36.63} & \bf \underline{24}s & \bf \underline{93.7} & \bf \underline{0.9709} & \bf \underline{7.2}s \\
\bottomrule
\end{tabular}
}
\label{tab:comprehensive_results}
\end{table*}

\begin{table}[tp]
\centering
\caption{\textbf{Video Frame Interpolation Results on UCF101.} We evaluate our full model following standard settings. All methods achieve comparable results, with our approach matching state-of-the-art EMA-VFI on PSNR.}
\small
\begin{tabular}{l|ccccc}
\toprule
Metrics & ToFlow & EMA-VFI & UPR-Net & VFIMamba & Ours \\
\midrule
PSNR$\uparrow$ & 34.58 & \bf 35.48 & 35.47 & 35.45 & \textbf{35.48} \\
SSIM$\uparrow$ & 0.9677 & 0.9701 & 0.9700 & \bf 0.9702 & 0.9700 \\
\bottomrule
\end{tabular}
\label{tab:vfi_results}
\end{table}

%% file: sections/5_capabilities.tex
\subsection{Extension Capabilities}
Beyond the primary generation tasks, we demonstrate \modelname{}'s strong zero-shot generalization capabilities across diverse video and multi-view applications without any task-specific training or architectural modifications. The layout-based design enables natural adaptation to various downstream tasks through prompt engineering alone.

\paragraph{Video Motion Clone}
Our framework enables natural video motion cloning through image redrawing without additional training. As demonstrated in Figure~\ref{fig:style_transfer}, we transform a cat video into videos featuring a fox, red panda, and tiger, while faithfully preserving the original motion patterns, temporal dynamics, and scene aesthetics. 

\paragraph{Video Restoration}
Our architecture's multi-scale processing capability enables effective video restoration without explicit training. Figure~\ref{fig:restoration} shows our model's performance in recovering high-quality videos from severely degraded inputs (with Gaussian blur and block masking). 

\paragraph{3D Editing}
We demonstrate our model's potential for practical 3D appearance editing through an innovative virtual try-on application. As shown in Figure~\ref{fig:rerender}, given an uncolored 3D human walking sequence from multiple viewpoints, our model can dress and style the figure through simple text prompts. This enables diverse appearance variations - from adding hair to rendering outfits - while maintaining consistent 3D structure and motion. 



More results and applications are shown in Appendix~\ref{appendix:pa}.

%% file: sections/2_related_work.tex
\section{Related Work} 
\label{sec:related_work}

\paragraph{Text-to-Image Generation}
Diffusion models~\cite{sohl2015deep,ddpm} have fundamentally transformed image generation by employing iterative denoising processes to synthesize high-quality outputs. Subsequent advancements~\cite{LDM,sdxl,ramesh2022hierarchical,Saharia2022} have refined this paradigm leveraging latent spaces with significantly reduced computational costs. 
Diffusion Transformers (DiT)~\cite{dit} further advanced this area by replacing the U-Net architecture with transformer-based designs. This architectural shift improved training efficiency, paving the way for more scalable and versatile generative frameworks. Building on these, flow matching~\cite{lipman2022flow,esser2024scaling} reformulates the generation process as a straight-path trajectory between data and noise distributions. 
More recently, FLUX~\cite{flux}, has combined the strengths of DiT and flow matching to achieve efficient and high-quality image generation. These models also integrate powerful language models~\cite{raffel2020exploring} and joint text-image attention mechanisms. This multimodal understanding has unlocked new possibilities for instruction-following and creative applications.
Beyond generating high-quality images, text-to-image models demonstrate a strong spatial understanding that can be naturally extended to temporal dimensions through layout representations, enabling diverse downstream tasks.

\paragraph{Task-Specific Generation}
Diffusion-based approaches have shown remarkable progress in generalized video generation tasks~\cite{ho2022imagen,blattmann2023align,zhang2023show,blattmann2023stable,he2023latent,zhou2022magicvideo,wang2023modelscope,ge2023preserve,wang2023internvid,wang2023videofactory,singer2022make,zhang2023show,zeng2023make,agarwal2025cosmos}. Notable works like VideoLDM~\cite{blattmann2023align}, Animatediff~\cite{guo2023animatediff}, and SVD~\cite{chai2023stablevideo} advance temporal modeling through specialized architectures. In the multi-view domain, various approaches~\cite{3DiM,zero1to3,mvdream,wonder3d,zero123++,Direct2.5,Instant3D,syncdreamer,Era3d,hunyuan3d,zhao2023animate124,yin20234dgen} focus on cross-view consistency through different attention mechanisms and feature space alignments. Recent 4D generation methods~\cite{dreamgaussian4d,diffusion4d,sv4d,dimensionx,cat4d} further extend to joint spatial-temporal synthesis, though often facing efficiency challenges or requiring multi-step generation. 
While these methods achieve remarkable results, they are typically tailored to specific tasks, relying on specialized architectures for image, video, or multi-view generation. Additionally, methods like VideoPoet~\cite{kondratyuk2023videopoet} employ complex cross-modal alignment mechanisms to bridge different generation modes.
In contrast, our approach introduces layout generation, an omni framework that transforms temporal and spatial generation into layout representations. This enables seamless multi-modal generation, to address a wide range of tasks through straightforward modifications to input representations, without the need for complex cross-modal alignment mechanisms.

%% file: sections/6_conclusion.tex
\vspace{-2mm}
\section{Conclusion}
\vspace{-1mm}
We present \modelname{}, an omni visual generation framework through grid representation. Our two-stage training strategy enables both robust generation and precise control, while the temporal refinement mechanism enhances motion coherence. Experiments demonstrate significant computational efficiency gains while maintaining competitive performance across tasks. The framework's strong zero-shot generalization capabilities further enable adaptation to diverse applications without task-specific training, suggesting a promising direction for efficient visual sequence generation.

%% file: sections/7_appendix.tex
\section{Appendix} \label{sec:appendix}

\subsection{Why Flux? Zero-shot Analysis of Foundation Models} \label{sec:appendix-zero-shot}

To better understand the layout capabilities of existing models before fine-tuning, we conducted a comprehensive zero-shot evaluation comparing three state-of-the-art models: DALLE-3, Flux, and Imagen3. Figure~\ref{fig:zero_eval} presents their generation results, with each row corresponding to DALLE-3 (top), Flux (middle), and Imagen3 (bottom) respectively.

\input{figures/layout_fig_zero_eval.tex}

Our analysis reveals varying degrees of grid layout understanding across models. While all models demonstrate basic grid comprehension, they exhibit different strengths and limitations. For motion control, we observe that precise directional instructions (e.g., clockwise rotation) often result in random orientations across all models, indicating limited spatial-temporal control capabilities.

In terms of grid structure accuracy, DALLE-3 shows inconsistent interpretation of specific layout requirements (e.g., 4×4 or 4×6 grids), while Flux and Imagen3 demonstrate better adherence to specified grid configurations. Notably, Flux exhibits superior understanding of spatial arrangements.

Content consistency across grid cells varies significantly. Both Imagen3 and DALLE-3 show noticeable variations in object appearance across frames, while Flux maintains better consistency in object characteristics throughout the sequence. This superior consistency, combined with its open-source nature, motivated our choice of Flux as the base model for our framework.


\subsection{Why is it Natural for \modelname{} to Leverage Built-in Attention Mechanism}
\label{subsec:attention}

Video generation fundamentally requires three key capabilities: spatial understanding within frames, temporal consistency between frames, and semantic control across the entire sequence. Traditional approaches tackle these requirements by implementing separate attention modules, as shown in Figure~\ref{fig:attention}(a). While this modular design directly addresses each requirement, it introduces architectural complexity and potential inconsistencies between modules.

\input{figures/layout/tex/attention}

Our key insight is that these seemingly distinct requirements can be unified through spatial reformulation. By organizing temporal sequences into grid layouts, we transform temporal relationships into spatial ones, allowing FLUX's native attention mechanism to naturally handle all requirements through a single, coherent process.

This unification works through two complementary mechanisms, as illustrated in Figure~\ref{fig:attention}(b). First, the original image self-attention $(I,I)$ automatically extends across the grid structure. When processing grid cells containing different temporal frames, this self-attention naturally splits into inner-frame attention $(I_i,I_i)$ and cross-frame attention $(I_i,I_j)$. The inner-frame component maintains spatial understanding within each frame, while the cross-frame component captures temporal relationships - effectively handling both spatial and temporal coherence through a single mechanism.

Second, the text-image cross-attention $(T,[I_i]_{i=0}^f)$ operates globally across all grid cells, enabling unified semantic control. This global operation ensures that textual instructions consistently influence all frames, maintaining semantic coherence throughout the sequence. The grid layout allows this semantic guidance to naturally incorporate both content and temporal specifications, as the attention mechanism can reference the spatial relationships between grid cells.

This reformulation fundamentally changes how temporal information is processed. Rather than treating temporal relationships as a separate problem requiring specialized mechanisms, we transform them into spatial relationships that existing attention mechanisms are already optimized to handle. This approach not only simplifies the architecture but also provides more robust temporal understanding, as it leverages the well-established capabilities of spatial attention mechanisms.

The elegance of this solution lies in its ability to achieve complex temporal processing without architectural modifications. By thoughtfully restructuring the problem space, we enable standard attention mechanisms to naturally extend their capabilities, demonstrating how strategic problem reformulation can be more powerful than architectural elaboration.

\subsection{Comparison with Existing Approaches and Computational Efficiency Analysis}

\label{subsec:computation_efficiency}
Current approaches to video generation can be categorized into two distinct paradigms, each with fundamental limitations in terms of architectural design and computational requirements. We provide a detailed analysis of these approaches and contrast them with our method:

\textbf{Paradigm 1: Image Models as Single-Frame Generators}\
Methods like SVD and AnimateDiff utilize pre-trained text-to-image models as frame generators while introducing separate modules for motion learning. This approach presents several fundamental limitations:

First, these methods require complex architectural additions for temporal modeling, introducing significant parameter overhead without leveraging the inherent capabilities of pre-trained image models. For instance, AnimateDiff introduces temporal attention layers that must be trained from scratch, while SVD requires separate motion estimation networks.

Second, the sequential nature of frame generation in these approaches leads to substantial computational overhead during inference. This sequential processing not only impacts generation speed but also limits the model's ability to maintain long-term temporal consistency, as each frame is generated with limited context from previous frames.

\textbf{Paradigm 2: End-to-End Video Architectures}\
Recent approaches like Sora, CogVideo, and Huanyuan Video attempt to solve video generation through end-to-end training of video-specific architectures. While theoretically promising, these methods face severe practical constraints:

The computational requirements are particularly striking:
\begin{itemize}
\item CogVideo requires approximately 35M video clips and an additional 2B filtered images from LAION-5B and COYO-700M datasets
\item Open-Sora necessitates more than 35M videos for training
\item These models typically demand multiple 80GB GPUs with sequence parallelism just for inference
\item Training typically requires thousands of GPU-days, making reproduction and iteration challenging for most research teams
\end{itemize}


\textbf{Our Grid-based Framework: A Resource-Efficient Alternative}\
In contrast, \modelname{} achieves competitive performance through a fundamentally different approach:

\textbf{1. Architectural Efficiency:}
Our grid-based framework requires only 160M additional parameters while maintaining competitive performance. This efficiency stems from:
\begin{itemize}
\item Treating temporal sequences as spatial layouts, enabling parallel processing
\item Leveraging existing image generation capabilities without architectural complexity
\item Efficient parameter sharing across temporal and spatial dimensions
\end{itemize}

\textbf{2. Data Efficiency:}
We achieve remarkable data efficiency improvements:
\begin{equation}
\text{Data Reduction} \approx \frac{>35M \text{ videos (previous methods)}}{<35K \text{ videos (our method)}} = 1000\times
\end{equation}

This efficiency is achieved through:
\begin{itemize}
\item Strategic use of grid-based training that maximizes information extraction from each video
\item Effective transfer learning from pre-trained image models
\item Focused training on essential video-specific components
\end{itemize}

\textbf{3. Computational Accessibility:}
Our approach enables high-quality video generation while maintaining accessibility for research environments with limited computational resources:
\begin{itemize}
\item Training can be completed on standard research GPUs
\item Inference requires significantly less memory compared to end-to-end approaches
\item The model maintains strong performance across both video and image tasks
\end{itemize}

This comprehensive analysis demonstrates that our approach not only addresses the limitations of existing methods but also achieves substantial improvements in computational efficiency while maintaining competitive performance. The significant reductions in data requirements and computational resources make our method particularly valuable for practical applications and research environments with limited resources.

\subsection{Distinction from Grid-based Methods}
\label{sec:appendix-grid-methods}

Several recent works utilize grid-based layouts for image generation, including IC-LoRA \cite{lhhuang2024iclora,lhhuang2024groupdiffusion}, Instant3D \cite{li2023instant3d}, DSD \cite{cai2024dsd}, OmniControl \cite{xie2024omnicontrol}. While these approaches might superficially appear similar to ours, a careful analysis reveals fundamental differences in both theoretical foundation and technical implementation.

\textbf{Different Theoretical Foundations:}
Prior grid-based methods primarily focus on appearance consistency in static image generation. For instance, Instant3D utilizes grids to generate consistent multi-view images but lacks dedicated modeling of temporal dynamics. Similarly, DSD leverages grid structures for self-distillation to improve image quality, while OmniControl introduces Spatial Control Signals to achieve static control but does not address temporal relationships. IC-LoRA uses grid layouts as a prompt engineering technique, where multiple images are arranged together to provide in-context examples for task adaptation.

In contrast, our approach fundamentally re-conceptualizes temporal sequences into spatial layouts. Rather than using grids for static consistency or example presentation, we treat them as an inherent representation of temporal information, where spatial relationships in the grid directly correspond to temporal relationships in the sequence. This enables our model to learn and generate temporal dynamics in a holistic manner.

\textbf{Distinct Technical Objectives:}
While Flux shows potential for grid-based generation, it exhibits several limitations. As demonstrated in our experiments (Figure 8), Flux struggles with understanding grid structures, leading to inconsistencies in structure and appearance, poor performance in cross-frame object consistency, and unreliable responses to motion instructions. Other methods like IC-LoRA rely on LoRA-based fine-tuning and natural language prompts to define relationships between grid elements, treating each grid element independently without explicit modeling of their temporal relationships.

Our method, specifically designed for temporal sequence generation, introduces parallel flow-matching and dedicated temporal loss functions that explicitly model motion patterns and temporal coherence. This allows our approach to capture and generate complex temporal dynamics that are beyond the capability of existing grid-based methods. We efficiently incorporate diverse data and world knowledge (3D, video) through our omni paradigm, transcending the limitations of traditional grid methods that only focus on appearance consistency.

\textbf{Different Application Scopes:}
While prior grid-based methods excel at static image generation or limited motion capabilities, they struggle with temporal sequence generation due to fundamental design limitations. Our method naturally handles both static and dynamic visual generation tasks while maintaining precise control over temporal dynamics, supporting a wide range of applications from text-to-video and multi-view generation to video editing.

These crucial differences are evidenced by our method's superior performance in temporal tasks and its ability to maintain consistent motion patterns across sequences - capabilities that are fundamentally beyond the scope of existing grid-based approaches.

\subsection{Implementation Details}
\label{subsec:implementation_details}

\textbf{Training Configurations:}
For training our model across different tasks, we employ the following configuration details:

\begin{itemize}
\item \textbf{Video Generation (High Resolution)}: 4×4 grid format with single frame resolution of 512×768, resulting in a total resolution of 2048×3072
\item \textbf{Video Generation (Extended Frames)}: 8×8 grid format with single frame resolution of 256×384, resulting in a total resolution of 2048×3072
\item \textbf{Multi-view Generation}: 4×6 grid format with single frame resolution of 256×256, resulting in a total resolution of 1024×1536
\end{itemize}

\textbf{Model Variant Details:}
\begin{itemize}
\item \textbf{Stage1}: Trained only on large-scale data like WebVid and TikTok, using automatically generated descriptions
\item \textbf{Stage2}: Fine-tuned on Stage1 using high-quality descriptions generated by GPT-4, but without temporal loss
\item \textbf{Full}: Built upon Stage2 with the addition of temporal loss ($\mathcal{L}_{flow}$)
\end{itemize}

\textbf{Extension Inference Details:}
For extension tasks (style transfer, restoration, and editing), we modify the omni-inference framework to process full sequences while maintaining temporal coherence. Unlike the reference-guided generation that requires partial initialization and masking, these tasks operate on complete sequences with controlled noise injection for appearance modification.

Given an input sequence represented as a grid structure $\mathbf{I}=(I_{ij})_{m\times n}$, we initialize the generation process with noise-injected states:
\begin{equation}
\mathbf{I}_T = (1-T)\mathbf{I} + T\epsilon, \quad \epsilon \sim \mathcal{N}(0, I)
\end{equation}
where $T \in [0.8,0.9]$ represents a lower noise level compared to the reference-guided generation. This lower $T$ value helps preserve the original temporal structure while allowing sufficient flexibility for appearance modifications.
\subsection{Post-Processing Pipeline}
\label{subsec:post_processing}

For multi-view generation results, we employ a two-stage enhancement process. First, the generated sequences are processed as video frames to ensure temporal consistency. Subsequently, we apply super-resolution using Real-ESRGAN~\cite{wang2021realesrgan} with anime-video-v3 weights, upscaling from 256×256 resolution to 1024×1024. This enhancement pipeline significantly improves visual quality while maintaining temporal coherence.

Table~\ref{tab:prompts} shows parts of our inference prompts for multyview generation.
We basically follow this prompt format.

\begin{table*}[tp]
\centering
\begin{tabular}{lp{11cm}}
\toprule
\textbf{Common Format} & A 24-frame sequence arranged in a 4x6 grid. Each frame captures a 3D model of [subject] from a different angle, rotating 360 degrees. The sequence begins with a front view and progresses through a complete clockwise rotation \\
\midrule
\textbf{Category} & \textbf{Subject Description} \\
\midrule
Creative Fusion & a skyscraper with knitted wool surface and cable-knit details \\
               & a mechanical hummingbird with clockwork wings and steampunk gears hovering near a neon flower \\
               & a bonsai tree with spiral galaxies and nebulae blooming from its twisted branches \\
               & a phoenix crafted entirely from woven bamboo strips with intricate basketwork details glowing from within \\
               & a jellyfish with a transparent porcelain bell decorated in blue-and-white patterns and ink-brush tentacles \\
               & a coral reef made entirely of rainbow-hued blown glass with intricate marine life formations \\
               & an urban street where buildings are shaped as giant functional musical instruments including a violin apartment and piano mall \\
               & a butterfly with stained glass wings depicting medieval scenes catching sunlight \\
               & a floating city where traditional Chinese pavilions rest on clouds made of flowing silk fabric in pastel colors \\
               & a lion composed of moving gears and pistons that transforms between mechanical and organic forms \\
               & a garden where geometric crystal formations grow and branch like plants with rainbow refractions \\
               & a tree whose trunk is a twisting pagoda with branches of miniature traditional buildings and roof tile leaves \\
               & a phoenix-dragon hybrid creature covered in mirrored scales that create fractal reflections \\
               & a celestial teapot with constellation etchings pouring a stream of stars and nebulae \\
               & an origami landscape where paper mountains continuously fold and unfold to reveal geometric cities and rivers \\
               & a sphere where traditional Chinese ink and wash paintings flow continuously between day and night scenes \\
\midrule
Natural Creatures & a Velociraptor in hunting pose with detailed scales and feathers \\
                 & a Mammoth with detailed fur and tusks \\
                 & a chameleon changing colors with detailed scales \\
                 & a white tiger in mid-stride with flowing muscles \\
                 & a Pterodactyl with spread wings in flight pose \\
                 & an orangutan showing intelligent behavior \\
                 & a polar bear with detailed fur texture \\
\bottomrule
\end{tabular}
\caption{\textbf{Prompt format for 360° object rotation generation.} All prompts follow the same structural template, varying only in the subject description. The subjects are categorized into creative fusion designs that combine different artistic elements and concepts, and natural creatures that focus on realistic animal representations.}
\label{tab:prompts}
\end{table*}

\subsection{Extended Video Generation Capabilities}
\label{subsec:extended_video}

\textbf{Maximum Frame Count and Resolution:}
GRID demonstrates significant scalability in video generation tasks. Our framework supports up to 64 frames (2.7 seconds at 24 FPS) in a single generation pass, substantially outperforming mainstream methods like AnimateDiff V3 and VideoCrafter2 which are typically limited to 16 frames. For resolution flexibility, GRID supports single-frame resolutions of 512×768 for 16-frame generation and 256×384 for 64-frame generation.

\textbf{Comparison with Existing Methods:}
Table~\ref{tab:video_length_comparison} provides a comprehensive comparison of maximum frame count capabilities across different video generation models. While CogVideo supports up to 81 frames and Wonder 2.1 similarly supports 81 frames, our qualitative analysis reveals that frame quality and semantic consistency in these models often degrade significantly beyond 40-50 frames, particularly for complex scenes and motions.

\begin{table}[h]
\centering
\caption{Maximum single-pass frame count comparison across video generation models}
\label{tab:video_length_comparison}
\begin{tabular}{lcc}
\toprule
\textbf{Model} & \textbf{Max Frames} & \textbf{Default Resolution} \\
\midrule
AnimateDiff V3 & 16 & 256×384 \\
VideoCrafter2 & 16 & 256×384 \\
GRID (Ours) & 64 & 256×384 \\
CogVideo & 81 & 720×480 \\
Wonder 2.1 & 81 & 512×512 \\
\bottomrule
\end{tabular}
\end{table}

\textbf{Zero-shot Extension Capability:}
A unique advantage of GRID is its inherent zero-shot extension capability. As demonstrated in Figure 11 of our main paper, models trained on 4×4 grid formats can directly extend to 4×8 formats without any additional training. This enables extending videos to 128 frames (5 seconds) while maintaining both semantic consistency and visual quality.

Unlike traditional diffusion methods that suffer from semantic drift during extension, our approach demonstrates:
\begin{itemize}
\item Superior temporal consistency across extended sequences
\item Better frame-to-frame coherence in action and motion
\item Maintained visual quality and semantic alignment throughout extended videos
\end{itemize}

Furthermore, this capability can be combined with our sequence splicing strategy to generate even longer videos if needed for specific applications. It's worth noting that for all existing methods, generating more than 81 frames while maintaining high quality and semantic consistency is a significant challenge. Claims of hundred-frame capabilities in some models often rely on additional frame interpolation or splicing techniques, which should be distinguished from native single-model capabilities.

\textbf{Resource Efficiency Scaling:}
An important aspect of GRID's design is its exceptional resource efficiency while maintaining quality. Compared to leading models, GRID achieves:
\begin{itemize}
\item 6-67× faster inference acceleration
\item Approximately 1000× reduction in training data requirements (35K videos vs. 35M videos)
\item 90-97
\end{itemize}

This efficiency advantage becomes even more pronounced when generating longer videos, making GRID particularly suitable for resource-constrained environments or real-time applications.

\subsection{Additional Benchmark Evaluations}
\label{subsec:additional_benchmarks}

\textbf{T2VCompBench Evaluation:}
To comprehensively evaluate our method's capabilities in video generation, we conducted evaluations on T2VCompBench~\cite{sun2024t2v}, which measures various dimensions of video generation quality. Table~\ref{tab:t2vcompbench} presents the results.

\begin{table}[h]
\centering
\caption{Evaluation results on T2VCompBench benchmark}
\label{tab:t2vcompbench}
\resizebox{\textwidth}{!}{
\begin{tabular}{lcccccccccc}
\toprule
\textbf{Model} & \textbf{Consist-attr$\uparrow$} & \textbf{Dynamic-attr$\uparrow$} & \textbf{Spatial$\uparrow$} & \textbf{Motion$\uparrow$} & \textbf{Action$\uparrow$} & \textbf{Interaction$\uparrow$} & \textbf{Numeracy$\uparrow$} & \textbf{Time(s)$\downarrow$} & \textbf{Data$\downarrow$} \\
\midrule
OpenSora 1.2 & 0.5639 & 0.0189 & 0.5063 & 0.2468 & 0.4833 & 0.5039 & 0.3719 & 12.0 & $>$35M \\
CogVideoX-5B & 0.6164 & \textbf{0.0219} & \textbf{0.5172} & \textbf{0.2658} & 0.5333 & \textbf{0.6069} & \textbf{0.3706} & 17.8 & 35M \\
GRID (Stage2) & 0.5669 & 0.0098 & 0.4711 & 0.1987 & 0.5103 & 0.4512 & 0.3579 & 7.2 & 35K \\
GRID (Ours) & \textbf{0.6132} & 0.0114 & 0.4918 & 0.2175 & \textbf{0.5342} & 0.4829 & 0.3686 & \textbf{7.2} & \textbf{35K} \\
\bottomrule
\end{tabular}
}
\end{table}

Our GRID model achieves comparable or even superior performance to specialized large models in attribute consistency (0.6132) and action expression (0.5342). The significant improvement from Stage2 (0.5669) to Full model (0.6132) in Consist-attr score demonstrates the effectiveness of our temporal loss in enhancing inter-frame consistency.

Furthermore, GRID demonstrates remarkable efficiency advantages:
\begin{itemize}
\item Computational efficiency: 1.7× faster inference than OpenSora and 2.5× faster than CogVideoX
\item Parameter efficiency: 91\% fewer parameters than OpenSora and 97\% fewer than CogVideoX
\item Data efficiency: Training data requirements are 1000× less than existing methods
\end{itemize}

\textbf{Inter-frame Consistency Evaluation:}
To specifically measure the improvement in inter-frame consistency provided by our temporal loss function, we conducted a detailed comparison between our Stage2 and Full models. As shown in Table~\ref{tab:consistency}, the addition of our temporal loss ($\mathcal{L}_{flow}$) significantly improves consistency measures.

\begin{table}[h]
\centering
\caption{Inter-frame consistency comparison between model variants}
\label{tab:consistency}
\begin{tabular}{lcc}
\toprule
\textbf{Method} & \textbf{Consist-attr$\uparrow$} & \textbf{Flow Consistency Score$\uparrow$} \\
\midrule
Ours (Stage2) & 0.5669 & 0.7214 \\
Ours (Full) & \textbf{0.6132} & \textbf{0.8567} \\
\bottomrule
\end{tabular}
\end{table}

The significant improvement in both Consist-attr (from T2VCompBench) and our Flow Consistency Score demonstrates that our temporal loss effectively enhances the model's ability to maintain consistent object attributes and motion patterns across frames, addressing a key limitation of existing methods.

\subsection{Potential Applications}
\label{appendix:pa}
Our framework demonstrates significant potential beyond its primary applications.

\subsubsection{Creative Multi-view Generation}
As shown in Figure~\ref{fig:creative}, our method exhibits remarkable flexibility in combining different conceptual elements to create novel multi-view compositions. The grid-based layout allows for intuitive arrangement and manipulation of various visual elements, enabling creative expressions that would be challenging for traditional approaches. This capability suggests promising applications in creative design, artistic visualization, and content creation.

\input{figures/layout/tex/multy}

\input{figures/layout/tex/man}
\input{figures/layout/tex/cat}

\input{figures/layout/tex/creative}

\subsubsection{Flexible Frame Extension}
Notably, our model demonstrates strong generalization capability in sequence length. Despite being trained on 4×4 (16-frame) driving scenarios, the model can effectively generate 4×8 (32-frame) sequences by simply adjusting the $c_layout$ prompt at inference time. As shown in Figure~\ref{fig:inf4-8}, the extended sequences maintain temporal consistency and visual quality comparable to the original training length. This flexibility suggests that our layout-based approach naturally accommodates variable-length generation without requiring explicit retraining, opening possibilities for dynamic content generation across different temporal scales.

\input{figures/layout/tex/inf4-8}

\input{figures/layout/tex/resto}

\subsubsection{Future Extension to Video Understanding}
Our layout-based framework shows potential in transforming traditional video understanding tasks into image-domain problems. Unlike conventional autoregressive approaches~\cite{bai20243d} that process frames sequentially, our method arranges frames in a grid layout, enabling parallel processing and global temporal modeling. This approach could benefit various video understanding tasks: for video-text retrieval, the layout representation allows direct comparison between video content and text embeddings across all frames simultaneously; for video question answering, it enables the model to attend to relevant frames across the entire sequence without sequential constraints; for video tracking and other analysis tasks, it avoids error accumulation common in traditional sequential processing. While we have not conducted specific experiments in these directions, our framework's ability to convert temporal relationships into spatial ones through layouts offers a promising alternative to conventional video understanding paradigms, potentially enabling more efficient and effective multi-modal video analysis.

\subsubsection{Maintained Image Generation Ability}
\label{appendix:ri}
\begin{figure}[t]
    \begin{subfigure}{\linewidth}
        \centering
        \includegraphics[width=0.24\linewidth]{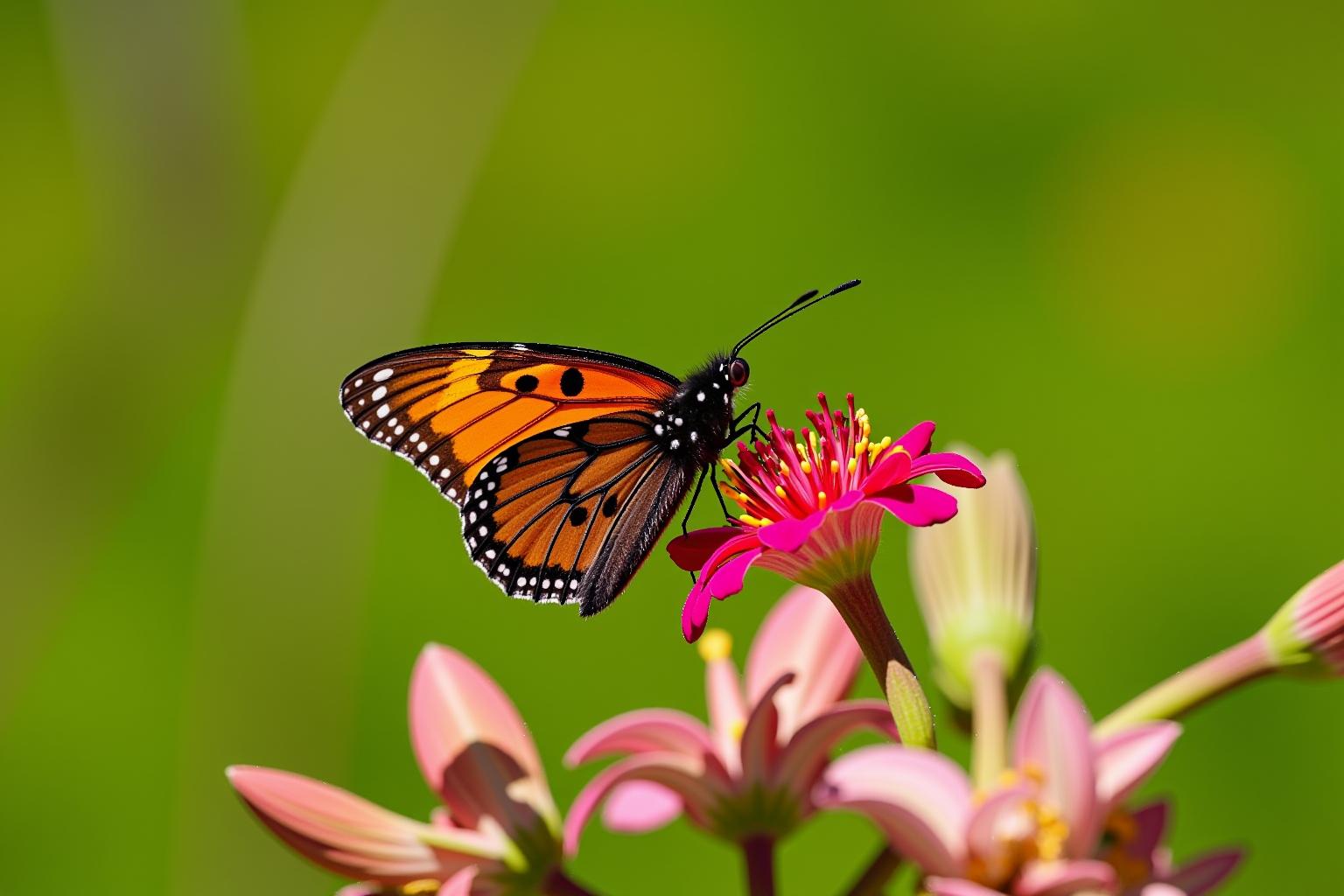}
        \hfill
        \includegraphics[width=0.24\linewidth]{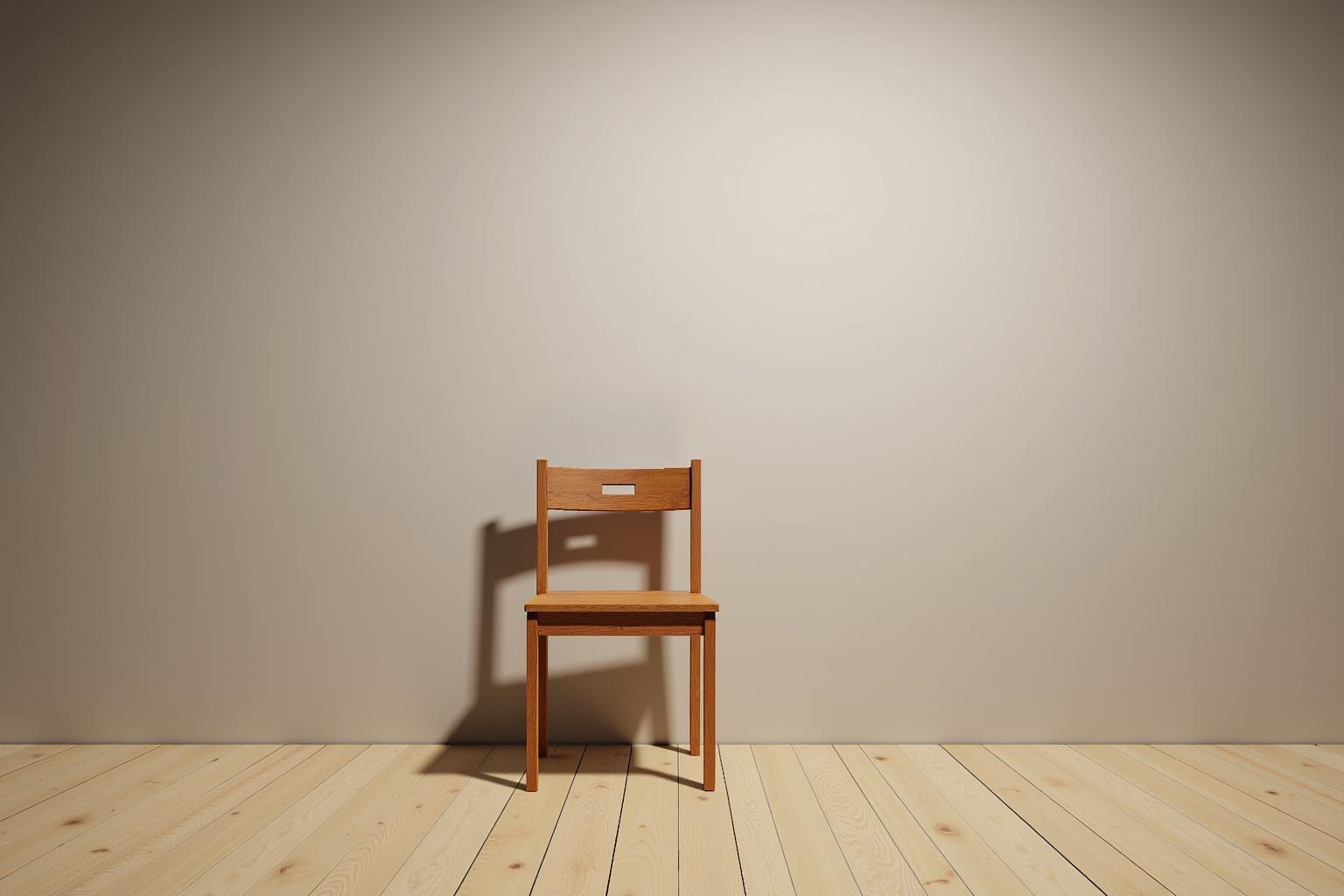}
        \hfill
        \includegraphics[width=0.24\linewidth]{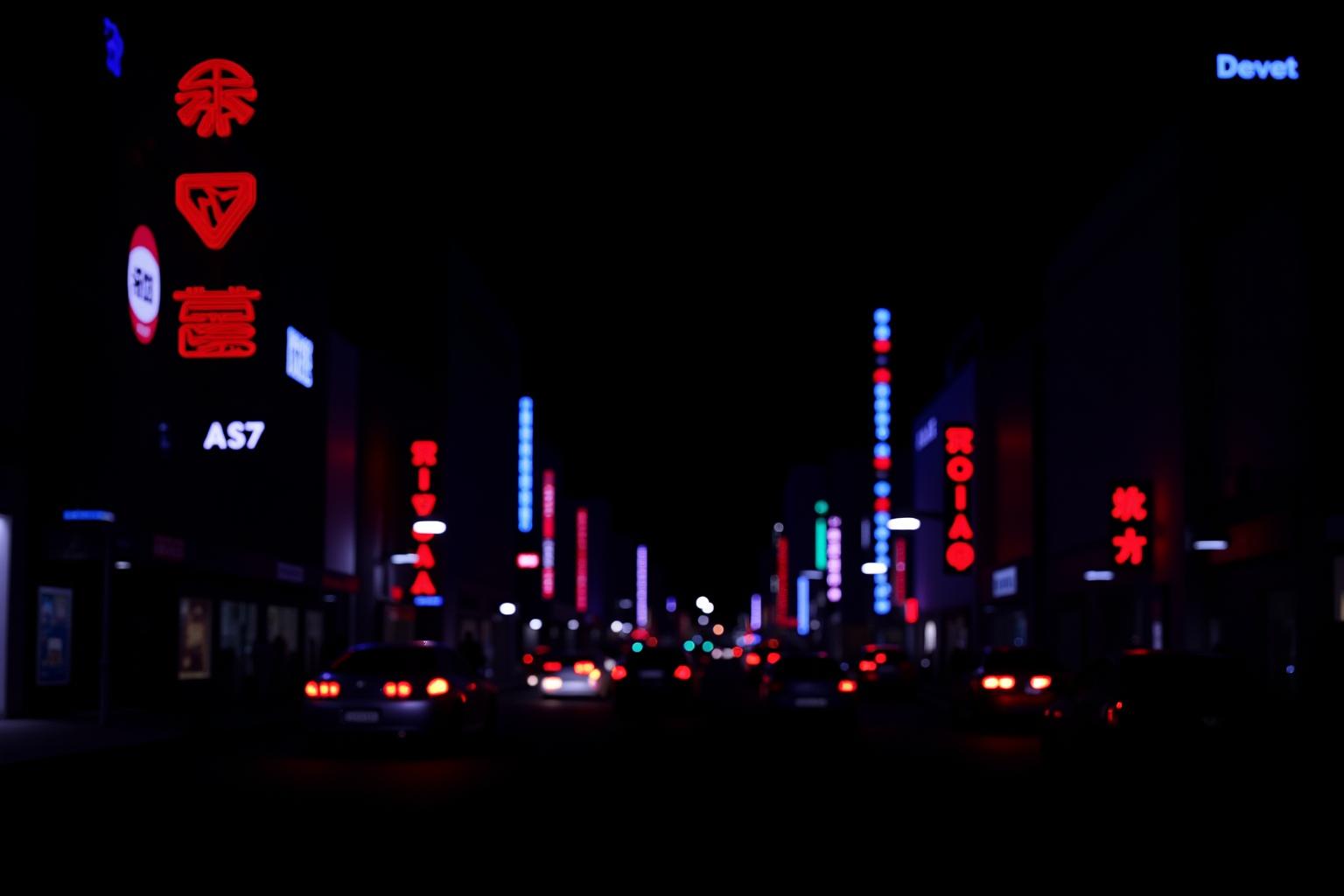}
        \hfill
        \includegraphics[width=0.24\linewidth]{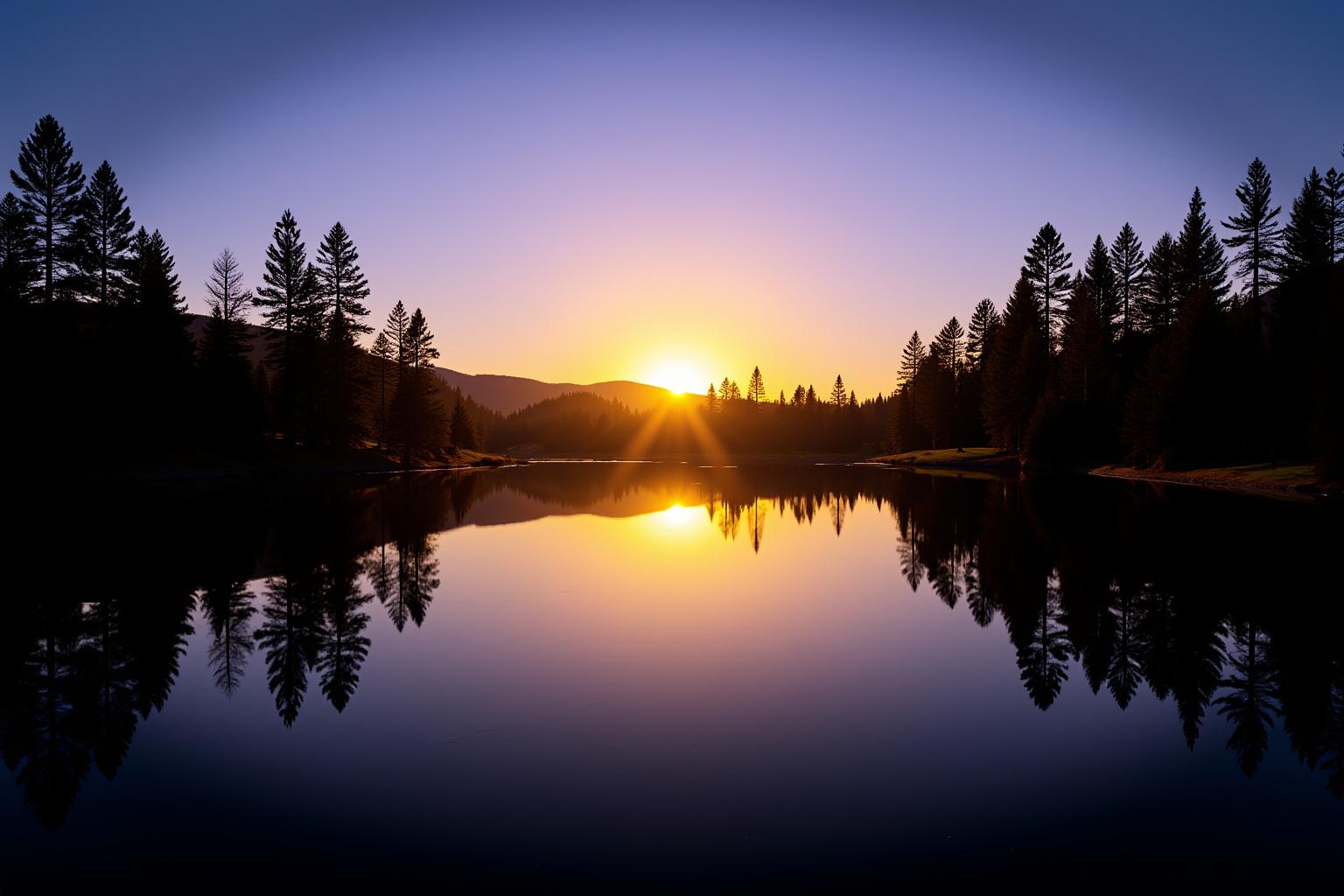}
    \end{subfigure}
    
    \begin{subfigure}{\linewidth}
        \centering
        \includegraphics[width=0.24\linewidth]{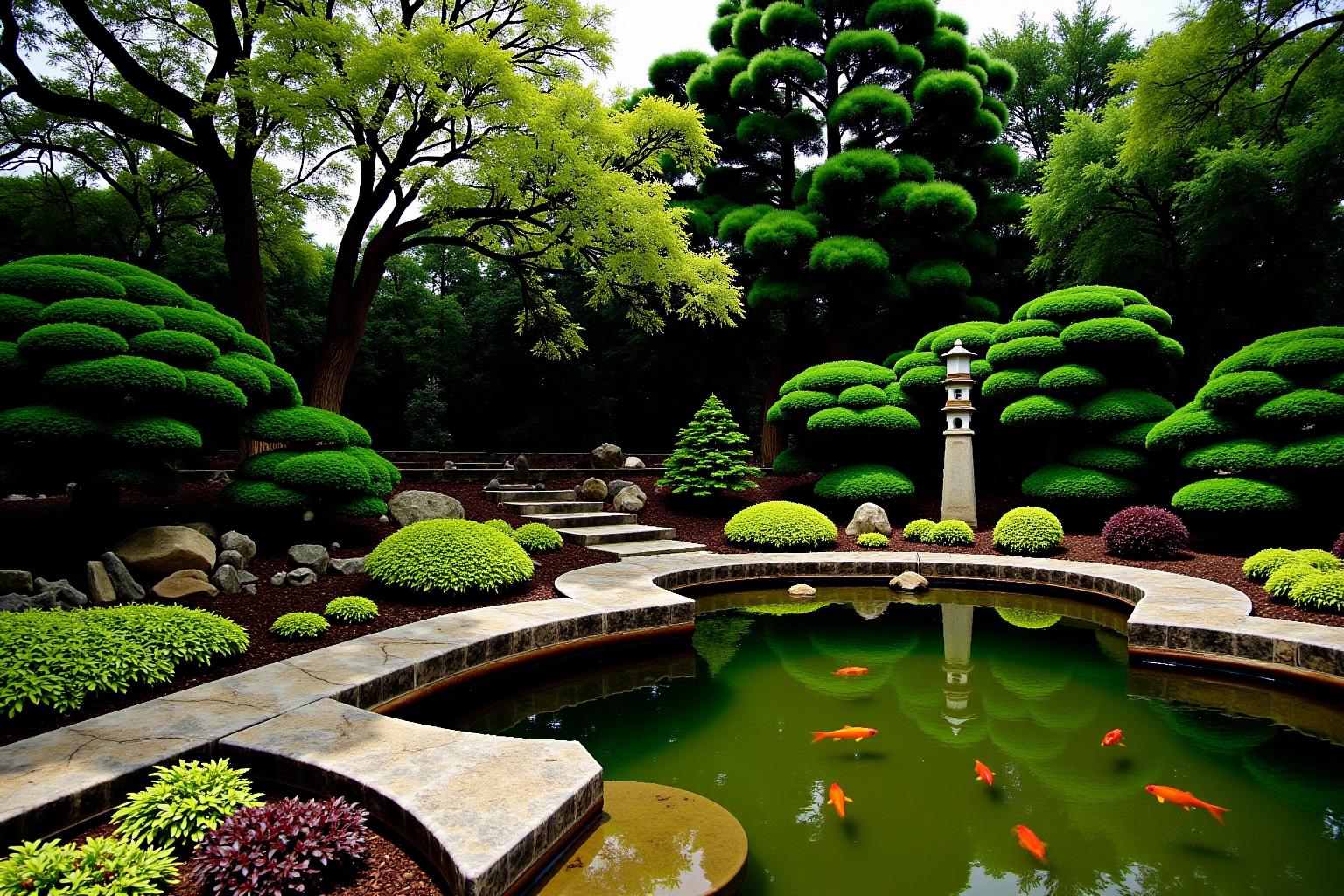}
        \hfill
        \includegraphics[width=0.24\linewidth]{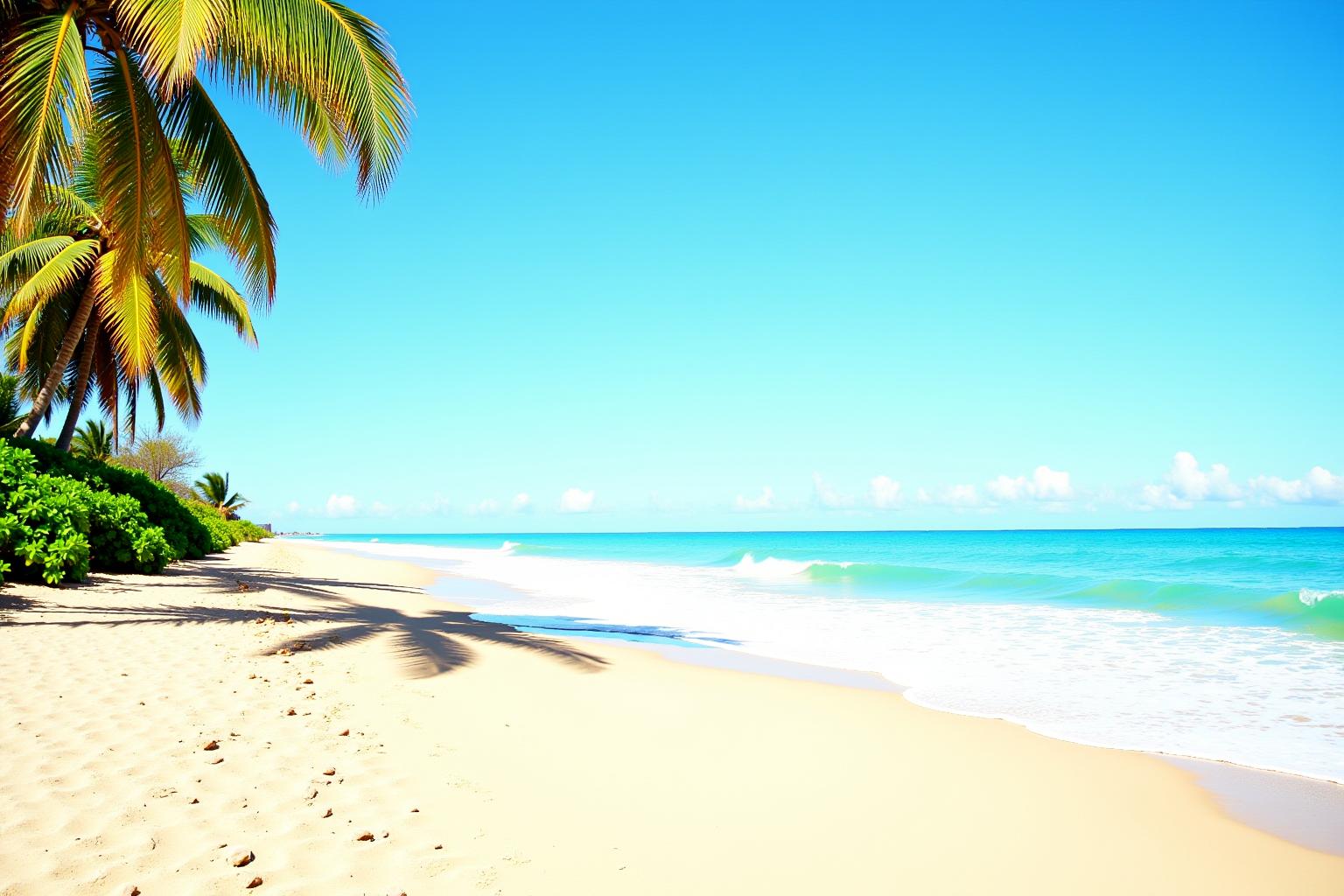}
        \hfill
        \includegraphics[width=0.24\linewidth]{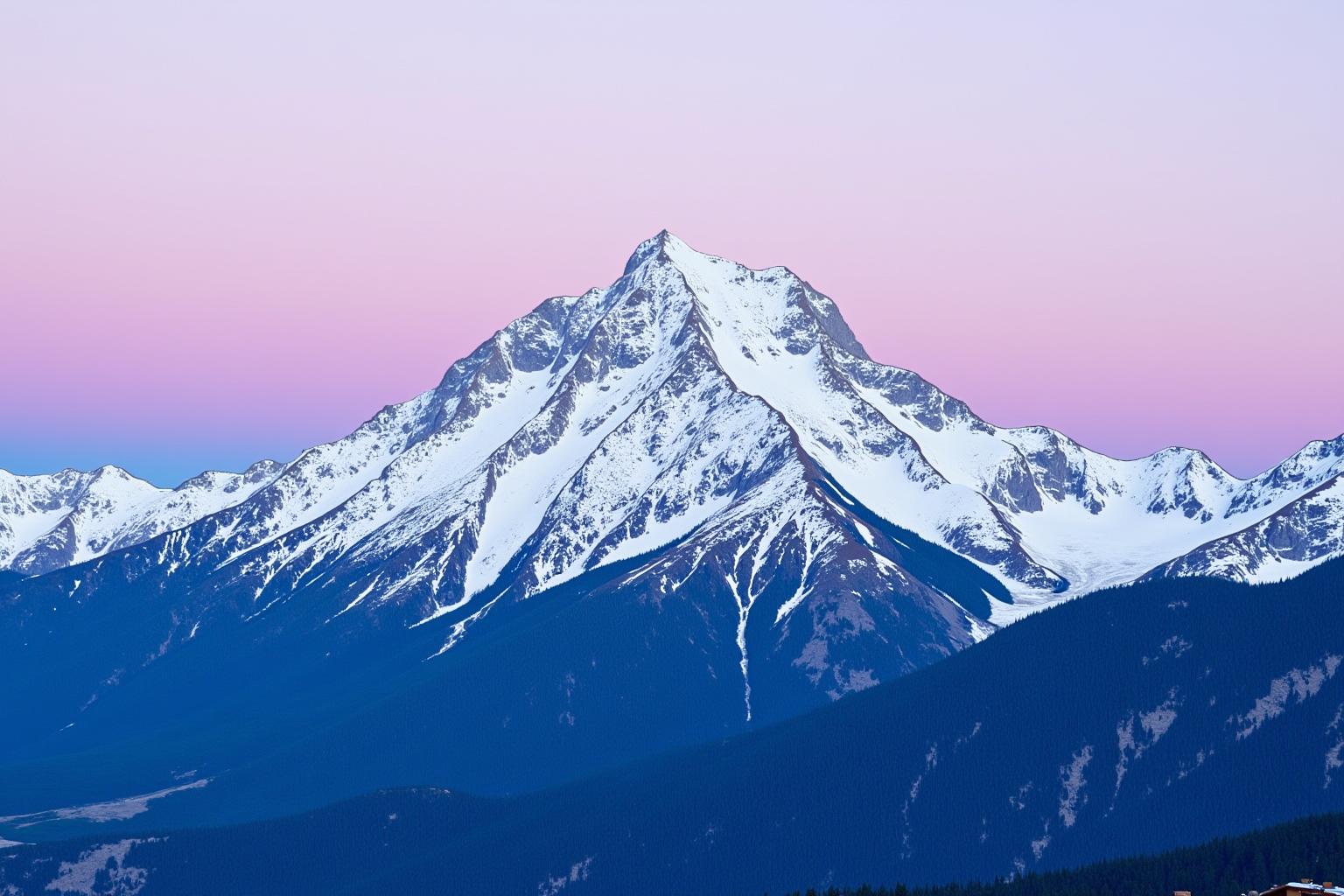}
        \hfill
        \includegraphics[width=0.24\linewidth]{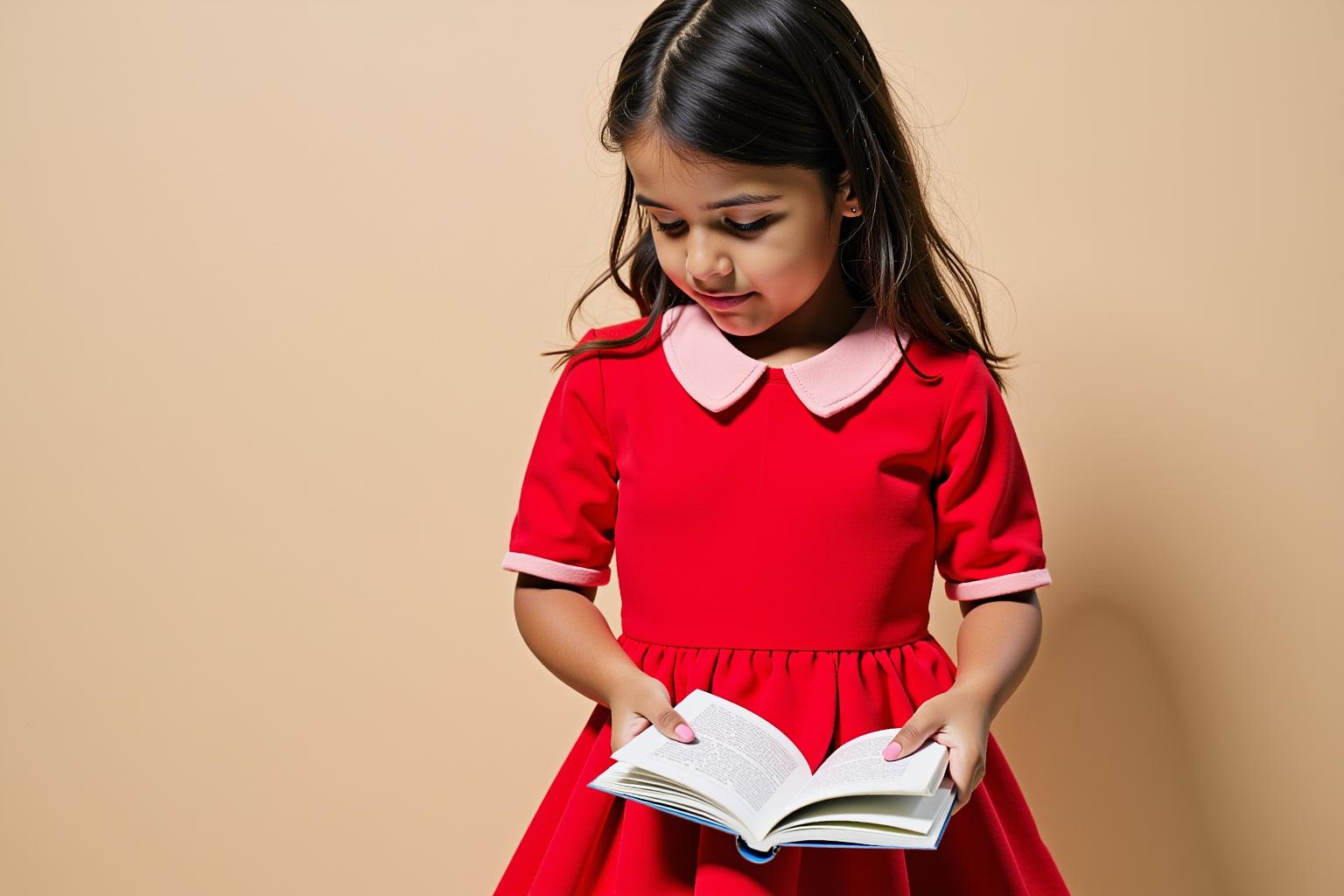}
    \end{subfigure}

    \begin{subfigure}{\linewidth}
        \centering
        \includegraphics[width=0.24\linewidth]{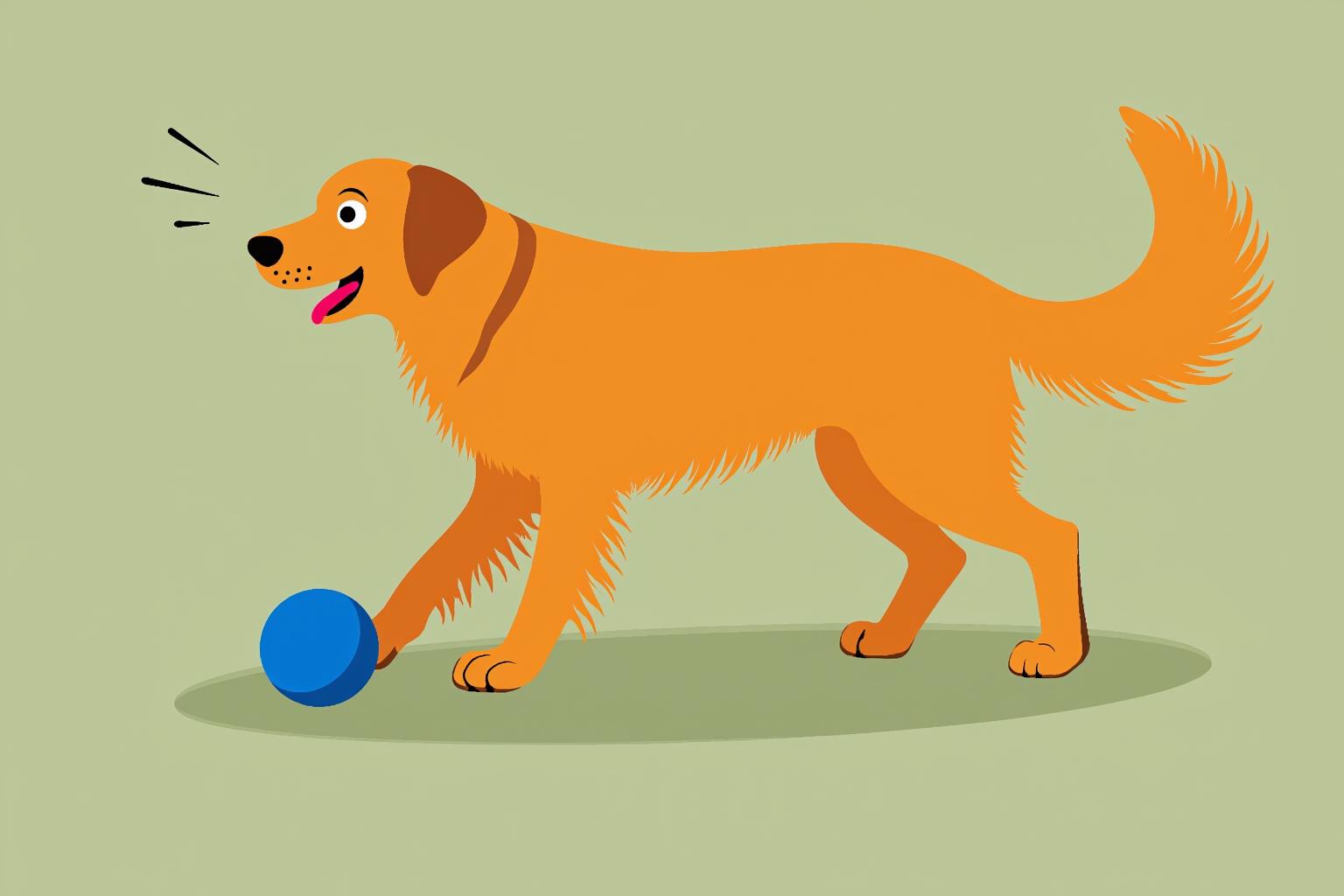}
        \hfill
        \includegraphics[width=0.24\linewidth]{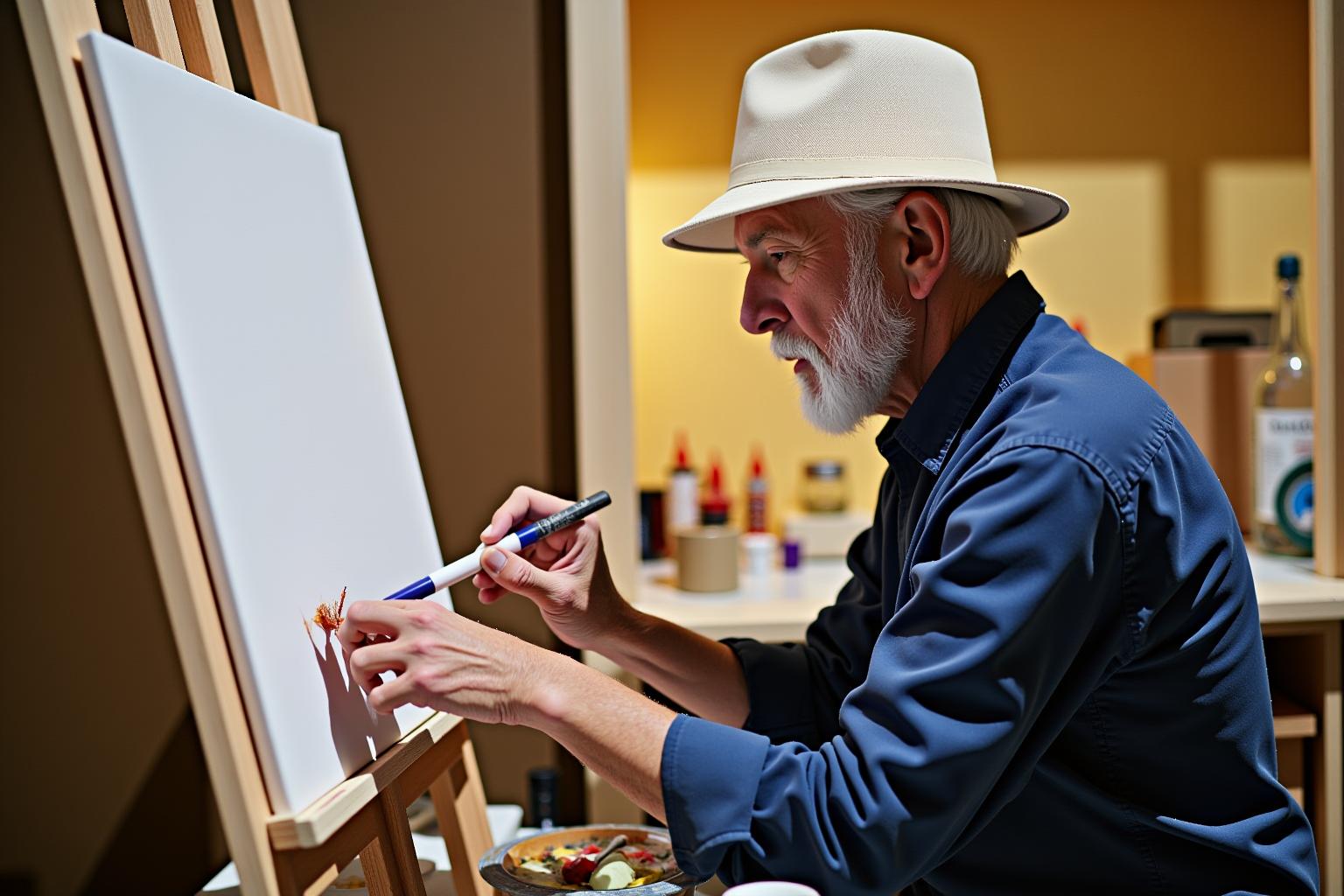}
        \hfill
        \includegraphics[width=0.24\linewidth]{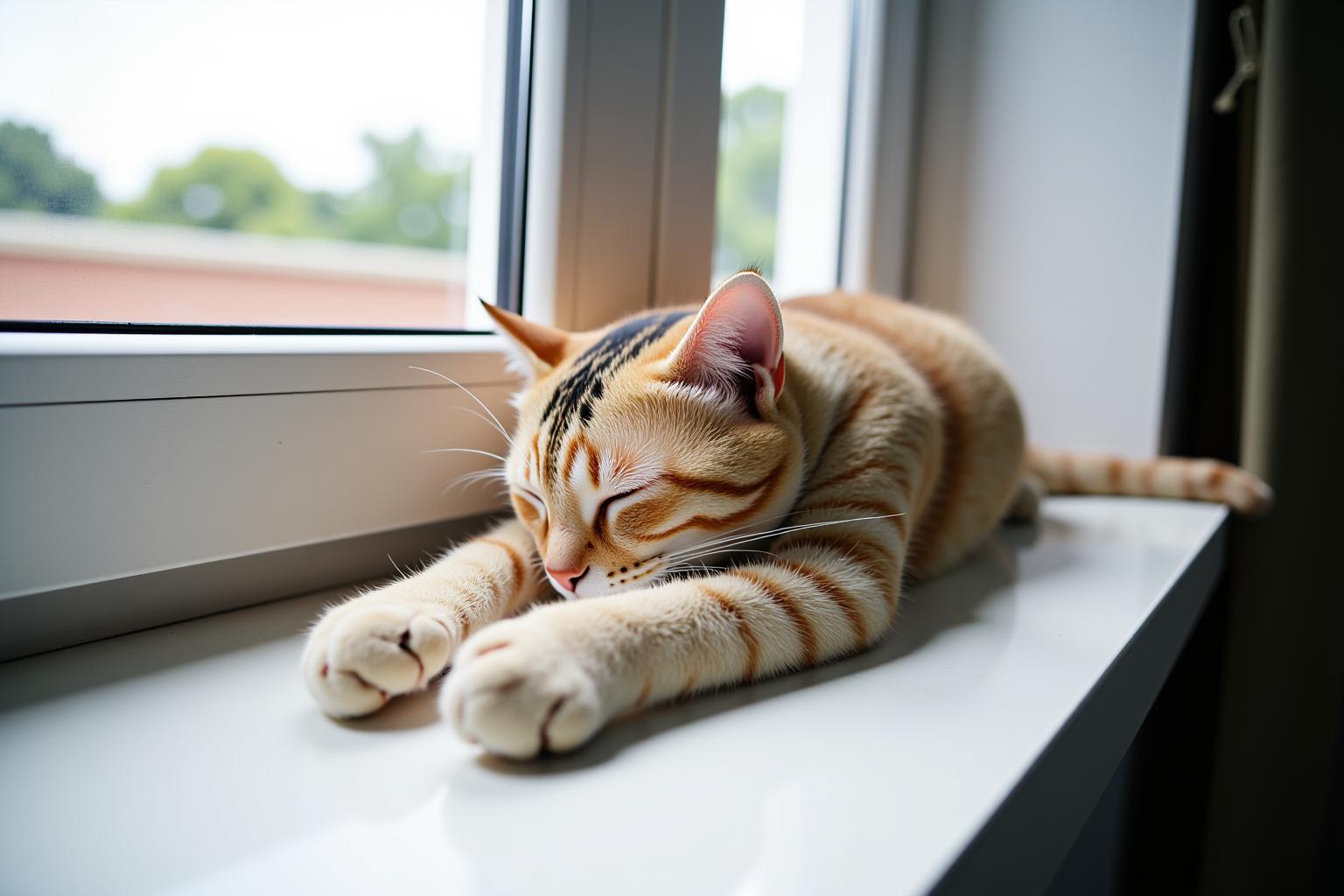}
        \hfill
        \includegraphics[width=0.24\linewidth]{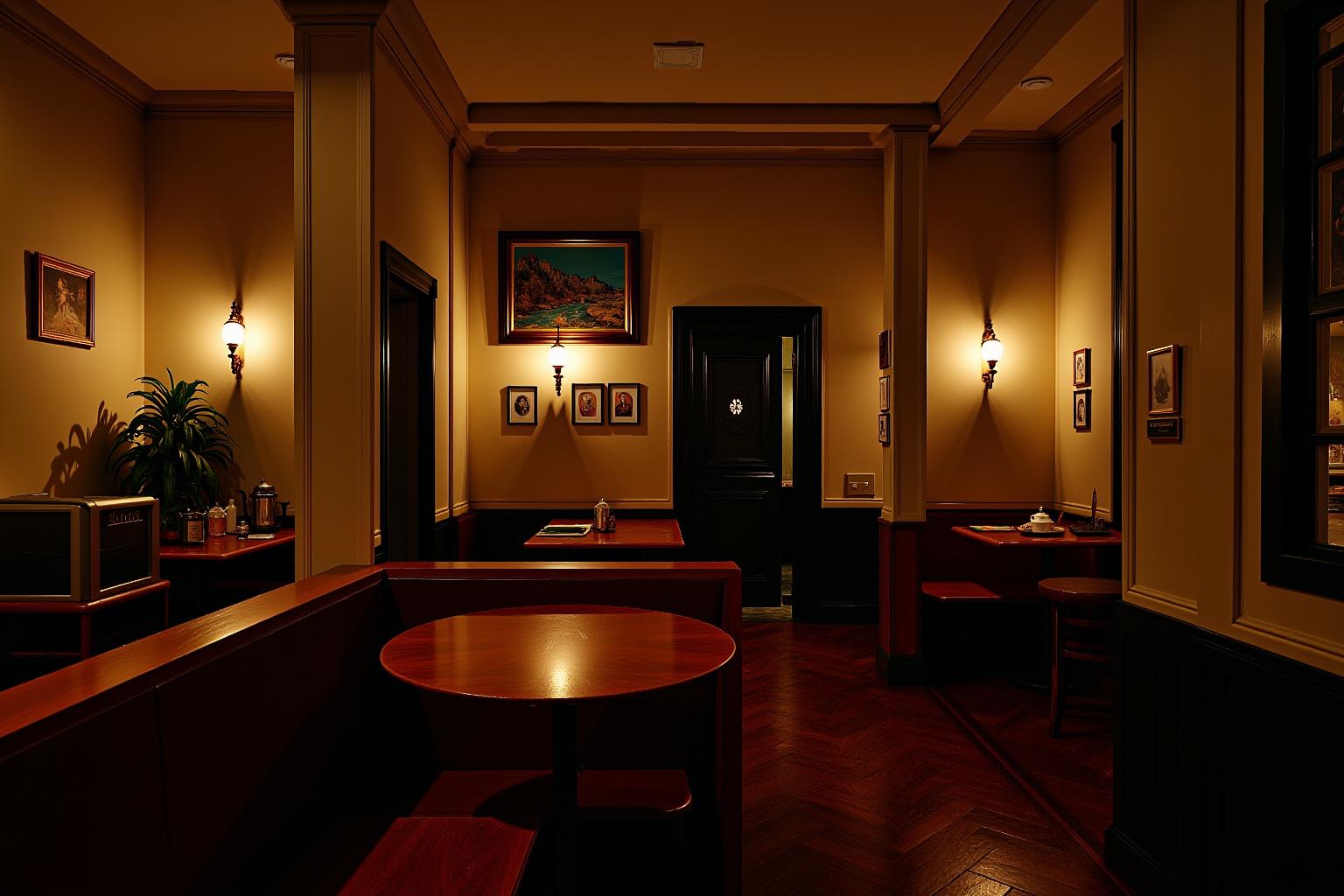}
    \end{subfigure}

    \caption{Demonstration of maintained image generation capabilities. Our model preserves high-quality single-image generation performance across diverse scenarios including: basic objects, nature scenes, character interactions, indoor/outdoor environments, artistic styles, and lighting effects. Each image is generated from text prompts testing different aspects of the model's generation abilities.}
    \label{fig:original}
\end{figure}

Our framework preserves the original Flux model's image generation capabilities while extending its functionality to handle video sequences.
As demonstrated in Figure~\ref{fig:original}, the model maintains high-quality performance on various image generation tasks such as text-to-image synthesis, image editing, and style transfer. 
This preservation of original capabilities alongside newly acquired video generation abilities creates a versatile model that can seamlessly handle both single-image and multi-frame tasks. The ability to maintain original image generation quality while adding new functionality demonstrates the effectiveness of our training approach and the robustness of the layout-based framework.

\subsection{Impact Statement}
This paper introduces research aimed at advancing visual sequence generation through an efficient layout-based framework. However, we must emphasize the potential risks associated with this technology~\cite{anydressing}, particularly in facial manipulation applications~\cite{spf, luo2024codeswap}, where our method could be misused to compromise identity security. Nevertheless, recent advances in adversarial perturbation protection mechanisms~\cite{wan2024prompt} provide solutions to help users protect their personal data against unauthorized model fine-tuning and malicious content generation. Therefore, we call for attention to these risks and encourage the adoption of defensive techniques to ensure the protection of personal content while advancing the development of generative AI technologies.

\subsection{Limitations}
Our approach faces two primary limitations. First, the grid-based layout design inherently constrains frame resolution due to limitations of the based Text-to-Image models when processing multiple frames simultaneously. Second, our training strategy, based on lora finetuning, shows limitations in text-to-video generation tasks that significantly deviate from the base model's capabilities. Combined with our relatively small training dataset, this makes it challenging to achieve competitive performance in open-world video generation scenarios requiring complex motion understanding.

\subsection{Multyview Camera Parameters}
\label{appendix:camera}

Building upon the dataset opensourced by Diffusion4D~\cite{diffusion4d}, Table~\ref{tab:camera} presents camera trajectory parameters, which serve as the foundation for consistent 4D content generation and subsequent reconstruction tasks.

Our camera configuration follows precise mathematical relationships, with cameras positioned at 15-degree intervals along a circle of radius 2 units in the horizontal plane. The systematic progression of coordinate bases ensures optimal coverage while maintaining consistent inter-frame relationships. Each camera's orientation is defined by orthogonal basis vectors, with the Y vector consistently aligned with the negative Z-axis to establish stable up-direction reference.

\begin{table*}[tp]
\centering

\begin{tabular}{|c|c|c|c|c|}
\hline
\textbf{Frame} & \textbf{X Vector} & \textbf{Y Vector} & \textbf{Z Vector} & \textbf{Origin} \\
\hline
1 & [1.0, 0.0, 0.0] & [-0.0, 0.0, -1.0] & [-0.0, 1.0, 0.0] & [0.0, -2.0, 0.0] \\
2 & [0.96, 0.27, -0.0] & [0.0, -0.0, -1.0] & [-0.27, 0.96, -0.0] & [0.54, -1.93, 0.0] \\
3 & [-0.92, 0.4, -0.0] & [0.0, 0.0, -1.0] & [-0.4, -0.92, -0.0] & [0.8, 1.83, 0.0] \\
4 & [-0.99, 0.14, -0.0] & [0.0, 0.0, -1.0] & [-0.14, -0.99, -0.0] & [0.27, 1.98, 0.0] \\
5 & [-0.99, -0.14, 0.0] & [-0.0, 0.0, -1.0] & [0.14, -0.99, -0.0] & [-0.27, 1.98, 0.0] \\
6 & [-0.92, -0.4, 0.0] & [-0.0, 0.0, -1.0] & [0.4, -0.92, -0.0] & [-0.8, 1.83, 0.0] \\
7 & [-0.78, -0.63, 0.0] & [-0.0, -0.0, -1.0] & [0.63, -0.78, 0.0] & [-1.26, 1.55, 0.0] \\
8 & [-0.58, -0.82, -0.0] & [0.0, 0.0, -1.0] & [0.82, -0.58, 0.0] & [-1.63, 1.15, 0.0] \\
9 & [-0.33, -0.94, -0.0] & [0.0, -0.0, -1.0] & [0.94, -0.33, 0.0] & [-1.88, 0.67, 0.0] \\
10 & [-0.07, -1.0, -0.0] & [0.0, 0.0, -1.0] & [1.0, -0.07, 0.0] & [-2.0, 0.14, 0.0] \\
11 & [0.2, -0.98, 0.0] & [0.0, -0.0, -1.0] & [0.98, 0.2, 0.0] & [-1.96, -0.41, 0.0] \\
12 & [0.46, -0.89, 0.0] & [0.0, -0.0, -1.0] & [0.89, 0.46, 0.0] & [-1.78, -0.92, 0.0] \\
13 & [0.85, 0.52, 0.0] & [-0.0, 0.0, -1.0] & [-0.52, 0.85, 0.0] & [1.04, -1.71, 0.0] \\
14 & [0.68, -0.73, -0.0] & [-0.0, 0.0, -1.0] & [0.73, 0.68, 0.0] & [-1.46, -1.37, 0.0] \\
15 & [0.85, -0.52, -0.0] & [0.0, 0.0, -1.0] & [0.52, 0.85, 0.0] & [-1.04, -1.71, 0.0] \\
16 & [0.96, -0.27, 0.0] & [-0.0, -0.0, -1.0] & [0.27, 0.96, -0.0] & [-0.54, -1.93, 0.0] \\
17 & [1.0, -0.0, 0.0] & [0.0, 0.0, -1.0] & [0.0, 1.0, 0.0] & [-0.0, -2.0, 0.0] \\
18 & [0.68, 0.73, 0.0] & [0.0, 0.0, -1.0] & [-0.73, 0.68, 0.0] & [1.46, -1.37, 0.0] \\
19 & [0.46, 0.89, -0.0] & [-0.0, -0.0, -1.0] & [-0.89, 0.46, 0.0] & [1.78, -0.92, 0.0] \\
20 & [0.2, 0.98, -0.0] & [-0.0, -0.0, -1.0] & [-0.98, 0.2, 0.0] & [1.96, -0.41, 0.0] \\
21 & [-0.07, 1.0, 0.0] & [-0.0, 0.0, -1.0] & [-1.0, -0.07, 0.0] & [2.0, 0.14, 0.0] \\
22 & [-0.33, 0.94, 0.0] & [-0.0, -0.0, -1.0] & [-0.94, -0.33, 0.0] & [1.88, 0.67, 0.0] \\
23 & [-0.58, 0.82, 0.0] & [-0.0, 0.0, -1.0] & [-0.82, -0.58, 0.0] & [1.63, 1.15, 0.0] \\
24 & [-0.78, 0.63, -0.0] & [0.0, -0.0, -1.0] & [-0.63, -0.78, 0.0] & [1.26, 1.55, 0.0] \\
\hline
\end{tabular}
\caption{Camera Parameters for 24 Frames}
\label{tab:camera}
\end{table*}

%% file: figures/layout_fig_zero_eval.tex
\begin{figure*}[h]
    \centering
    \includegraphics[width=\textwidth]{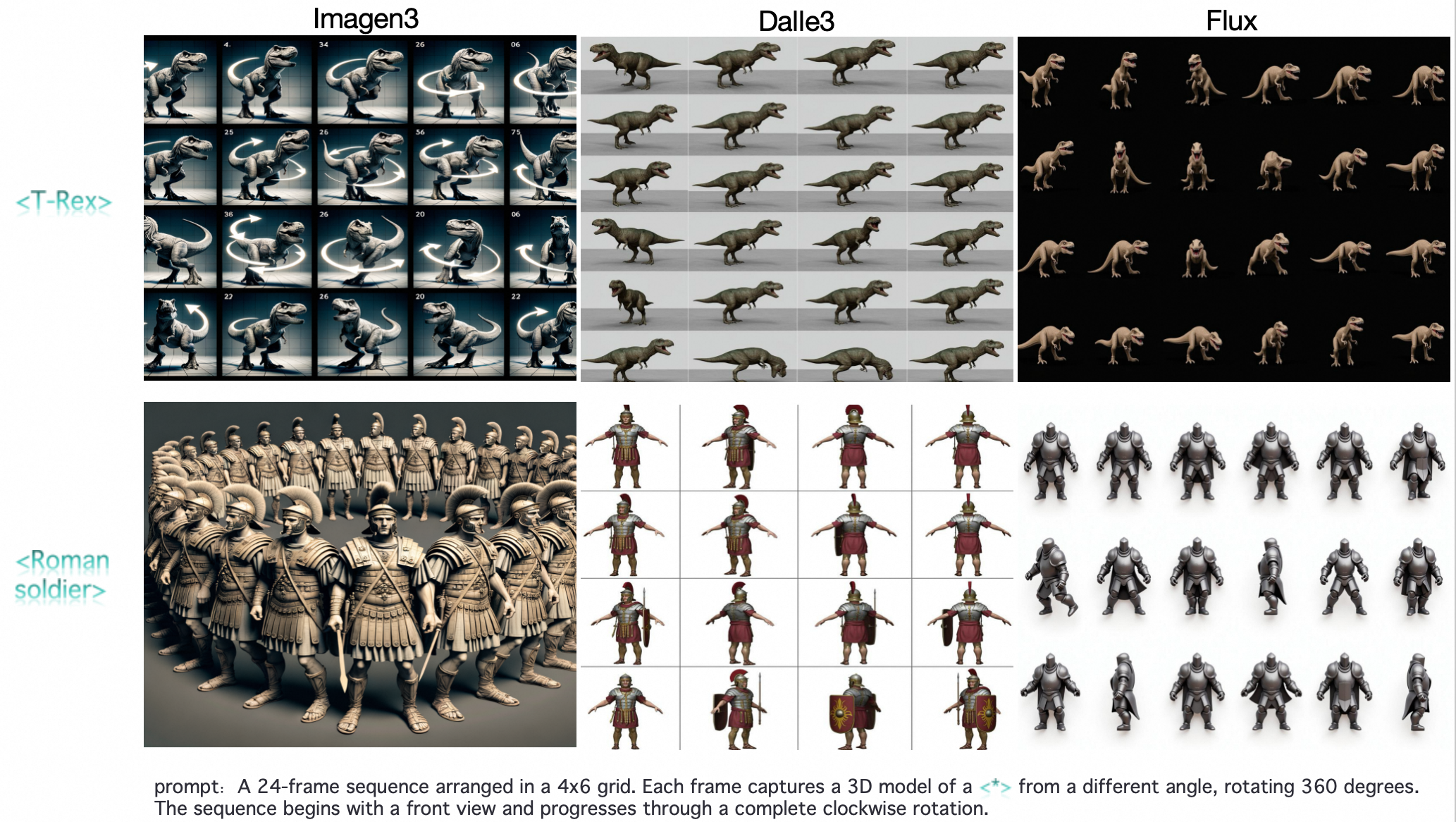}
    \caption{Zero-shot evaluation of foundation models on grid-based multi-view generation tasks before we begin to train. Using the prompt "a * from different angles in a mxn grid layout,"}
    \label{fig:zero_eval}
\end{figure*}

%% file: figures/layout/tex/attention.tex
\begin{figure*}[h]
    \centering
    \includegraphics[width=\textwidth]{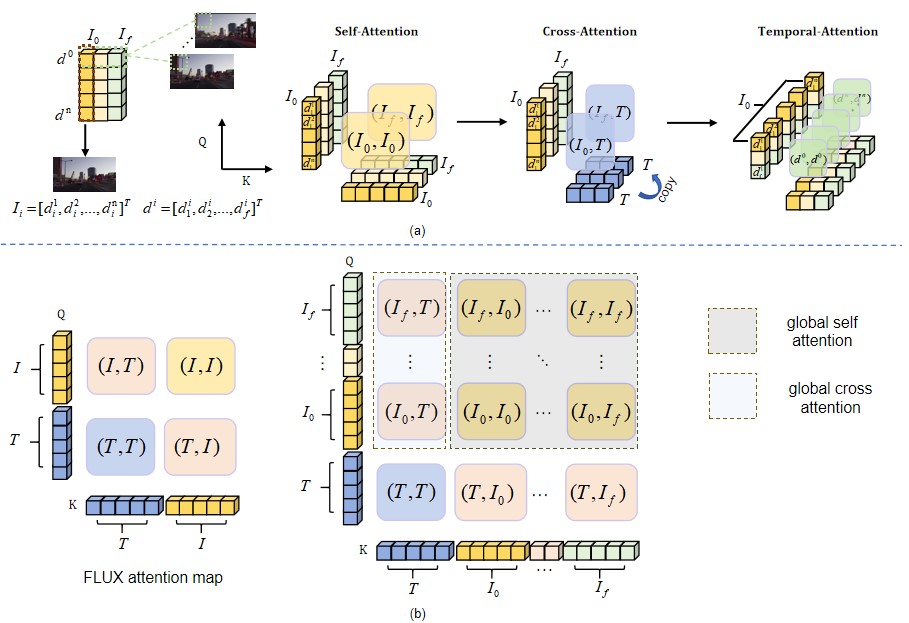}
    \caption{Comparison of attention mechanisms. (a) Traditional video diffusion models rely on three separate attention modules to handle spatial understanding, semantic guidance, and temporal consistency respectively. (b) Through our grid layout reformulation, FLUX's unified self-attention naturally encompasses both inner-frame ($I_i,I_i$) and cross-frame ($I_i,I_j$) relationships, while its global text-image attention ($T,I$) enables consistent control across all frames. This simplification eliminates the need for specialized temporal modules while maintaining effective spatio-temporal understanding.}
    \label{fig:attention}
\end{figure*}

%% file: figures/layout/tex/multy.tex
\begin{figure}[t]
    \centering
    \includegraphics[width=\linewidth]{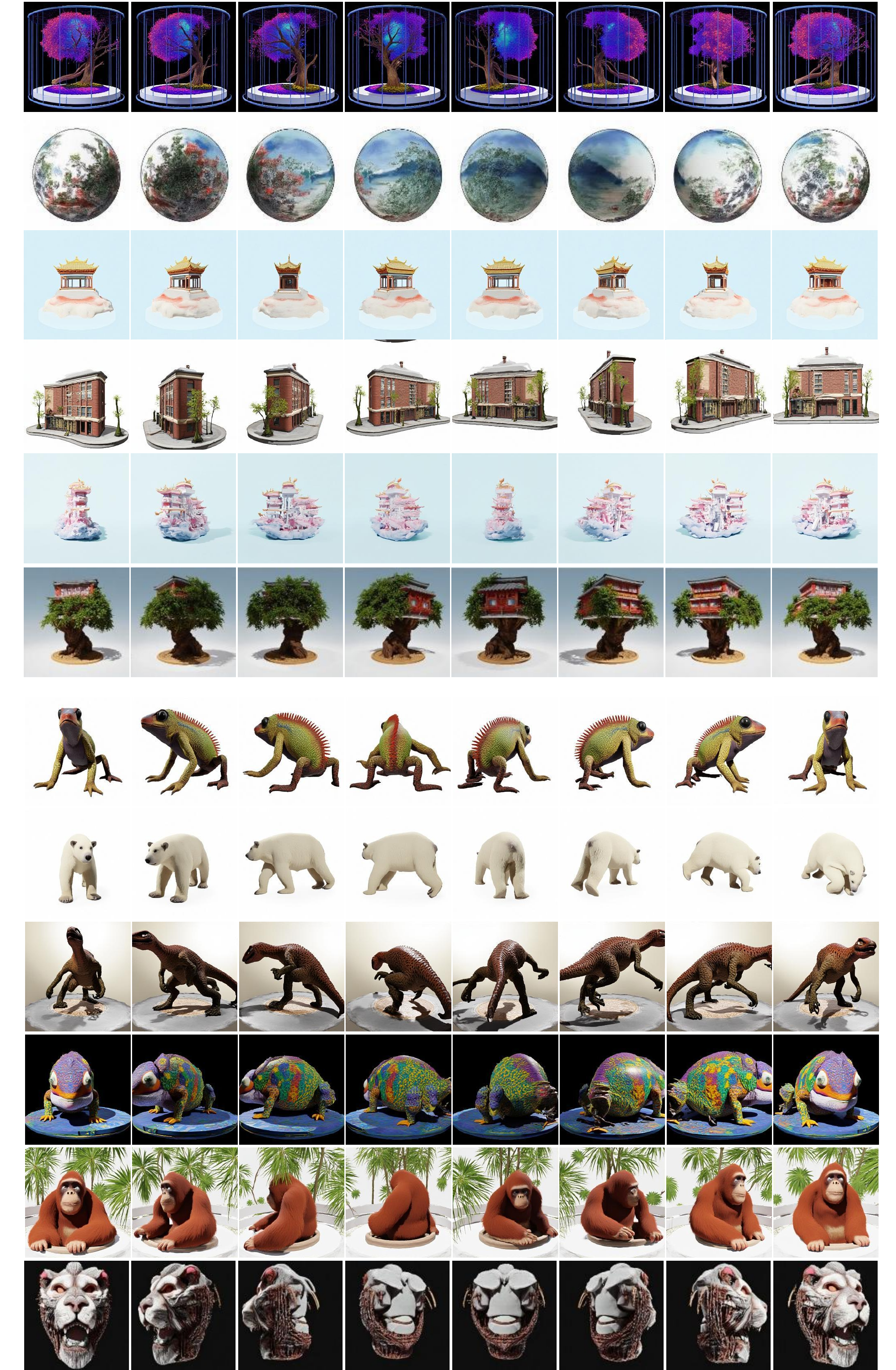}
\caption{Multi-view generation results for static objects (top six rows) and dynamic subjects (bottom six rows), demonstrating consistent appearance and structure across different viewpoints.}
\label{fig:4D}
\vspace{-4mm}

\end{figure}


%% file: figures/layout/tex/man.tex
\begin{figure}[t]
    \centering
    \includegraphics[width=\linewidth]{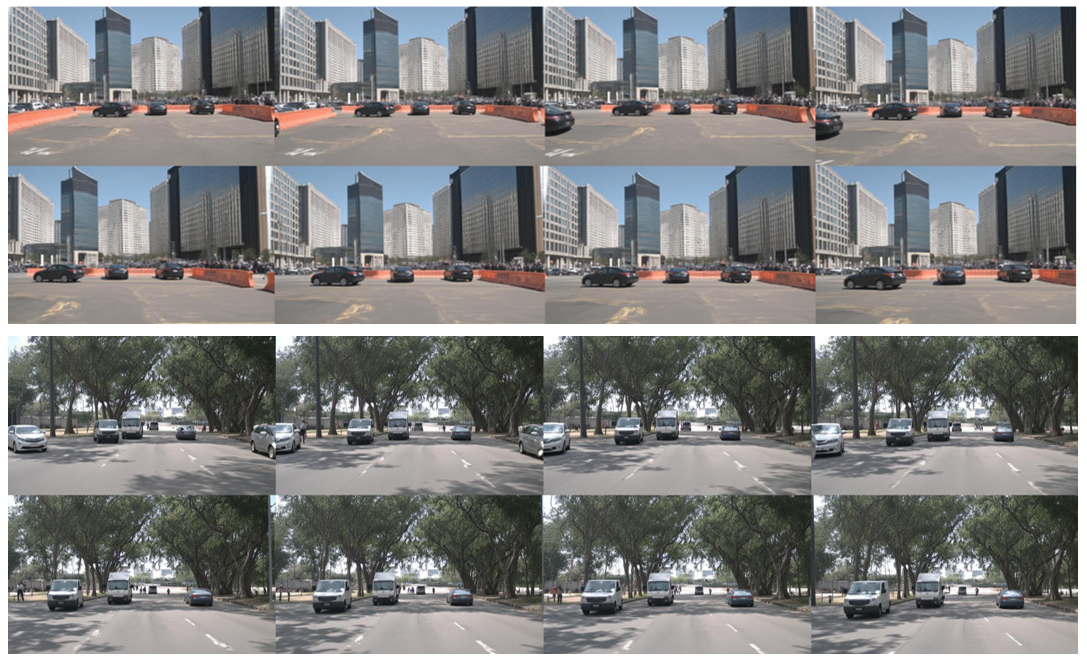}
\caption{Text-to-Video Generation of driving scenes, showcasing complex multi-vehicle scenarios which represent the most challenging aspects of driving scene generation.}
    \label{fig:driving}
\end{figure}

\begin{figure}[t]
    \centering
    \includegraphics[width=\linewidth]{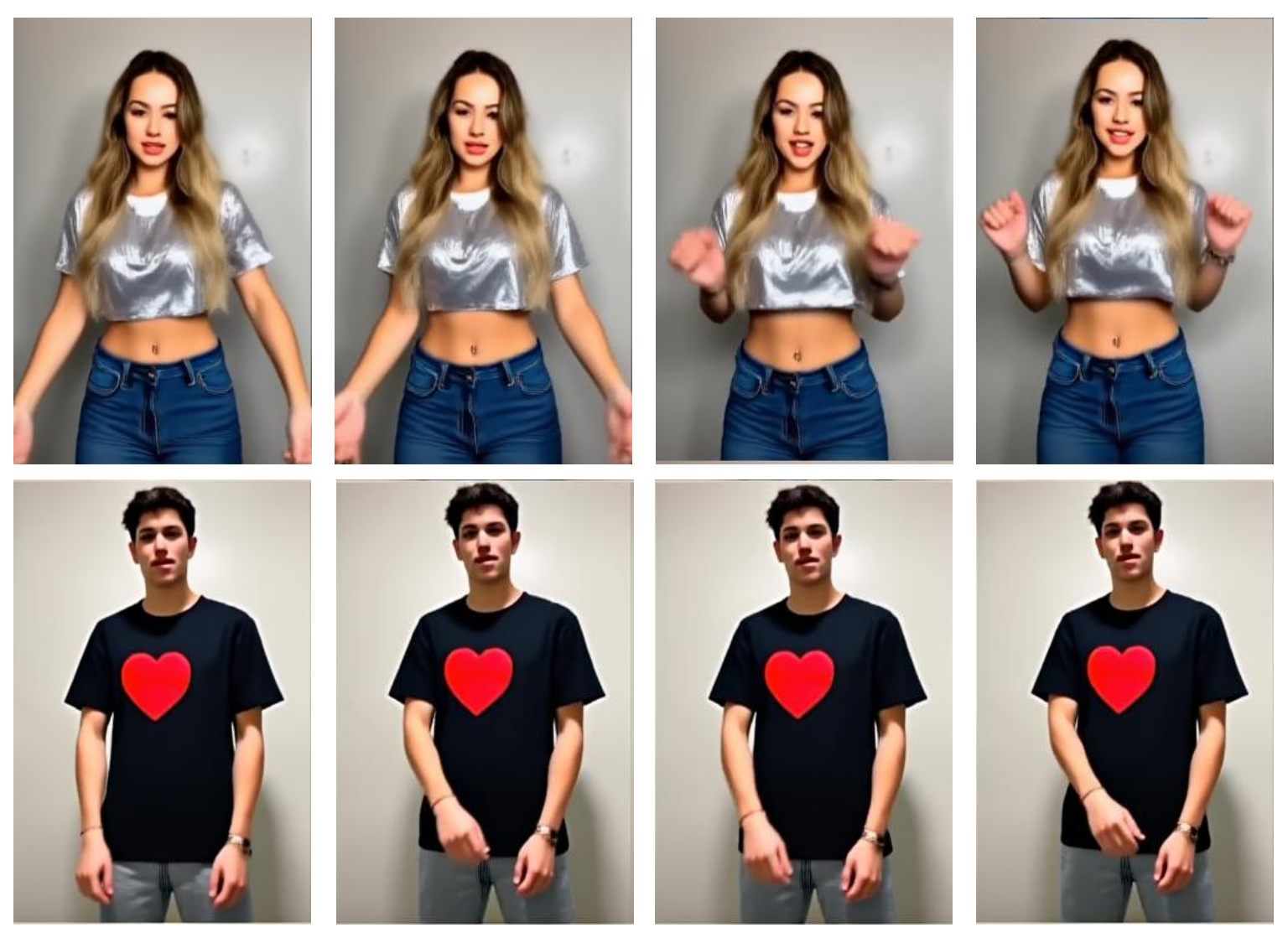}
\caption{Image-to-Video Generation of dance sequences from TikTok dataset. The leftmost column shows the input reference image, followed by generated motion sequences.}
\label{fig:dancing}
\end{figure}








%% file: figures/layout/tex/cat.tex

\begin{figure}[t]
    \centering
    \includegraphics[width=\linewidth]{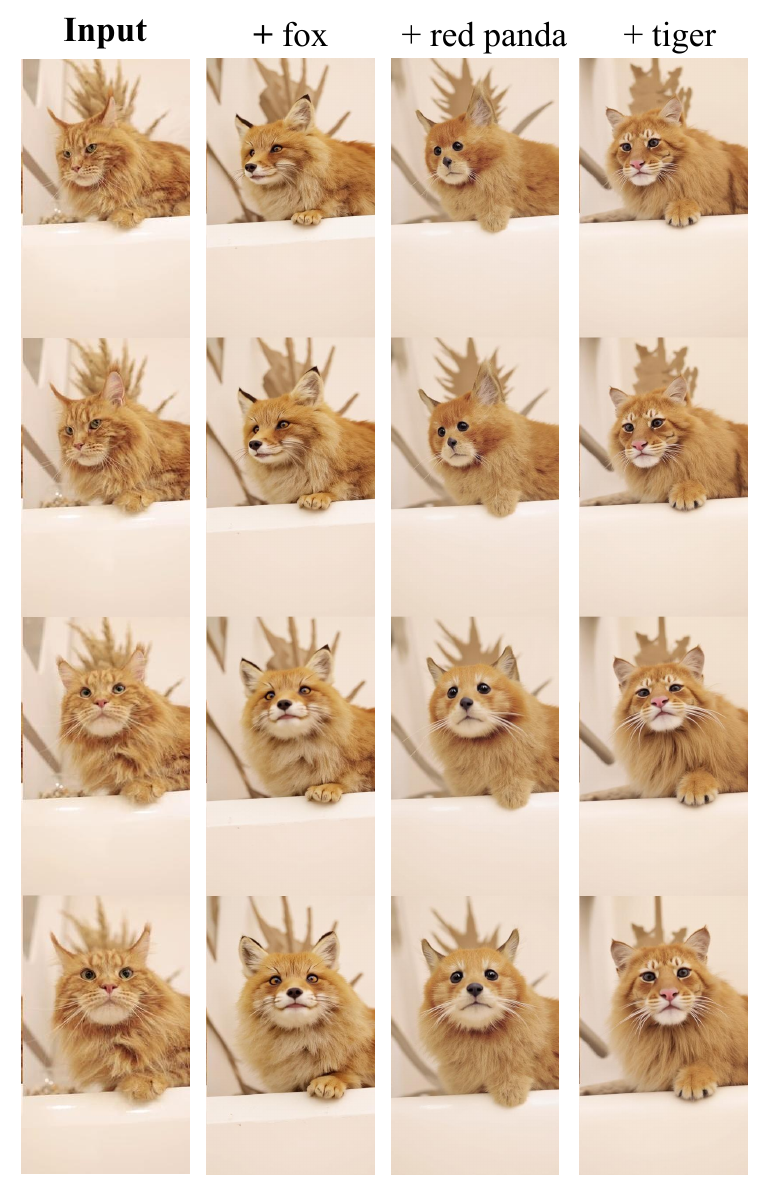}
\caption{Zero-shot video motion clone results. Our model incorporates characteristics from different animals (fox, red panda, tiger) while maintaining motion pattern.}
\label{fig:style_transfer}
     \vspace{-3mm}
\end{figure}


%% file: figures/layout/tex/creative.tex
\begin{figure}[t]
\centering
    \begin{subfigure}{0.8\linewidth}
        \centering
        \includegraphics[width=0.48\linewidth]{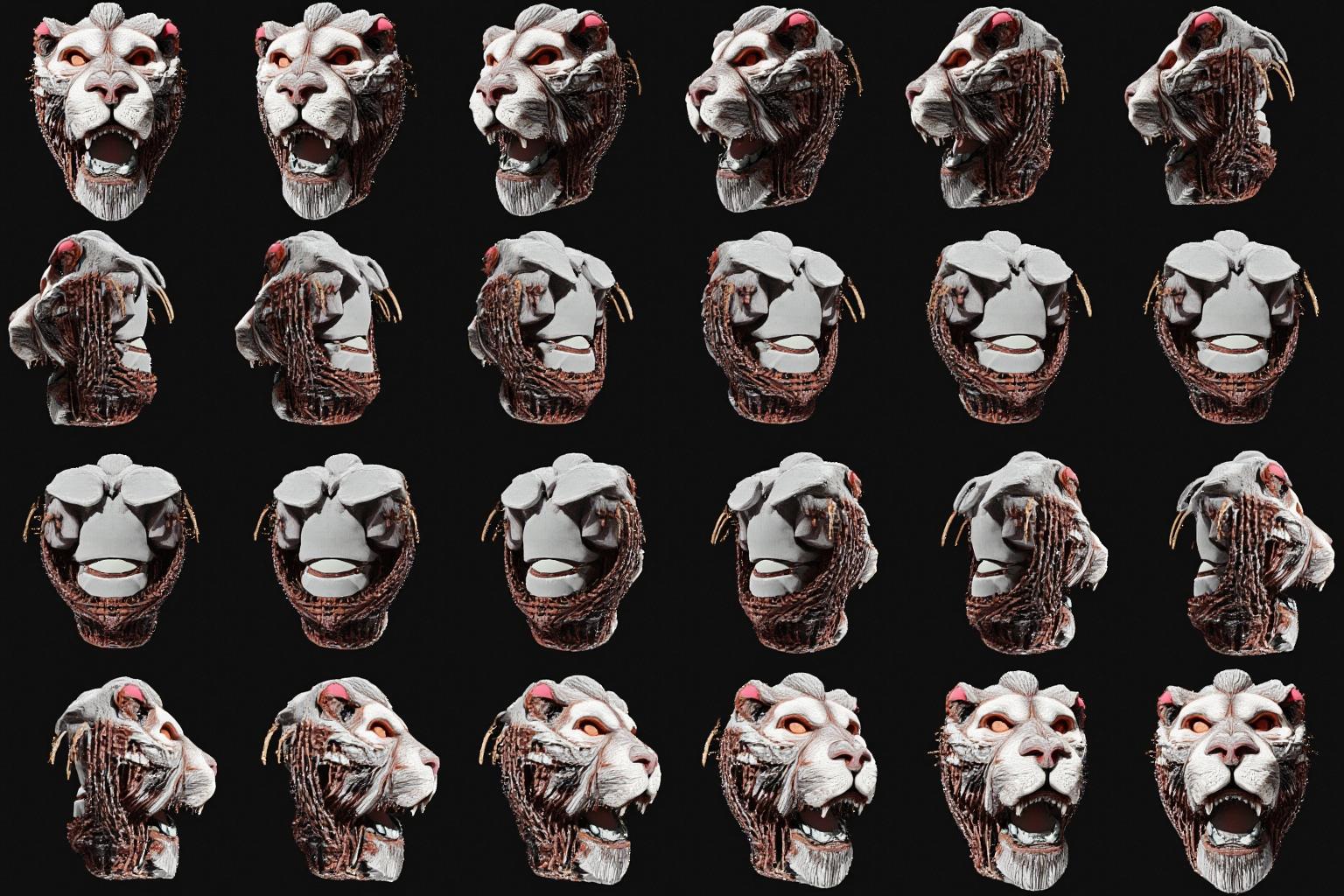}
        \hfill
        \includegraphics[width=0.48\linewidth]{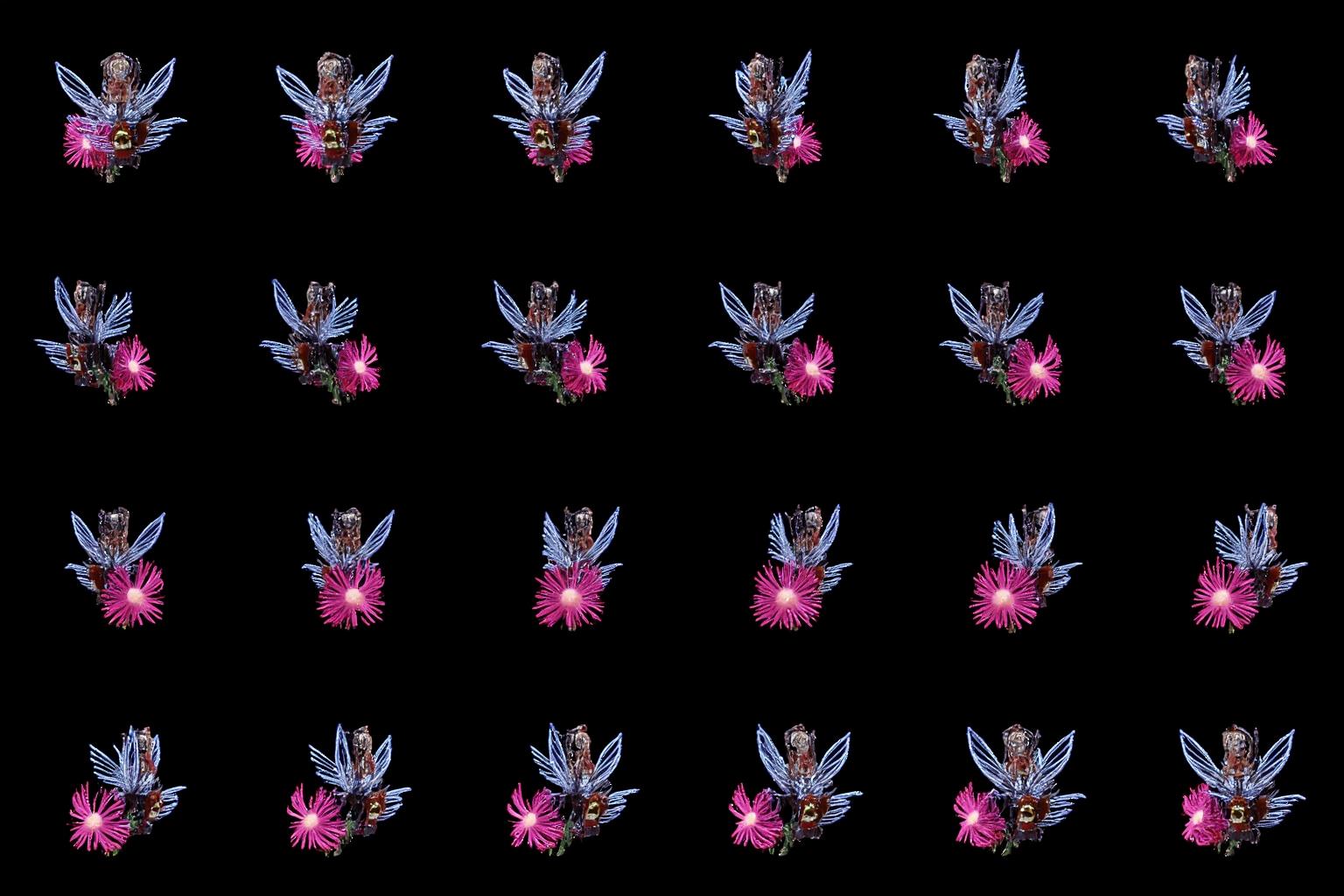}
     
    \end{subfigure}
    
    \begin{subfigure}{0.8\linewidth}
        \centering
        \includegraphics[width=0.48\linewidth]{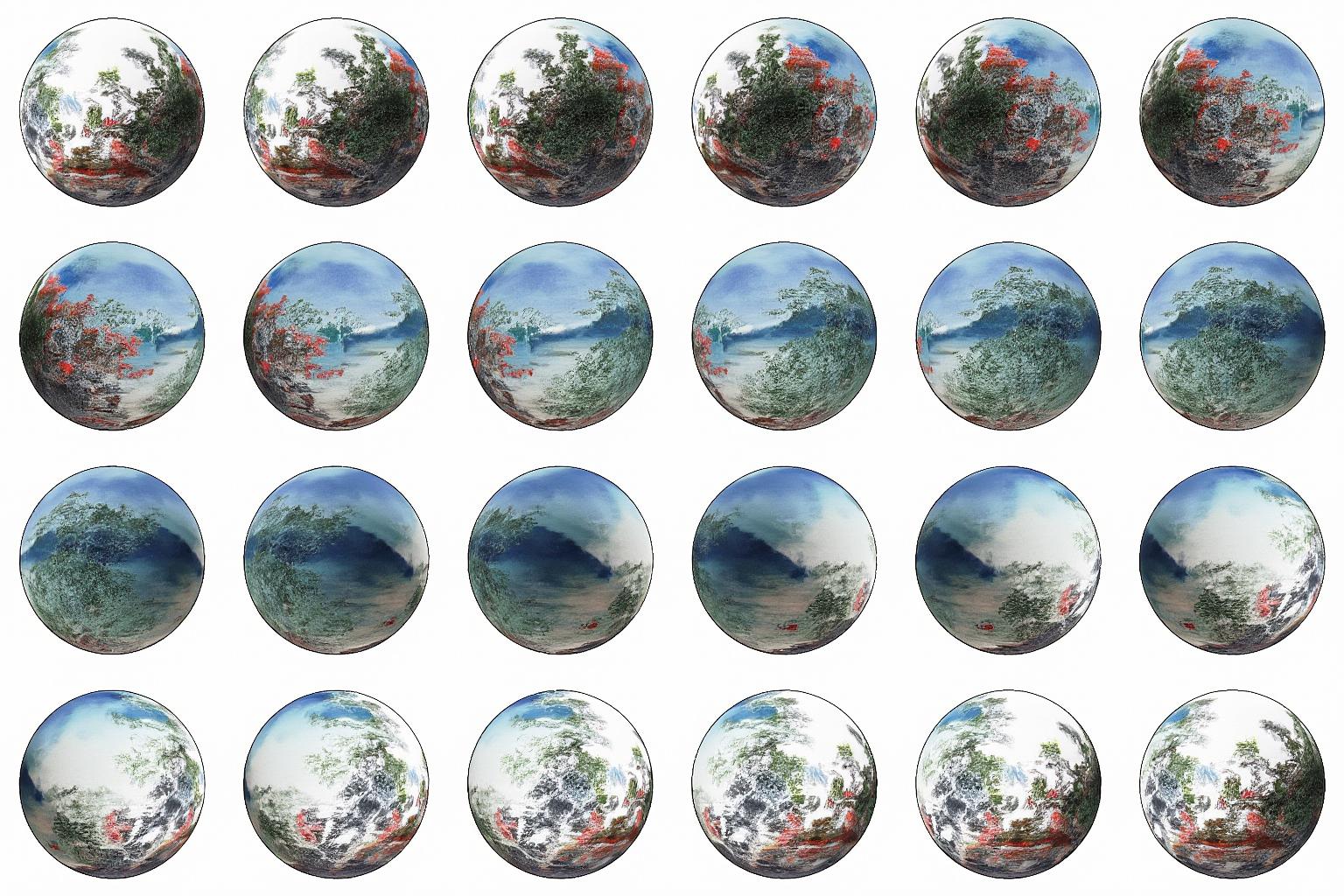}
        \hfill
        \includegraphics[width=0.48\linewidth]{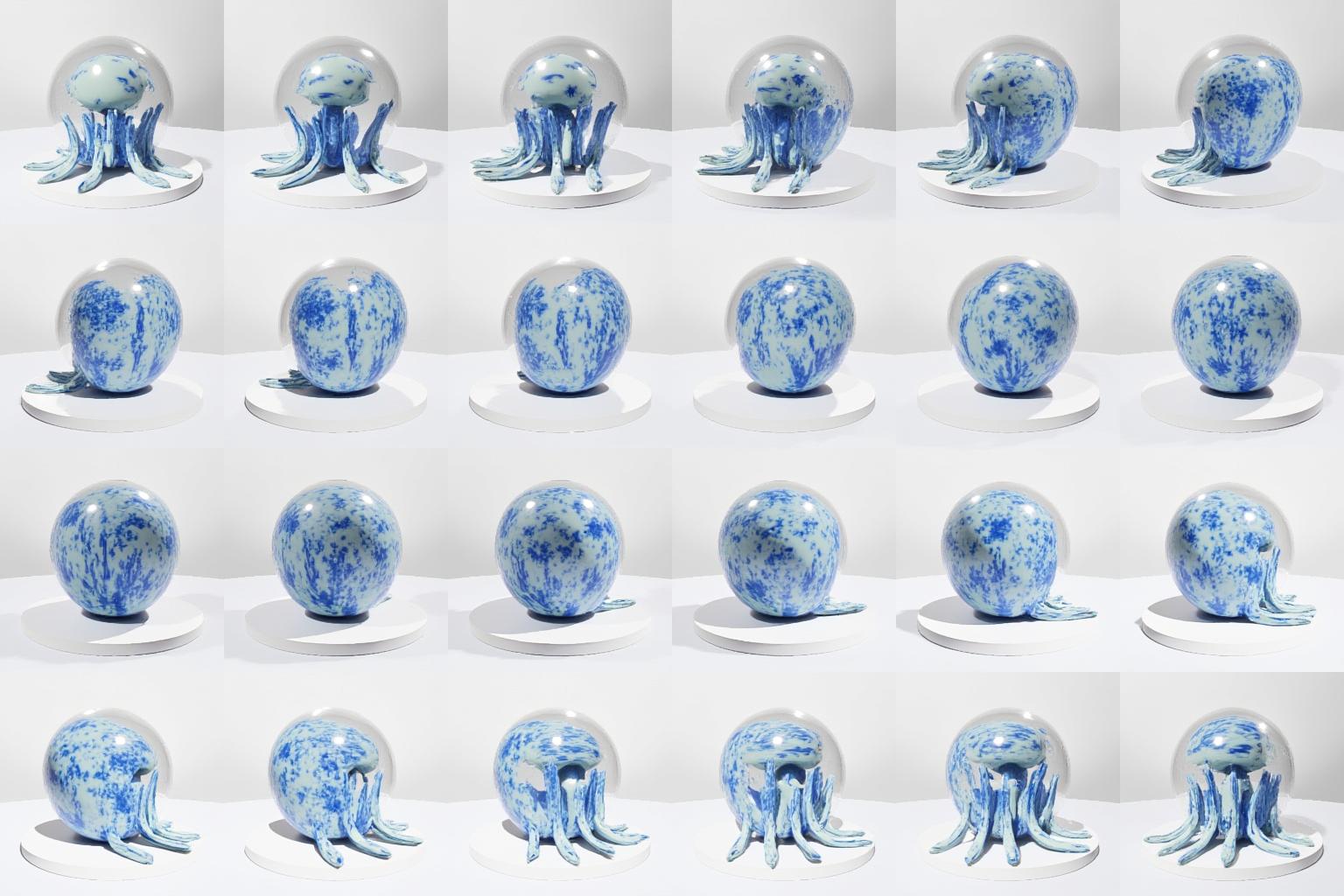}
   
    \end{subfigure}

\begin{subfigure}{0.8\linewidth}
        \centering
        \includegraphics[width=0.48\linewidth]{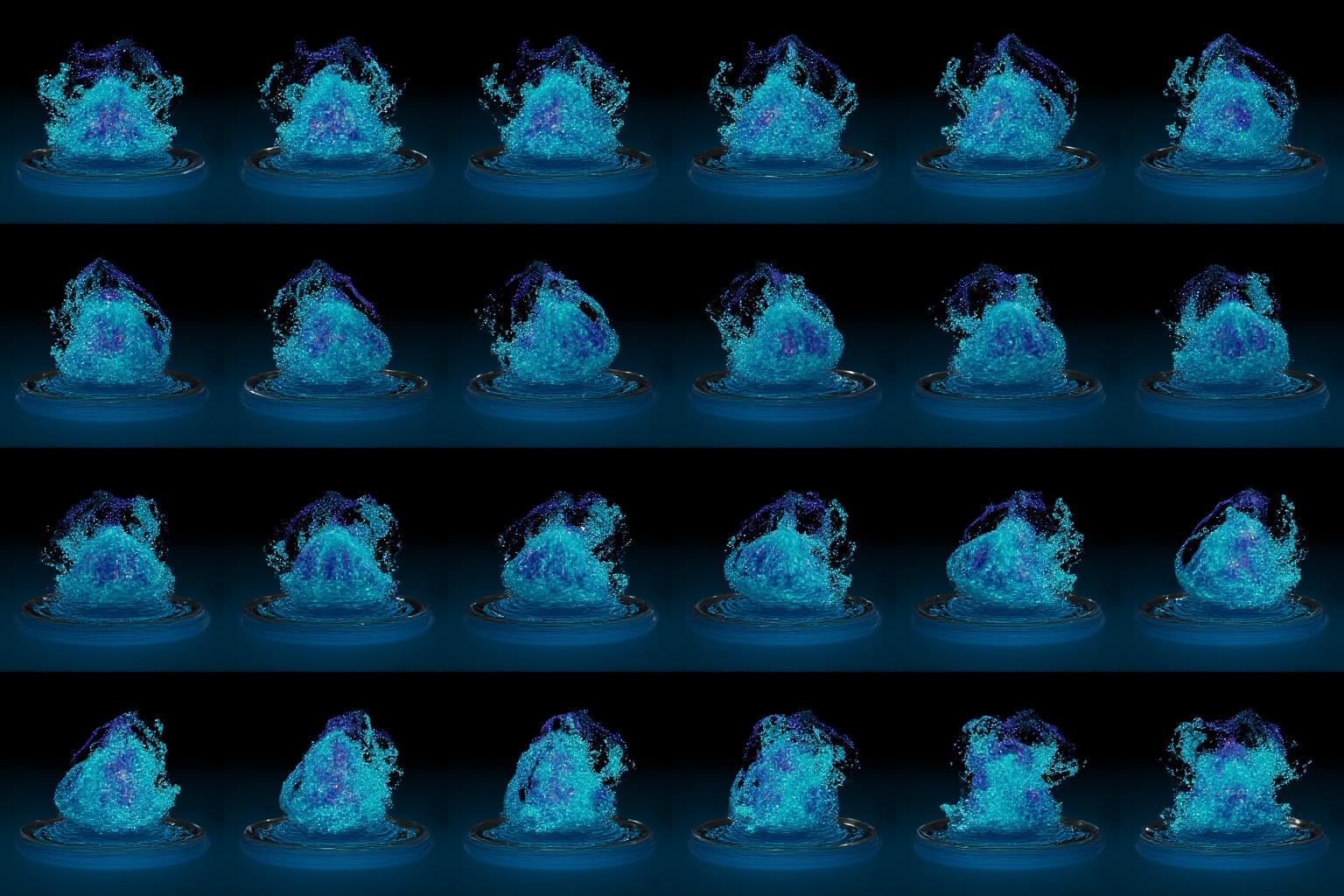}
        \hfill
        \includegraphics[width=0.48\linewidth]{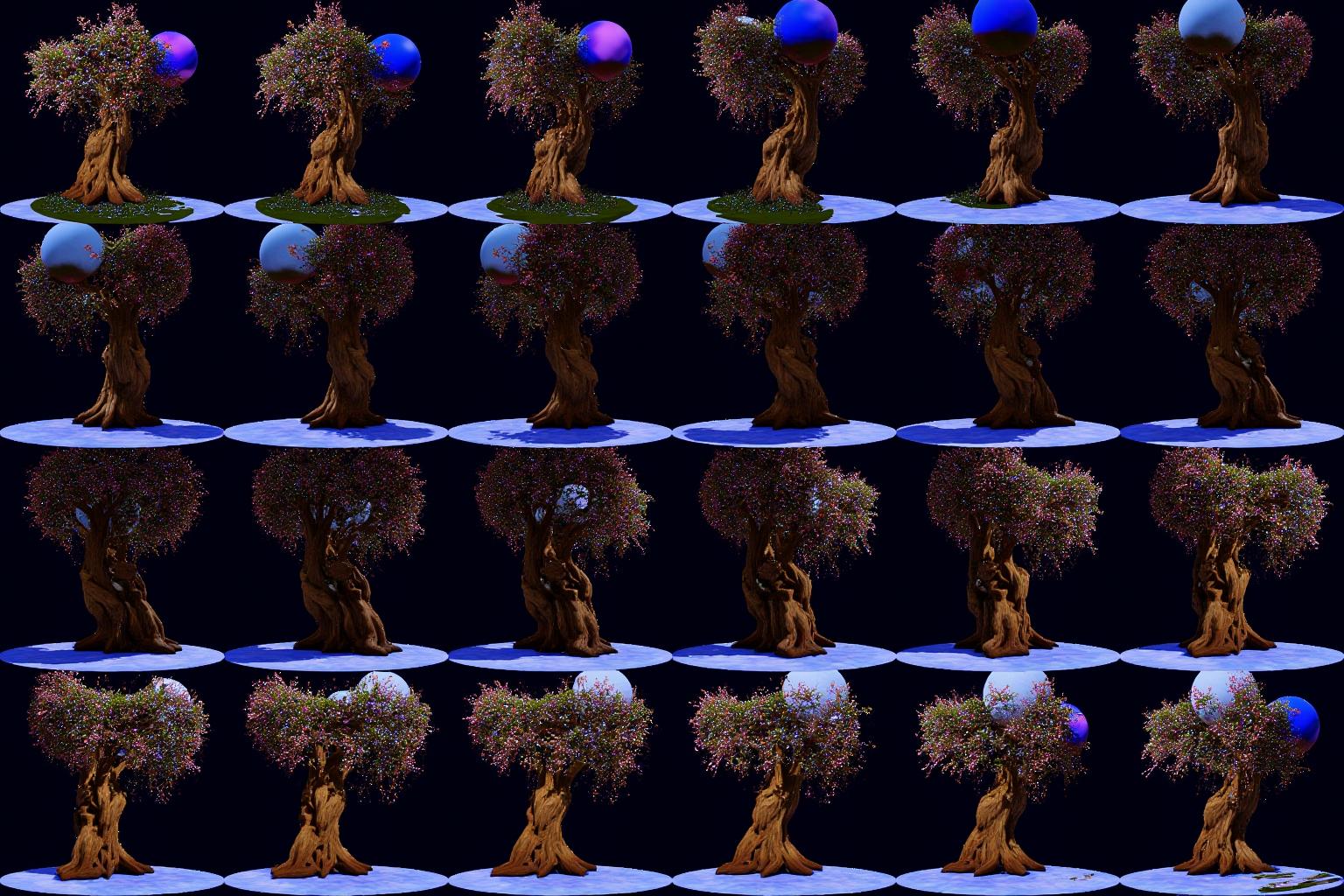}
      
    \end{subfigure}
    
    \begin{subfigure}{0.8\linewidth}
        \centering
        \includegraphics[width=0.48\linewidth]{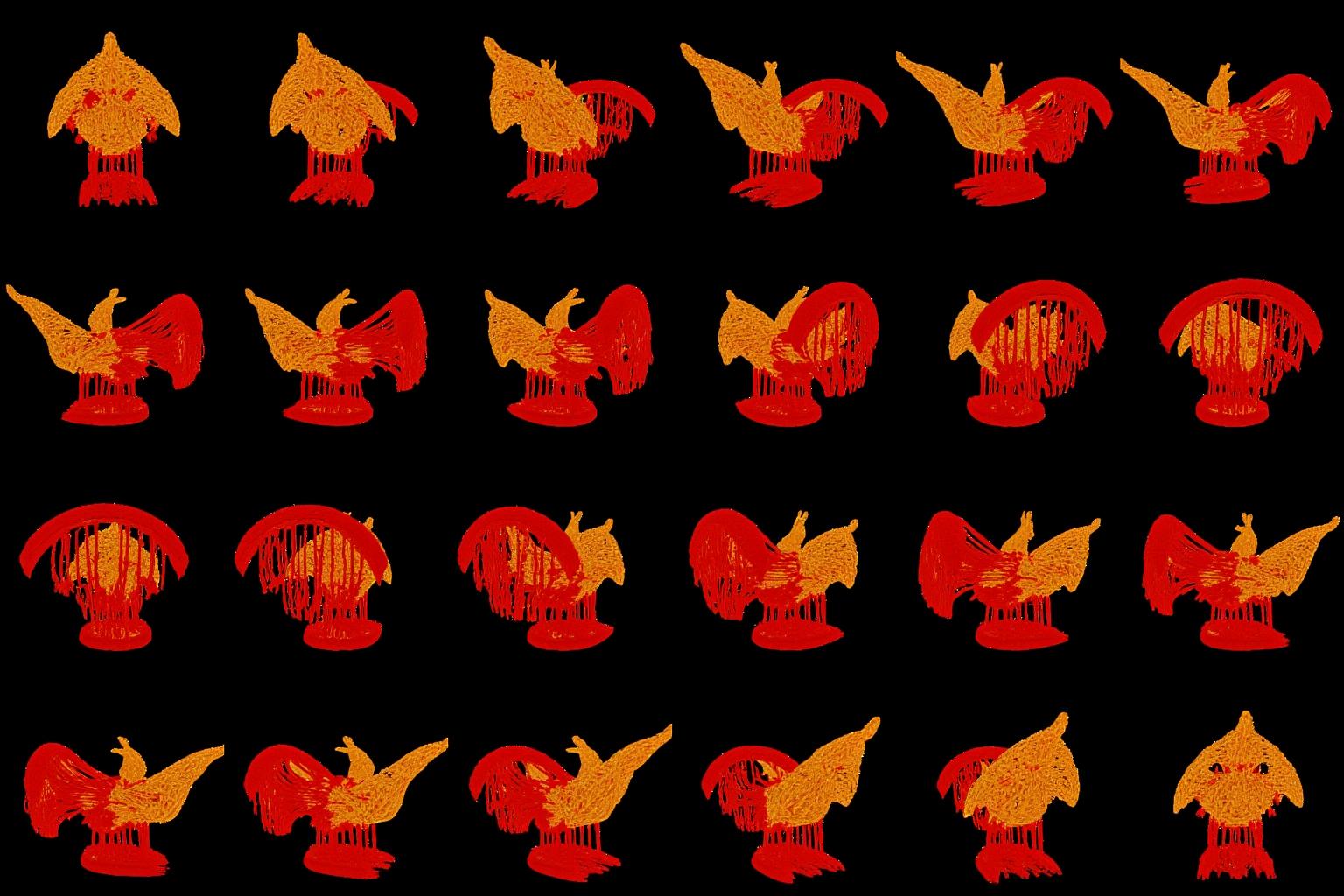}
        \hfill
        \includegraphics[width=0.48\linewidth]{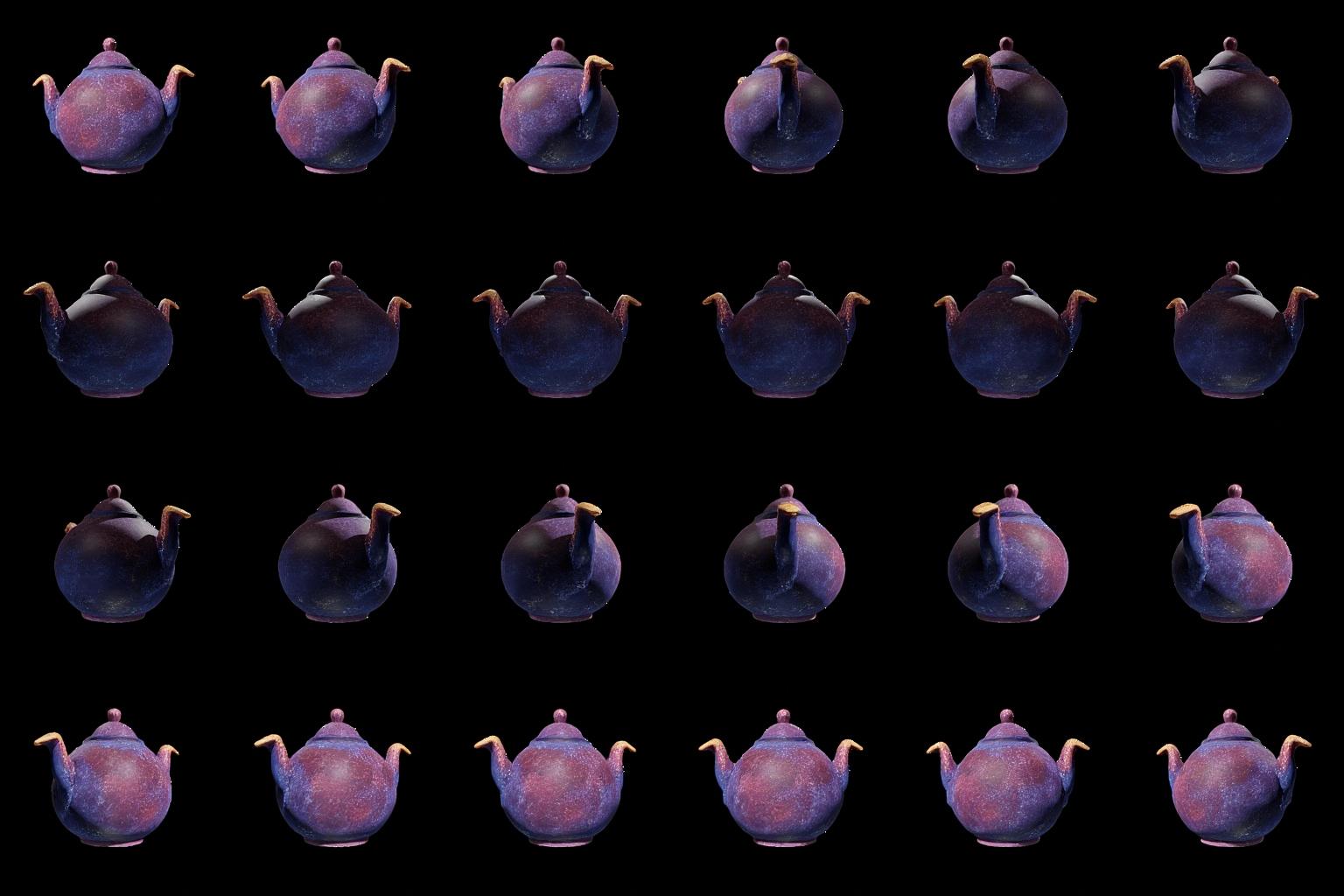}
    \end{subfigure}

    \caption{Creative multy-view concept generation.}

    \label{fig:creative}
\end{figure}

%% file: figures/layout/tex/inf4-8.tex
\begin{figure}[t]
    \centering
    \includegraphics[width=0.5\linewidth]{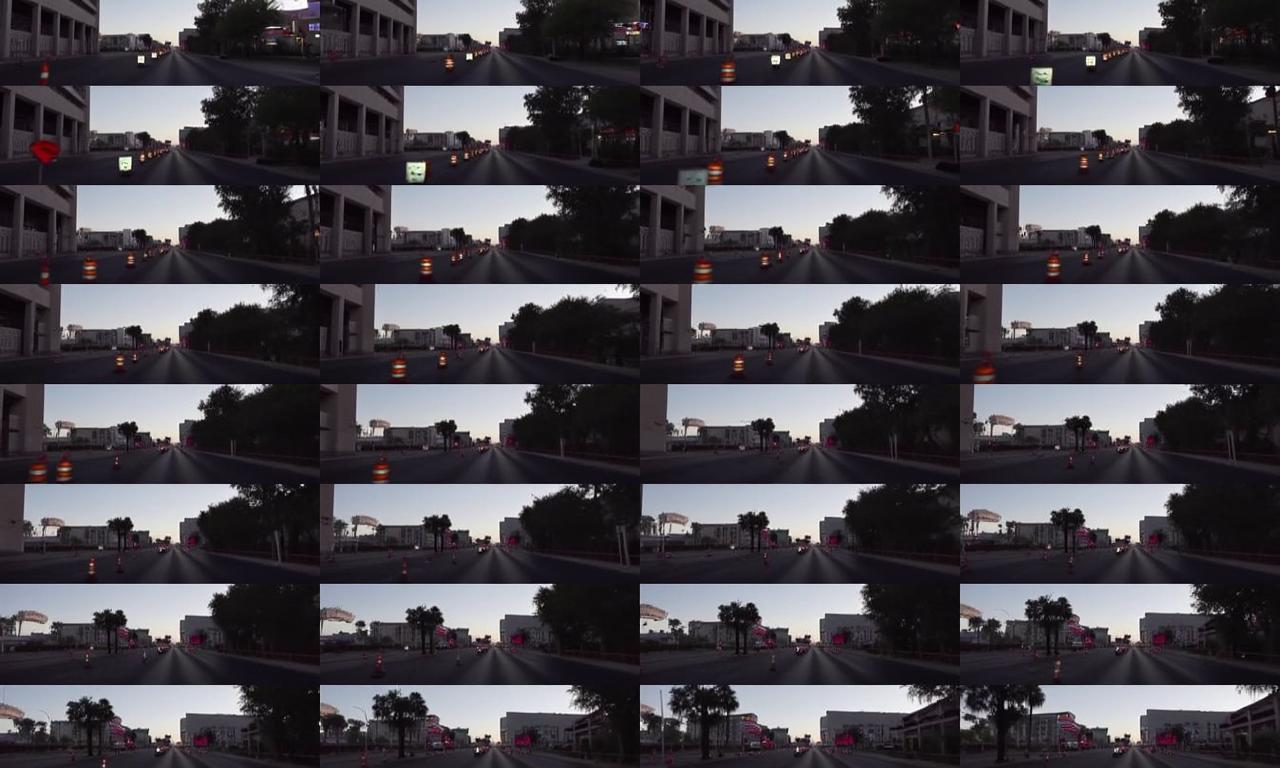}
\caption{We only train our model using 4$\times$4 datasets, but when at inference, we directly change prompt to ask to layout 4$\times$8 grid. The model has not trained on these kind of dataset, but show a zero-shot generalization ability.}
\label{fig:inf4-8}
\end{figure}

%% file: figures/layout/tex/resto.tex
\begin{figure*}[t]
    \centering
    \includegraphics[width=\textwidth]{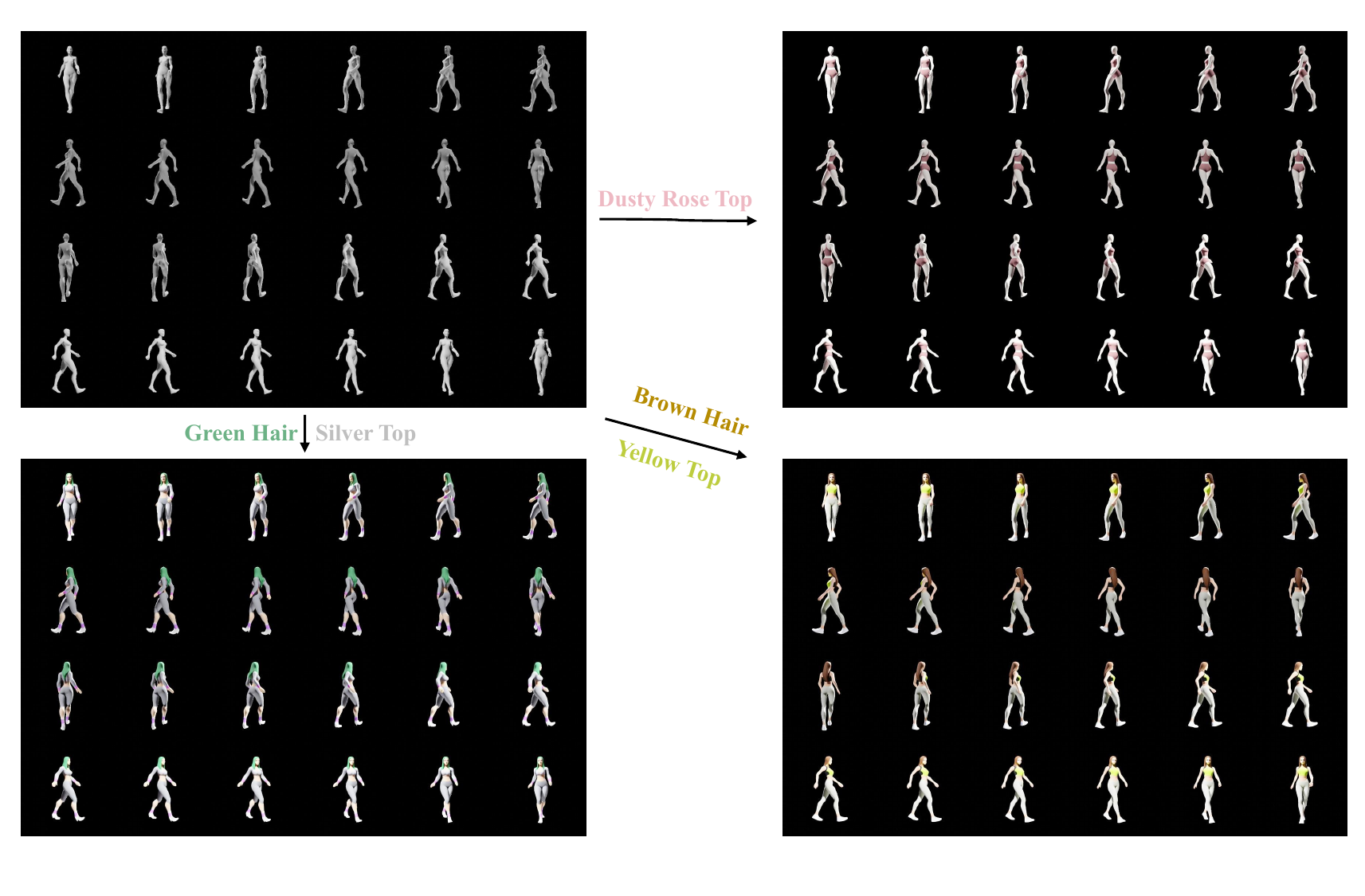}
\caption{Zero-shot 3D editing with attribute control. Our model generates diverse variations by modifying appearance attributes through text prompts while preserving motion patterns.}
\label{fig:rerender}
\end{figure*}

\begin{figure}[t]
    \centering
    \includegraphics[width=0.8\linewidth]{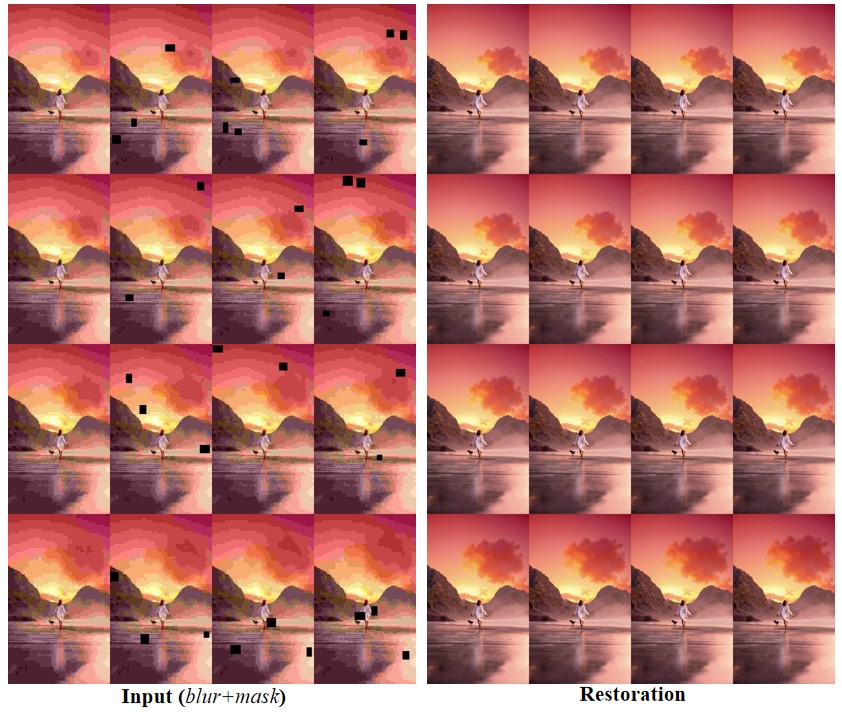}
\caption{Video restoration from degraded inputs. Left: Input sequences with Gaussian blur and block masking. Right: Restored high-quality outputs maintaining temporal consistency.}
\label{fig:restoration}
\end{figure}

%% file: main.bbl
\begin{thebibliography}{74}
\providecommand{\natexlab}[1]{#1}
\providecommand{\url}[1]{\texttt{#1}}
\expandafter\ifx\csname urlstyle\endcsname\relax
  \providecommand{\doi}[1]{doi: #1}\else
  \providecommand{\doi}{doi: \begingroup \urlstyle{rm}\Url}\fi

\bibitem[Esser et~al.(2024)Esser, Kulal, Blattmann, Entezari, M{\"u}ller, Saini, Levi, Lorenz, Sauer, Boesel, et~al.]{esser2024scaling}
Patrick Esser, Sumith Kulal, Andreas Blattmann, Rahim Entezari, Jonas M{\"u}ller, Harry Saini, Yam Levi, Dominik Lorenz, Axel Sauer, Frederic Boesel, et~al.
\newblock Scaling rectified flow transformers for high-resolution image synthesis.
\newblock In \emph{Forty-first International Conference on Machine Learning}, 2024.

\bibitem[Baldridge et~al.(2024)Baldridge, Bauer, Bhutani, Brichtova, Bunner, Chan, Chen, Dieleman, Du, Eaton-Rosen, et~al.]{baldridge2024imagen}
Jason Baldridge, Jakob Bauer, Mukul Bhutani, Nicole Brichtova, Andrew Bunner, Kelvin Chan, Yichang Chen, Sander Dieleman, Yuqing Du, Zach Eaton-Rosen, et~al.
\newblock Imagen 3.
\newblock \emph{arXiv preprint arXiv:2408.07009}, 2024.

\bibitem[Betker et~al.(2023)Betker, Goh, Jing, Brooks, Wang, Li, Ouyang, Zhuang, Lee, Guo, et~al.]{betker2023improving}
James Betker, Gabriel Goh, Li~Jing, Tim Brooks, Jianfeng Wang, Linjie Li, Long Ouyang, Juntang Zhuang, Joyce Lee, Yufei Guo, et~al.
\newblock Improving image generation with better captions.
\newblock \emph{Computer Science. https://cdn. openai. com/papers/dall-e-3. pdf}, 2\penalty0 (3):\penalty0 8, 2023.

\bibitem[Tian et~al.(2024)Tian, Jiang, Yuan, Peng, and Wang]{tian2024visual}
Keyu Tian, Yi~Jiang, Zehuan Yuan, Bingyue Peng, and Liwei Wang.
\newblock Visual autoregressive modeling: Scalable image generation via next-scale prediction.
\newblock \emph{arXiv preprint arXiv:2404.02905}, 2024.

\bibitem[Du et~al.(2022)Du, Qian, Liu, Ding, Qiu, Yang, and Tang]{DBLP:conf/acl/DuQLDQY022}
Zhengxiao Du, Yujie Qian, Xiao Liu, Ming Ding, Jiezhong Qiu, Zhilin Yang, and Jie Tang.
\newblock {GLM:} general language model pretraining with autoregressive blank infilling.
\newblock pages 320--335, 2022.

\bibitem[OpenAI(2023)]{openai2023gpt4}
OpenAI.
\newblock {GPT}-4 technical report.
\newblock \emph{arXiv:2303.08774}, 2023.

\bibitem[Bain et~al.(2021)Bain, Nagrani, Varol, and Zisserman]{bain2021frozen}
Max Bain, Arsha Nagrani, G{\"u}l Varol, and Andrew Zisserman.
\newblock Frozen in time: A joint video and image encoder for end-to-end retrieval.
\newblock In \emph{Proceedings of the IEEE/CVF international conference on computer vision}, pages 1728--1738, 2021.

\bibitem[Jafarian and Park(2022)]{jafarian2022self}
Yasamin Jafarian and Hyun~Soo Park.
\newblock Self-supervised 3d representation learning of dressed humans from social media videos.
\newblock \emph{IEEE Transactions on Pattern Analysis and Machine Intelligence}, 45\penalty0 (7):\penalty0 8969--8983, 2022.

\bibitem[Deitke et~al.(2023)Deitke, Schwenk, Salvador, Weihs, Michel, VanderBilt, Schmidt, Ehsani, Kembhavi, and Farhadi]{deitke2023objaverse}
Matt Deitke, Dustin Schwenk, Jordi Salvador, Luca Weihs, Oscar Michel, Eli VanderBilt, Ludwig Schmidt, Kiana Ehsani, Aniruddha Kembhavi, and Ali Farhadi.
\newblock Objaverse: A universe of annotated 3d objects.
\newblock In \emph{Proceedings of the IEEE/CVF Conference on Computer Vision and Pattern Recognition}, pages 13142--13153, 2023.

\bibitem[Soomro et~al.(2012)Soomro, Zamir, and Shah]{soomro2012ucf101}
Khurram Soomro, Amir~Roshan Zamir, and Mubarak Shah.
\newblock Ucf101: A dataset of 101 human actions classes from videos in the wild.
\newblock \emph{arXiv preprint arXiv:1212.0402}, 2012.

\bibitem[Unterthiner et~al.(2019)Unterthiner, van Steenkiste, Kurach, Marinier, Michalski, and Gelly]{unterthiner2019fvd}
Thomas Unterthiner, Sjoerd van Steenkiste, Karol Kurach, Rapha{\"e}l Marinier, Marcin Michalski, and Sylvain Gelly.
\newblock Fvd: A new metric for video generation.
\newblock 2019.

\bibitem[Xu et~al.(2018)Xu, Huang, Yuan, Guo, Sun, Wu, and Weinberger]{xu2018empirical}
Qiantong Xu, Gao Huang, Yang Yuan, Chuan Guo, Yu~Sun, Felix Wu, and Kilian Weinberger.
\newblock An empirical study on evaluation metrics of generative adversarial networks.
\newblock \emph{arXiv preprint arXiv:1806.07755}, 2018.

\bibitem[Liang et~al.(2024)Liang, Yin, Xu, Liang, Wang, Plataniotis, Zhao, and Wei]{diffusion4d}
Hanwen Liang, Yuyang Yin, Dejia Xu, Hanxue Liang, Zhangyang Wang, Konstantinos~N Plataniotis, Yao Zhao, and Yunchao Wei.
\newblock Diffusion4d: Fast spatial-temporal consistent 4d generation via video diffusion models.
\newblock \emph{arXiv preprint arXiv:2405.16645}, 2024.

\bibitem[Zhao et~al.(2023)Zhao, Yan, Xie, Hong, Li, and Lee]{zhao2023animate124}
Yuyang Zhao, Zhiwen Yan, Enze Xie, Lanqing Hong, Zhenguo Li, and Gim~Hee Lee.
\newblock Animate124: Animating one image to 4d dynamic scene.
\newblock \emph{arXiv preprint arXiv:2311.14603}, 2023.

\bibitem[Bahmani et~al.(2024)Bahmani, Skorokhodov, Rong, Wetzstein, Guibas, Wonka, Tulyakov, Park, Tagliasacchi, and Lindell]{bahmani20244dfy}
Sherwin Bahmani, Ivan Skorokhodov, Victor Rong, Gordon Wetzstein, Leonidas Guibas, Peter Wonka, Sergey Tulyakov, Jeong~Joon Park, Andrea Tagliasacchi, and David~B. Lindell.
\newblock 4d-fy: Text-to-4d generation using hybrid score distillation sampling.
\newblock \emph{IEEE Conference on Computer Vision and Pattern Recognition ({CVPR})}, 2024.

\bibitem[Zeng et~al.(2024)Zeng, Jiang, Zhu, Lu, Lin, Zhu, Hu, Cao, and Yao]{zeng2024stag4d}
Yifei Zeng, Yanqin Jiang, Siyu Zhu, Yuanxun Lu, Youtian Lin, Hao Zhu, Weiming Hu, Xun Cao, and Yao Yao.
\newblock Stag4d: Spatial-temporal anchored generative 4d gaussians.
\newblock 2024.

\bibitem[Yin et~al.(2023)Yin, Xu, Wang, Zhao, and Wei]{yin20234dgen}
Yuyang Yin, Dejia Xu, Zhangyang Wang, Yao Zhao, and Yunchao Wei.
\newblock 4dgen: Grounded 4d content generation with spatial-temporal consistency.
\newblock \emph{arXiv preprint arXiv:2312.17225}, 2023.

\bibitem[Guo et~al.(2023)Guo, Yang, Rao, Wang, Qiao, Lin, and Dai]{guo2023animatediff}
Yuwei Guo, Ceyuan Yang, Anyi Rao, Yaohui Wang, Yu~Qiao, Dahua Lin, and Bo~Dai.
\newblock Animatediff: Animate your personalized text-to-image diffusion models without specific tuning.
\newblock \emph{arXiv preprint arXiv:2307.04725}, 2023.

\bibitem[Zheng et~al.(2024)Zheng, Peng, Yang, Shen, Li, Liu, Zhou, Li, and You]{opensora}
Zangwei Zheng, Xiangyu Peng, Tianji Yang, Chenhui Shen, Shenggui Li, Hongxin Liu, Yukun Zhou, Tianyi Li, and Yang You.
\newblock Open-sora: Democratizing efficient video production for all.
\newblock \emph{arXiv preprint arXiv:2412.20404}, 2024.

\bibitem[Agarwal et~al.(2025)Agarwal, Ali, Bala, Balaji, Barker, Cai, Chattopadhyay, Chen, Cui, Ding, et~al.]{agarwal2025cosmos}
Niket Agarwal, Arslan Ali, Maciej Bala, Yogesh Balaji, Erik Barker, Tiffany Cai, Prithvijit Chattopadhyay, Yongxin Chen, Yin Cui, Yifan Ding, et~al.
\newblock Cosmos world foundation model platform for physical ai.
\newblock \emph{arXiv preprint arXiv:2501.03575}, 2025.

\bibitem[Yang et~al.(2024{\natexlab{a}})Yang, Teng, Zheng, Ding, Huang, Xu, Yang, Hong, Zhang, Feng, et~al.]{cogvideox}
Zhuoyi Yang, Jiayan Teng, Wendi Zheng, Ming Ding, Shiyu Huang, Jiazheng Xu, Yuanming Yang, Wenyi Hong, Xiaohan Zhang, Guanyu Feng, et~al.
\newblock Cogvideox: Text-to-video diffusion models with an expert transformer.
\newblock \emph{arXiv preprint arXiv:2408.06072}, 2024{\natexlab{a}}.

\bibitem[Zhang et~al.(2023{\natexlab{a}})Zhang, Zhu, Wang, Chen, Wu, and Wang]{zhang2023extracting}
Guozhen Zhang, Yuhan Zhu, Haonan Wang, Youxin Chen, Gangshan Wu, and Limin Wang.
\newblock Extracting motion and appearance via inter-frame attention for efficient video frame interpolation.
\newblock In \emph{Proceedings of the IEEE/CVF Conference on Computer Vision and Pattern Recognition}, pages 5682--5692, 2023{\natexlab{a}}.

\bibitem[Jin et~al.(2023)Jin, Wu, Chen, Chen, Koo, and Hahm]{jin2023unified}
Xin Jin, Longhai Wu, Jie Chen, Youxin Chen, Jayoon Koo, and Cheul-hee Hahm.
\newblock A unified pyramid recurrent network for video frame interpolation.
\newblock In \emph{Proceedings of the IEEE conference on computer vision and pattern recognition}, 2023.

\bibitem[Zhang et~al.(2024)Zhang, Liu, Cui, Zhao, Ma, and Wang]{zhang2024vfimambavideoframeinterpolation}
Guozhen Zhang, Chunxu Liu, Yutao Cui, Xiaotong Zhao, Kai Ma, and Limin Wang.
\newblock Vfimamba: Video frame interpolation with state space models, 2024.
\newblock URL \url{https://arxiv.org/abs/2407.02315}.

\bibitem[Sohl-Dickstein et~al.(2015)Sohl-Dickstein, Weiss, Maheswaranathan, and Ganguli]{sohl2015deep}
Jascha Sohl-Dickstein, Eric Weiss, Niru Maheswaranathan, and Surya Ganguli.
\newblock Deep unsupervised learning using nonequilibrium thermodynamics.
\newblock In \emph{International conference on machine learning}, pages 2256--2265. PMLR, 2015.

\bibitem[Ho et~al.(2020)Ho, Jain, and Abbeel]{ddpm}
Jonathan Ho, Ajay Jain, and Pieter Abbeel.
\newblock Denoising diffusion probabilistic models.
\newblock \emph{Advances in neural information processing systems}, 33:\penalty0 6840--6851, 2020.

\bibitem[Rombach et~al.(2022)Rombach, Blattmann, Lorenz, Esser, and Ommer]{LDM}
Robin Rombach, Andreas Blattmann, Dominik Lorenz, Patrick Esser, and Bj{\"o}rn Ommer.
\newblock High-resolution image synthesis with latent diffusion models.
\newblock In \emph{Proceedings of the IEEE/CVF conference on computer vision and pattern recognition}, pages 10684--10695, 2022.

\bibitem[Podell et~al.(2023)Podell, English, Lacey, Blattmann, Dockhorn, M{\"u}ller, Penna, and Rombach]{sdxl}
Dustin Podell, Zion English, Kyle Lacey, Andreas Blattmann, Tim Dockhorn, Jonas M{\"u}ller, Joe Penna, and Robin Rombach.
\newblock Sdxl: Improving latent diffusion models for high-resolution image synthesis.
\newblock \emph{arXiv preprint arXiv:2307.01952}, 2023.

\bibitem[Ramesh et~al.(2022)Ramesh, Dhariwal, Nichol, Chu, and Chen]{ramesh2022hierarchical}
Aditya Ramesh, Prafulla Dhariwal, Alex Nichol, Casey Chu, and Mark Chen.
\newblock Hierarchical text-conditional image generation with clip latents.
\newblock \emph{arXiv preprint arXiv:2204.06125}, 1\penalty0 (2):\penalty0 3, 2022.

\bibitem[Saharia et~al.(2022)Saharia, Chan, Saxena, Li, Whang, Denton, Ghasemipour, Karagol~Ayan, Mahdavi, Gontijo~Lopes, Salimans, Ho, Fleet, and Norouzi]{Saharia2022}
Chitwan Saharia, William Chan, Saurabh Saxena, Lala Li, Jay Whang, Emily Denton, Seyed Kamyar~Seyed Ghasemipour, Burcu Karagol~Ayan, S.~Sara Mahdavi, Rapha Gontijo~Lopes, Tim Salimans, Jonathan Ho, David Fleet, and Mohammad Norouzi.
\newblock Imagen: unprecedented photorealism × deep level of language understanding, 2022.

\bibitem[Peebles and Xie(2023)]{dit}
William Peebles and Saining Xie.
\newblock Scalable diffusion models with transformers.
\newblock In \emph{Proceedings of the IEEE/CVF International Conference on Computer Vision}, pages 4195--4205, 2023.

\bibitem[Lipman et~al.(2022)Lipman, Chen, Ben-Hamu, Nickel, and Le]{lipman2022flow}
Yaron Lipman, Ricky~TQ Chen, Heli Ben-Hamu, Maximilian Nickel, and Matt Le.
\newblock Flow matching for generative modeling.
\newblock \emph{arXiv preprint arXiv:2210.02747}, 2022.

\bibitem[BlackForest(2024)]{flux}
BlackForest.
\newblock Flux.
\newblock \url{https://github.com/black-forest-labs/flux}, 2024.

\bibitem[Raffel et~al.(2020)Raffel, Shazeer, Roberts, Lee, Narang, Matena, Zhou, Li, and Liu]{raffel2020exploring}
Colin Raffel, Noam Shazeer, Adam Roberts, Katherine Lee, Sharan Narang, Michael Matena, Yanqi Zhou, Wei Li, and Peter~J Liu.
\newblock Exploring the limits of transfer learning with a unified text-to-text transformer.
\newblock \emph{Journal of Machine Learning Research}, 21\penalty0 (1):\penalty0 5485--5551, 2020.

\bibitem[Ho et~al.(2022)Ho, Chan, Saharia, Whang, Gao, Gritsenko, Kingma, Poole, Norouzi, Fleet, et~al.]{ho2022imagen}
Jonathan Ho, William Chan, Chitwan Saharia, Jay Whang, Ruiqi Gao, Alexey Gritsenko, Diederik~P Kingma, Ben Poole, Mohammad Norouzi, David~J Fleet, et~al.
\newblock Imagen video: High definition video generation with diffusion models.
\newblock \emph{arXiv preprint arXiv:2210.02303}, 2022.

\bibitem[Blattmann et~al.(2023{\natexlab{a}})Blattmann, Rombach, Ling, Dockhorn, Kim, Fidler, and Kreis]{blattmann2023align}
Andreas Blattmann, Robin Rombach, Huan Ling, Tim Dockhorn, Seung~Wook Kim, Sanja Fidler, and Karsten Kreis.
\newblock Align your latents: High-resolution video synthesis with latent diffusion models.
\newblock In \emph{CVPR}, pages 22563--22575, 2023{\natexlab{a}}.

\bibitem[Zhang et~al.(2023{\natexlab{b}})Zhang, Wu, Liu, Zhao, Ran, Gu, Gao, and Shou]{zhang2023show}
David~Junhao Zhang, Jay~Zhangjie Wu, Jia-Wei Liu, Rui Zhao, Lingmin Ran, Yuchao Gu, Difei Gao, and Mike~Zheng Shou.
\newblock Show-1: Marrying pixel and latent diffusion models for text-to-video generation.
\newblock \emph{arXiv preprint arXiv:2309.15818}, 2023{\natexlab{b}}.

\bibitem[Blattmann et~al.(2023{\natexlab{b}})Blattmann, Dockhorn, Kulal, Mendelevitch, Kilian, Lorenz, Levi, English, Voleti, Letts, et~al.]{blattmann2023stable}
Andreas Blattmann, Tim Dockhorn, Sumith Kulal, Daniel Mendelevitch, Maciej Kilian, Dominik Lorenz, Yam Levi, Zion English, Vikram Voleti, Adam Letts, et~al.
\newblock Stable video diffusion: Scaling latent video diffusion models to large datasets.
\newblock \emph{arXiv preprint arXiv:2311.15127}, 2023{\natexlab{b}}.

\bibitem[He et~al.(2023)He, Yang, Zhang, Shan, and Chen]{he2023latent}
Yingqing He, Tianyu Yang, Yong Zhang, Ying Shan, and Qifeng Chen.
\newblock Latent video diffusion models for high-fidelity long video generation.
\newblock \emph{arXiv preprint arXiv:2211.13221}, 2\penalty0 (3):\penalty0 4, 2023.

\bibitem[Zhou et~al.(2022)Zhou, Wang, Yan, Lv, Zhu, and Feng]{zhou2022magicvideo}
Daquan Zhou, Weimin Wang, Hanshu Yan, Weiwei Lv, Yizhe Zhu, and Jiashi Feng.
\newblock Magicvideo: Efficient video generation with latent diffusion models.
\newblock \emph{arXiv preprint arXiv:2211.11018}, 2022.

\bibitem[Wang et~al.(2023{\natexlab{a}})Wang, Yuan, Chen, Zhang, Wang, and Zhang]{wang2023modelscope}
Jiuniu Wang, Hangjie Yuan, Dayou Chen, Yingya Zhang, Xiang Wang, and Shiwei Zhang.
\newblock Modelscope text-to-video technical report.
\newblock \emph{arXiv preprint arXiv:2308.06571}, 2023{\natexlab{a}}.

\bibitem[Ge et~al.(2023)Ge, Nah, Liu, Poon, Tao, Catanzaro, Jacobs, Huang, Liu, and Balaji]{ge2023preserve}
Songwei Ge, Seungjun Nah, Guilin Liu, Tyler Poon, Andrew Tao, Bryan Catanzaro, David Jacobs, Jia-Bin Huang, Ming-Yu Liu, and Yogesh Balaji.
\newblock Preserve your own correlation: A noise prior for video diffusion models.
\newblock In \emph{CVPR}, pages 22930--22941, 2023.

\bibitem[Wang et~al.(2023{\natexlab{b}})Wang, He, Li, Li, Yu, Ma, Chen, Wang, Luo, Liu, et~al.]{wang2023internvid}
Yi~Wang, Yinan He, Yizhuo Li, Kunchang Li, Jiashuo Yu, Xin Ma, Xinyuan Chen, Yaohui Wang, Ping Luo, Ziwei Liu, et~al.
\newblock Internvid: A large-scale video-text dataset for multimodal understanding and generation.
\newblock \emph{arXiv preprint arXiv:2307.06942}, 2023{\natexlab{b}}.

\bibitem[Wang et~al.(2023{\natexlab{c}})Wang, Yang, Tuo, He, Zhu, Fu, and Liu]{wang2023videofactory}
Wenjing Wang, Huan Yang, Zixi Tuo, Huiguo He, Junchen Zhu, Jianlong Fu, and Jiaying Liu.
\newblock Videofactory: Swap attention in spatiotemporal diffusions for text-to-video generation.
\newblock \emph{arXiv preprint arXiv:2305.10874}, 2023{\natexlab{c}}.

\bibitem[Singer et~al.(2022)Singer, Polyak, Hayes, Yin, An, Zhang, Hu, Yang, Ashual, Gafni, et~al.]{singer2022make}
Uriel Singer, Adam Polyak, Thomas Hayes, Xi~Yin, Jie An, Songyang Zhang, Qiyuan Hu, Harry Yang, Oron Ashual, Oran Gafni, et~al.
\newblock Make-a-video: Text-to-video generation without text-video data.
\newblock \emph{arXiv preprint arXiv:2209.14792}, 2022.

\bibitem[Zeng et~al.(2023)Zeng, Wei, Zheng, Zou, Wei, Zhang, and Li]{zeng2023make}
Yan Zeng, Guoqiang Wei, Jiani Zheng, Jiaxin Zou, Yang Wei, Yuchen Zhang, and Hang Li.
\newblock Make pixels dance: High-dynamic video generation.
\newblock \emph{arXiv preprint arXiv:2311.10982}, 2023.

\bibitem[Chai et~al.(2023)Chai, Guo, Wang, and Lu]{chai2023stablevideo}
Wenhao Chai, Xun Guo, Gaoang Wang, and Yan Lu.
\newblock Stablevideo: Text-driven consistency-aware diffusion video editing.
\newblock In \emph{CVPR}, pages 23040--23050, 2023.

\bibitem[Watson et~al.(2022)Watson, Chan, Martin-Brualla, Ho, Tagliasacchi, and Norouzi]{3DiM}
Daniel Watson, William Chan, Ricardo Martin-Brualla, Jonathan Ho, Andrea Tagliasacchi, and Mohammad Norouzi.
\newblock Novel view synthesis with diffusion models.
\newblock \emph{arXiv preprint arXiv:2210.04628}, 2022.

\bibitem[Liu et~al.(2023{\natexlab{a}})Liu, Wu, Van~Hoorick, Tokmakov, Zakharov, and Vondrick]{zero1to3}
Ruoshi Liu, Rundi Wu, Basile Van~Hoorick, Pavel Tokmakov, Sergey Zakharov, and Carl Vondrick.
\newblock Zero-1-to-3: Zero-shot one image to 3d object.
\newblock In \emph{Proceedings of the IEEE/CVF international conference on computer vision}, pages 9298--9309, 2023{\natexlab{a}}.

\bibitem[Shi et~al.(2023{\natexlab{a}})Shi, Wang, Ye, Long, Li, and Yang]{mvdream}
Yichun Shi, Peng Wang, Jianglong Ye, Mai Long, Kejie Li, and Xiao Yang.
\newblock Mvdream: Multi-view diffusion for 3d generation.
\newblock \emph{arXiv preprint arXiv:2308.16512}, 2023{\natexlab{a}}.

\bibitem[Long et~al.(2024)Long, Guo, Lin, Liu, Dou, Liu, Ma, Zhang, Habermann, Theobalt, et~al.]{wonder3d}
Xiaoxiao Long, Yuan-Chen Guo, Cheng Lin, Yuan Liu, Zhiyang Dou, Lingjie Liu, Yuexin Ma, Song-Hai Zhang, Marc Habermann, Christian Theobalt, et~al.
\newblock Wonder3d: Single image to 3d using cross-domain diffusion.
\newblock In \emph{Proceedings of the IEEE/CVF Conference on Computer Vision and Pattern Recognition}, pages 9970--9980, 2024.

\bibitem[Shi et~al.(2023{\natexlab{b}})Shi, Chen, Zhang, Liu, Xu, Wei, Chen, Zeng, and Su]{zero123++}
Ruoxi Shi, Hansheng Chen, Zhuoyang Zhang, Minghua Liu, Chao Xu, Xinyue Wei, Linghao Chen, Chong Zeng, and Hao Su.
\newblock Zero123++: a single image to consistent multi-view diffusion base model.
\newblock \emph{arXiv preprint arXiv:2310.15110}, 2023{\natexlab{b}}.

\bibitem[Lu et~al.(2024)Lu, Zhang, Li, Fang, McKinnon, Tsin, Quan, Cao, and Yao]{Direct2.5}
Yuanxun Lu, Jingyang Zhang, Shiwei Li, Tian Fang, David McKinnon, Yanghai Tsin, Long Quan, Xun Cao, and Yao Yao.
\newblock Direct2. 5: Diverse text-to-3d generation via multi-view 2.5 d diffusion.
\newblock In \emph{Proceedings of the IEEE/CVF Conference on Computer Vision and Pattern Recognition}, pages 8744--8753, 2024.

\bibitem[Li et~al.(2023{\natexlab{a}})Li, Tan, Zhang, Xu, Luan, Xu, Hong, Sunkavalli, Shakhnarovich, and Bi]{Instant3D}
Jiahao Li, Hao Tan, Kai Zhang, Zexiang Xu, Fujun Luan, Yinghao Xu, Yicong Hong, Kalyan Sunkavalli, Greg Shakhnarovich, and Sai Bi.
\newblock Instant3d: Fast text-to-3d with sparse-view generation and large reconstruction model.
\newblock \emph{arXiv preprint arXiv:2311.06214}, 2023{\natexlab{a}}.

\bibitem[Liu et~al.(2023{\natexlab{b}})Liu, Lin, Zeng, Long, Liu, Komura, and Wang]{syncdreamer}
Yuan Liu, Cheng Lin, Zijiao Zeng, Xiaoxiao Long, Lingjie Liu, Taku Komura, and Wenping Wang.
\newblock Syncdreamer: Generating multiview-consistent images from a single-view image.
\newblock \emph{arXiv preprint arXiv:2309.03453}, 2023{\natexlab{b}}.

\bibitem[Li et~al.(2024{\natexlab{a}})Li, Liu, Long, Zhang, Lin, Li, Qi, Zhang, Luo, Tan, et~al.]{Era3d}
Peng Li, Yuan Liu, Xiaoxiao Long, Feihu Zhang, Cheng Lin, Mengfei Li, Xingqun Qi, Shanghang Zhang, Wenhan Luo, Ping Tan, et~al.
\newblock Era3d: High-resolution multiview diffusion using efficient row-wise attention.
\newblock \emph{arXiv preprint arXiv:2405.11616}, 2024{\natexlab{a}}.

\bibitem[Yang et~al.(2024{\natexlab{b}})Yang, Shi, Zhang, Yang, Wang, Zhao, Liu, Wang, Lin, Yu, et~al.]{hunyuan3d}
Xianghui Yang, Huiwen Shi, Bowen Zhang, Fan Yang, Jiacheng Wang, Hongxu Zhao, Xinhai Liu, Xinzhou Wang, Qingxiang Lin, Jiaao Yu, et~al.
\newblock Hunyuan3d-1.0: A unified framework for text-to-3d and image-to-3d generation.
\newblock \emph{arXiv preprint arXiv:2411.02293}, 2024{\natexlab{b}}.

\bibitem[Ren et~al.(2023)Ren, Pan, Tang, Zhang, Cao, Zeng, and Liu]{dreamgaussian4d}
Jiawei Ren, Liang Pan, Jiaxiang Tang, Chi Zhang, Ang Cao, Gang Zeng, and Ziwei Liu.
\newblock Dreamgaussian4d: Generative 4d gaussian splatting.
\newblock \emph{arXiv preprint arXiv:2312.17142}, 2023.

\bibitem[Xie et~al.(2024{\natexlab{a}})Xie, Yao, Voleti, Jiang, and Jampani]{sv4d}
Yiming Xie, Chun-Han Yao, Vikram Voleti, Huaizu Jiang, and Varun Jampani.
\newblock Sv4d: Dynamic 3d content generation with multi-frame and multi-view consistency.
\newblock \emph{arXiv preprint arXiv:2407.17470}, 2024{\natexlab{a}}.

\bibitem[Sun et~al.(2024{\natexlab{a}})Sun, Chen, Liu, Chen, Duan, Zhang, and Wang]{dimensionx}
Wenqiang Sun, Shuo Chen, Fangfu Liu, Zilong Chen, Yueqi Duan, Jun Zhang, and Yikai Wang.
\newblock Dimensionx: Create any 3d and 4d scenes from a single image with controllable video diffusion.
\newblock \emph{arXiv preprint arXiv:2411.04928}, 2024{\natexlab{a}}.

\bibitem[Wu et~al.(2024)Wu, Gao, Poole, Trevithick, Zheng, Barron, and Holynski]{cat4d}
Rundi Wu, Ruiqi Gao, Ben Poole, Alex Trevithick, Changxi Zheng, Jonathan~T Barron, and Aleksander Holynski.
\newblock Cat4d: Create anything in 4d with multi-view video diffusion models.
\newblock \emph{arXiv preprint arXiv:2411.18613}, 2024.

\bibitem[Kondratyuk et~al.(2023)Kondratyuk, Yu, Gu, Lezama, Huang, Schindler, Hornung, Birodkar, Yan, Chiu, et~al.]{kondratyuk2023videopoet}
Dan Kondratyuk, Lijun Yu, Xiuye Gu, Jos{\'e} Lezama, Jonathan Huang, Grant Schindler, Rachel Hornung, Vighnesh Birodkar, Jimmy Yan, Ming-Chang Chiu, et~al.
\newblock Videopoet: A large language model for zero-shot video generation.
\newblock \emph{arXiv preprint arXiv:2312.14125}, 2023.

\bibitem[Huang et~al.(2024{\natexlab{a}})Huang, Wang, Wu, Shi, Dou, Liang, Feng, Liu, and Zhou]{lhhuang2024iclora}
Lianghua Huang, Wei Wang, Zhi-Fan Wu, Yupeng Shi, Huanzhang Dou, Chen Liang, Yutong Feng, Yu~Liu, and Jingren Zhou.
\newblock In-context lora for diffusion transformers.
\newblock \emph{arXiv preprint arxiv:2410.23775}, 2024{\natexlab{a}}.

\bibitem[Huang et~al.(2024{\natexlab{b}})Huang, Wang, Wu, Dou, Shi, Feng, Liang, Liu, and Zhou]{lhhuang2024groupdiffusion}
Lianghua Huang, Wei Wang, Zhi-Fan Wu, Huanzhang Dou, Yupeng Shi, Yutong Feng, Chen Liang, Yu~Liu, and Jingren Zhou.
\newblock Group diffusion transformers are unsupervised multitask learners.
\newblock \emph{arXiv preprint arxiv:2410.15027}, 2024{\natexlab{b}}.

\bibitem[Li et~al.(2023{\natexlab{b}})Li, Tan, Zhang, Xu, Luan, Xu, Hong, Sunkavalli, Shakhnarovich, and Bi]{li2023instant3d}
Jiahao Li, Hao Tan, Kai Zhang, Zexiang Xu, Fujun Luan, Yinghao Xu, Yicong Hong, Kalyan Sunkavalli, Greg Shakhnarovich, and Sai Bi.
\newblock Instant3d: Fast text-to-3d with sparse-view generation and large reconstruction model.
\newblock \emph{arXiv preprint arXiv:2311.06214}, 2023{\natexlab{b}}.

\bibitem[Cai et~al.(2025)Cai, Chan, Zhang, Guibas, Wu, and Wetzstein]{cai2024dsd}
Shengqu Cai, Eric Chan, Yunzhi Zhang, Leonidas Guibas, Jiajun Wu, and Gordon. Wetzstein.
\newblock Diffusion self-distillation for zero-shot customized image generation.
\newblock In \emph{CVPR}, 2025.

\bibitem[Xie et~al.(2024{\natexlab{b}})Xie, Jampani, Zhong, Sun, and Jiang]{xie2024omnicontrol}
Yiming Xie, Varun Jampani, Lei Zhong, Deqing Sun, and Huaizu Jiang.
\newblock Omnicontrol: Control any joint at any time for human motion generation.
\newblock In \emph{The Twelfth International Conference on Learning Representations}, 2024{\natexlab{b}}.
\newblock URL \url{https://openreview.net/forum?id=gd0lAEtWso}.

\bibitem[Wang et~al.()Wang, Xie, Dong, and Shan]{wang2021realesrgan}
Xintao Wang, Liangbin Xie, Chao Dong, and Ying Shan.
\newblock Real-esrgan: Training real-world blind super-resolution with pure synthetic data.
\newblock In \emph{International Conference on Computer Vision Workshops (ICCVW)}.

\bibitem[Sun et~al.(2024{\natexlab{b}})Sun, Huang, Liu, Wu, Xu, Li, and Liu]{sun2024t2v}
Kaiyue Sun, Kaiyi Huang, Xian Liu, Yue Wu, Zihan Xu, Zhenguo Li, and Xihui Liu.
\newblock T2v-compbench: A comprehensive benchmark for compositional text-to-video generation.
\newblock \emph{arXiv preprint arXiv:2407.14505}, 2024{\natexlab{b}}.

\bibitem[Bai et~al.(2024)Bai, Wu, Liu, Jia, Mao, Zhang, Zhao, Shen, Wei, Wang, et~al.]{bai20243d}
Yifan Bai, Dongming Wu, Yingfei Liu, Fan Jia, Weixin Mao, Ziheng Zhang, Yucheng Zhao, Jianbing Shen, Xing Wei, Tiancai Wang, et~al.
\newblock Is a 3d-tokenized llm the key to reliable autonomous driving?
\newblock \emph{arXiv preprint arXiv:2405.18361}, 2024.

\bibitem[Li et~al.(2024{\natexlab{b}})Li, Sun, Zhang, Ye, Liao, Feng, Zhao, and He]{anydressing}
Xinghui Li, Qichao Sun, Pengze Zhang, Fulong Ye, Zhichao Liao, Wanquan Feng, Songtao Zhao, and Qian He.
\newblock Anydressing: Customizable multi-garment virtual dressing via latent diffusion models.
\newblock \emph{arXiv preprint arXiv:2412.04146}, 2024{\natexlab{b}}.

\bibitem[Xian et~al.(2025)Xian, Liao, Li, Qin, Wan, Xie, Zeng, Shen, and Feng]{spf}
Xiaole Xian, Zhichao Liao, Qingyu Li, Wenyu Qin, Pengfei Wan, Weicheng Xie, Long Zeng, Linlin Shen, and Pingfa Feng.
\newblock Spf-portrait: Towards pure portrait customization with semantic pollution-free fine-tuning.
\newblock \emph{arXiv preprint arXiv:2504.00396}, 2025.

\bibitem[Luo et~al.(2024)Luo, Zhang, Xie, Tong, Yu, Chang, Ma, and Yu]{luo2024codeswap}
Xiangyang Luo, Xin Zhang, Yifan Xie, Xinyi Tong, Weijiang Yu, Heng Chang, Fei Ma, and Fei~Richard Yu.
\newblock Codeswap: Symmetrically face swapping based on prior codebook.
\newblock In \emph{Proceedings of the 32nd ACM International Conference on Multimedia}, pages 6910--6919, 2024.

\bibitem[Wan et~al.(2024)Wan, He, Song, and Gong]{wan2024prompt}
Cong Wan, Yuhang He, Xiang Song, and Yihong Gong.
\newblock Prompt-agnostic adversarial perturbation for customized diffusion models.
\newblock \emph{arXiv preprint arXiv:2408.10571}, 2024.

\end{thebibliography}
